%% file: main.tex
\PassOptionsToClass{cleveref}{jmlr}
\documentclass{l4dc2027}

\usepackage[T1]{fontenc}
\usepackage{times}
\usepackage{booktabs}
\usepackage{array}
\usepackage{nicefrac}
\usepackage{enumitem}
\usepackage{csquotes}
\usepackage{mathtools}
\usepackage{wrapfig}
\usepackage{multirow}
\usepackage{xcolor}
\usepackage{colortbl}
\usepackage{capt-of}
\usepackage{placeins}
\usepackage[nohyperlinks]{acronym}
\makeatletter
\@ifundefined{AC@verridelabel}{}{\renewcommand*{\AC@verridelabel}[1]{}}
\makeatother
\input{acronyms}

\definecolor{armPlateau}{HTML}{E69F00}\definecolor{armLearnAdd}{HTML}{D55E00}
\definecolor{armLearnMult}{HTML}{0072B2}\definecolor{armCalib}{HTML}{009E73}
\definecolor{armLinear}{HTML}{CC79A7}\definecolor{armVdot}{HTML}{56B4E9}
\definecolor{armSigmaV}{HTML}{8C5A2B}\definecolor{armNominal}{HTML}{8F8E88}
\newcommand{\swatch}[2][0.9pt]{\textcolor{#2}{\rule[0.5ex]{1.1em}{#1}}\,}
\newcommand{\dswatch}[1]{\textcolor{#1}{\rule[0.5ex]{0.45em}{0.9pt}\hspace{0.2em}\rule[0.5ex]{0.45em}{0.9pt}}\,}
\newcommand{\algcomment}[1]{\footnotesize\textit{#1}}
\newcommand{\eqptr}[1]{\text{\algcomment{$\triangleright$ #1}}}%
\newcommand{\rkA}[1]{\cellcolor{green!40}\textbf{#1}}%
\newcommand{\rkB}[1]{\cellcolor{green!22}#1}%
\newcommand{\rkC}[1]{\cellcolor{green!9}#1}%
\newcommand{\sgB}[1]{\cellcolor{green!25}#1}%
\newcommand{\oursrow}{\rowcolor{blue!10}}%
\newcommand{\sgW}[1]{\cellcolor{red!20}#1}%

\title[Task-Convergent Flow Matching]{Task Convergent Flow Matching for Generative Robotics}

\author{%
  \Name{Lorenzo Paiola}\textsuperscript{1} \Email{lorenzo.paiola@iit.it}\\
  \Name{Maximilian St{\"o}lzle}\textsuperscript{2} \Email{mstolzle@mit.edu}\\
  \Name{Giorgio Grioli}\textsuperscript{1,3} \Email{giorgio.grioli@unipi.it}\\
  \Name{Daniela Rus}\textsuperscript{2} \Email{rus@csail.mit.edu}\\
  \Name{Cosimo {Della Santina}}\textsuperscript{4} \Email{c.dellasantina@tudelft.nl}\\[0.4ex]
  \addr \textsuperscript{1}SoftBots for Human Cooperation and Rehabilitation, Istituto Italiano di Tecnologia, Genova, Italy\\
  \textsuperscript{2}Computer Science and Artificial Intelligence Laboratory, Massachusetts Institute of Technology, USA\\
  \textsuperscript{3}Research Center ``E.~Piaggio'' and Dept.\ of Information Engineering, University of Pisa, Italy\\
  \textsuperscript{4}Department of Cognitive Robotics, Delft University of Technology, The Netherlands
}

\begin{document}

\graphicspath{{./images/}}

\maketitle

\begin{abstract}
Successful manipulation requires the manipulated objects to progress toward a desired task outcome. Generative policies can produce plausible robot motions without ensuring this progress, allowing successive action chunks to stall or undo earlier changes to the scene. We introduce a task-centric inference-time refinement framework for flow matching policies that evaluates candidate action chunks through parallel physics-based rollouts and searches for motions satisfying sufficient conditions for task convergence. Task-level performance funnels bound temporary regressions while requiring net progress over each chunk. We establish sufficient conditions under which repeated funnel satisfaction yields convergence throughout execution, and use interval-wise Lipschitz bounds to verify continuous-time funnel satisfaction from sampled rollouts. To accommodate contact-rich hybrid dynamics, we refine the flow's source seeds using derivative-free Covariance Matrix Adaptation Evolution Strategy (CMA-ES), constrained to radial quantile bands of the source distribution to help retain the learned motion prior. This formulation supports both Euclidean and pose-valued Riemannian Flow Matching (RMFM) policies. On PushT and three-cube stacking, refinement raises success from 31.3\% to 88.5\% and from 77.0\% to 86.0\%. On PushT, whose coverage score responds continuously to the motion of the block, a learned funnel lowers the average maximum loss of achieved progress by 19\% relative to constraining the endpoint of each chunk alone, at a similar success rate. With the radial band, 11.3\% of the executed motion windows are out of distribution, a rate within the range of held-out demonstrations, compared with 34.1\% for an unconstrained search.
\end{abstract}


\section{Introduction}
\input{sections/intro.tex}

\section{Overview}
\input{sections/setup.tex}

\section{Task-Centric Refinement}
\input{sections/method.tex}

\section{Generation and Refinement}
\label{sec:generation_refinement}
\input{sections/background.tex}
\input{sections/optimization.tex}

\section{Experiments}
\input{sections/experiments.tex}

\section[Conclusion, Limitations, and Future Work]{Conclusion, Limitations, and Future Work}
\input{sections/conclusions.tex}
\input{sections/limitations.tex}
\clearpage
\input{sections/acks.tex}

\bibliography{main}

\clearpage
\appendix
\input{sections/appendix/related.tex}
\input{sections/appendix/manifoldflow.tex}
\input{sections/appendix/phasedlyapunov.tex}
\input{sections/appendix/funnels.tex}

\input{sections/appendix/implementationdetails.tex}

\input{sections/appendix/proofs.tex}
\input{sections/appendix/additionalresults.tex}
\input{sections/appendix/compute.tex}
\input{sections/appendix/limitations.tex}

\end{document}

%% file: acronyms.tex
\newacro{CEM}{Cross-Entropy Method}
\newacro{CMA-ES}{Covariance Matrix Adaptation Evolution Strategy}
\newacro{CVaR}{Conditional Value at Risk}
\newacro{DD}{Drawdown}
\newacro{DiT}{Diffusion Transformer}
\newacro{DLS}{Damped Least Squares}
\newacro{EE}{End Effector}
\newacro{EMA}{Exponential Moving Average}
\newacro{FiLM}{Feature-Wise Linear Modulation}
\newacro{GELU}{Gaussian Error Linear Unit}
\newacro{IK}{Inverse Kinematics}
\newacro{KID}{Kernel Inception Distance}
\newacro{KL}{Kullback--Leibler}
\newacro{L-BFGS}{Limited-Memory Broyden--Fletcher--Goldfarb--Shanno}
\newacro{LDJ}{Log Dimensionless Jerk}
\newacro{MLP}{Multilayer Perceptron}
\newacro{MPC}{Model Predictive Control}
\newacro{MPPI}{Model Predictive Path Integral}
\newacro{OOD}{Out of Distribution}
\newacro{OT}{Optimal Transport}
\newacro{PCA}{Principal Component Analysis}
\newacro{PD}{Proportional--Derivative}
\newacro{RKHS}{Reproducing Kernel Hilbert Space}
\newacro{RMFM}{Riemannian Flow Matching}
\newacro{Sep-CMA-ES}{Separable CMA-ES}
\newacro{SiLU}{Sigmoid Linear Unit}
\newacro{adaLN-Zero}{Adaptive Layer Normalization with Zero Initialization}

%% file: sections/intro.tex
\begin{figure}[t]
    \centering
    \includegraphics[width=\linewidth]{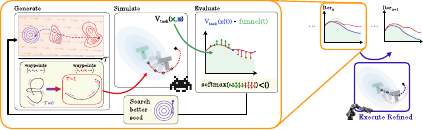}
    \caption{Generative policy candidates are evaluated through their predicted task evolution, refined against a performance funnel, and executed before replanning from the new scene.}
    \vspace{-0.9cm}
    \label{fig:overview}
\end{figure}
Diffusion and flow matching policies learn multimodal distributions over action chunks, enabling complex manipulation behaviors from demonstrations~\citep{chiDiffusionPolicyVisuomotor2024,rouxelFlowMatchingImitation2024,blackpi0VisionLanguageActionFlow2024a}. Yet plausible motions do not by themselves ensure task completion: successive chunks can stall or undo earlier progress. For example, a robot folding a shirt may generate plausible grasping and sweeping motions while repeatedly disturbing an earlier fold. Progress must therefore be assessed through the evolution of the manipulated objects, whose desired arrangement defines completion even when many robot configurations are compatible with it.

Existing methods refine generative plans at inference time to satisfy safety and feasibility constraints, using barrier functions or constrained sampling~\citep{xiaoSafeDiffuserSafePlanning2023a,zhangConstrainedDiffusersSafe2025a}. Stability-oriented methods also guide generation or constrain learned motion dynamics through conditions that encourage or require a Lyapunov function to decrease at each step~\citep{mizutaCoBLDiffusionDiffusionBasedConditional2024,perez-dattariLetDynamicsFlow2026a}. These approaches address collision avoidance, admissible trajectories, and convergence of robot motion. Such requirements do not by themselves ensure that successive action chunks advance the manipulated objects toward task completion.

A recent line of work instead evaluates and refines generated actions through their consequences predicted by learned world models~\citep{qiInferenceTimeEnhancementGenerative2025,duDynaGuideSteeringDiffusion2025}. Task-oriented objectives include predicted progress or distance to a goal at the end of a chunk or planning horizon~\citep{yanProgressVLAProgressGuidedDiffusion2026,hegdeLeWAM2026}, as well as accumulated rewards and terminal value estimates~\citep{hamelFlowMPCImprovingFlow2026}. These objectives improve predicted outcomes, but do not specify how much intermediate regression is acceptable or provide sufficient conditions for convergence under repeated execution. In particular, a favorable endpoint can conceal substantial loss of progress within the chunk.

What remains unresolved is how to formulate task-progress conditions for successive action chunks of generative manipulation policies that accommodate temporary setbacks while ensuring convergence, and how to use these conditions for inference-time refinement through physics-based simulation. We introduce a framework for flow matching policies that evaluates candidate action chunks through parallel simulated world rollouts and searches for motions satisfying sufficient conditions for task convergence. The pretrained policy supplies candidate motions, while simulation predicts their effects on the manipulated task before execution. This framework is supported by three contributions: task-level performance funnels, verification from sampled rollouts, and constrained optimization of the flow's source seed.

First, manipulation can require temporary setbacks, such as repositioning fabric before completing a fold~\citep{maoARMAdvantageReward2026}. We introduce task-level performance funnels that bound these setbacks while requiring net progress over each action chunk, and establish sufficient conditions under which repeated funnel satisfaction yields convergence throughout execution. These funnels build on finite-step Lyapunov conditions, which permit intermediate increases while requiring net decrease~\citep{norooziControlDiscretetimeNonlinear2019}, and on funnel \ac{MPC}, which constrains transient error~\citep{bergerFunnelModelPredictive2022}, applying both ideas to task error over action chunks.

Second, sampled rollouts can miss funnel violations between prediction times. We use interval-wise Lipschitz bounds~\citep{leraAccelerationUnivariateGlobal2013} to verify satisfaction throughout continuous execution, and introduce a practical variant based on locally estimated rates of change.

Third, contact-rich hybrid dynamics motivate sampling-based optimization. To exploit parallel candidate evaluation while seeking effective refinement within few sequential iterations, we use the \ac{CMA-ES}~\citep{hansenCMAEvolutionStrategy2023}. We constrain the search to a radial band of the source distribution's typical set to help retain the learned motion prior, extending earlier approaches based on noise standardization and typical-radius projection~\citep{stolzlePhysibleMDMEnforcingInferenceTime2026,hegdeLeWAM2026}.

In summary, our contributions are
\begin{enumerate}[noitemsep]
    \item \textbf{Inference-time refinement for task convergence.} A framework for flow matching policies that evaluates candidate action chunks through parallel physics-simulated world rollouts and searches for motions satisfying sufficient conditions for task convergence.
    \item \textbf{Task-level performance funnels.} Bounds on task error that allow temporary setbacks while requiring net progress over each action chunk, with sufficient conditions for task convergence throughout execution.
    \item \textbf{Verification from sampled rollouts.} A procedure for checking continuous-time task-funnel satisfaction using interval-wise Lipschitz bounds, with a practical variant based on locally estimated residual rates.
    \item \textbf{Seed refinement with radial quantile bands.} An extension of prior seed-refinement methods~\citep{stolzlePhysibleMDMEnforcingInferenceTime2026,hegdeLeWAM2026} that replaces fixed-radius normalization with radial quantile bands and applies \ac{CMA-ES} in the Gaussian source coordinates of both Euclidean and pose-valued \ac{RMFM} policies.
\end{enumerate}
An extended discussion of related work is provided in Appendix~\ref{app:related_work}.

%% file: sections/setup.tex
\label{sec:overview}

\begin{wrapfigure}{L}{0.65\textwidth}
\vspace{-\baselineskip}
\begin{minipage}{0.630\textwidth}
    \small
    \begin{equation}
        \begin{aligned}
            \min_{z_0}\quad & \colorbox{red!15}{$\displaystyle \sup_{t\in[0,H]}\!\left[V(T(x_w^k(t)))-\rho_k(t)\right]$} && \eqptr{\S\ref{sec:task_refinement}, \ref{sec:generation_refinement}}\\[1ex]
            \text{s.t.}\quad
            & \colorbox{blue!15}{$\displaystyle w=\psi_\theta(1\mid z_0,o_k)$} && \eqptr{\S\ref{sec:flow_prior}}\\[1ex]
            & \colorbox{green!15}{$\displaystyle u(t)=\kappa(x_w^k(t),w,t)$} && \\[1ex]
            & \colorbox{orange!15}{$\displaystyle \dot{x}_w^k(t)=f(x_w^k(t),u(t)),\ x_w^k(0)=x_k$} && \eqptr{\S\ref{sec:world_task}}
        \end{aligned}
        \label{eq:overview_optimization}
    \end{equation}
\end{minipage}
\caption{\raggedright\colorbox{red!15}{Whole-horizon task residual}, \colorbox{blue!15}{Flow prior}, \colorbox{green!15}{Tracking controller}, \colorbox{orange!15}{World dynamics}.}
\vspace{-1\baselineskip}
\end{wrapfigure}
Our aim is to turn the motions proposed by a learned policy into sustained progress toward task completion. At each planning step $k$, the policy observes the scene and proposes a finite-horizon motion, which is refined against its predicted effect on the task and executed before a new observation starts the next step (Figure~\ref{fig:overview}).

\paragraph{Task Progress.}
A task-error measure $V$ quantifies how far the scene is from the desired outcome. We prescribe an envelope for this error over the duration of a motion and evaluate whether the predicted task evolution stays within it. The envelope allows useful temporary regressions while requiring a net improvement by the end. Repeating this requirement connects individual motion choices to task convergence.

\paragraph{Refining a policy proposal.}
Conditioned on the observation history $o_k$, the learned flow map $\psi_\theta$ transports a random seed $z_0$ into a waypoint action chunk $w=\psi_\theta(1\mid z_0,o_k)$. Refinement searches over the seed while leaving this map fixed. A low-level controller $\kappa$ tracks the chunk's waypoints and produces the applied input $u(t)$, e.g. motor torques. In \eqref{eq:overview_optimization}, $x_w^k(t)$ is the world trajectory from the current state $x_k$, $T$ extracts its task state, $V$ measures task error, and $\rho_k$ is the envelope over the horizon $[0,H]$. The objective measures the largest departure from the desired task evolution over the entire horizon, and a plan with nonpositive worst-case residual satisfies the funnel throughout its rollout. The search ranks plans by a smooth maximum of the sampled residuals, which accounts for the worst residual together with the values close to it (\Cref{sec:optimization}).

%% file: sections/method.tex
\label{sec:task_refinement}
\paragraph{World dynamics, task state, and task potential.}
\label{sec:world_task}
Let $x(t)$ denote the state of the robot and its environment, $u(t)$ the applied input, and $w$ a waypoint action chunk, which the controller $\kappa$ tracks according to its execution timing. Under the controlled dynamics of \eqref{eq:overview_optimization}, at iteration $k$ the plan induces a rollout $x_w^k(t)=\mathcal{R}_\kappa(x_k,w;t)$ from the current world state over $t\in[0,H]$, including tracking and contact dynamics (Appendix~\ref{app:simulator_regularity} states the regularity assumptions).

We formalize task completion through the manipulated objects rather than a desired robot configuration. In stacking, for instance, the relative cube placement describes the target, while the arm can take many configurations compatible with it. We introduce a task state $y=T(x)\in\mathcal{Y}$ and a goal set $\mathcal{G}\subseteq\mathcal{Y}$.\footnote{The map $T$ may be analytical or learned. Appendix~\ref{app:simulator_regularity} states the regularity assumptions on its composition with the task measure, and Appendix~\ref{app:phased_lyapunov} gives the geometric construction used here.} The task rollout is $y_w^k(t)=T(x_w^k(t))$. Progress is measured by a task-level Lyapunov candidate $V:\mathcal{Y}\to\mathbb{R}_{\ge0}$, which we call the \emph{task potential}, with goal set $\mathcal{G}=\{y:V(y)\le v_{\mathrm{goal}}\}$, $v_{\mathrm{goal}}\ge0$.
An exact goal has $v_{\mathrm{goal}}=0$, with $V$ positive outside $\mathcal{G}$, and a positive threshold describes completion within tolerance. The potential encodes the goal and a useful ordering of intermediate task states, but it need not decrease under any prescribed controller.

\paragraph{Performance funnels and task convergence.}
\label{sec:funnels}
Although decreasing $V$ expresses progress toward the goal, requiring it to decrease at every instant would exclude useful preparatory motions. We therefore describe acceptable task evolution through a nonnegative envelope $\rho_k:[0,H]\to\mathbb{R}_{\ge0}$. Starting from the current value $V_k=V(y_k)$, this envelope allows temporary increases while requiring a smaller terminal value, so that those maneuvers produce net progress:
\begin{equation}
    \vcenter{\hbox{\includegraphics[width=0.19\linewidth]{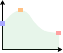}}}
    \quad
    \begin{cases}
        \colorbox{blue!15}{$\rho_k(0)=V_k$}, & \text{initial anchoring},\\[3pt]
        \colorbox{orange!15}{$\sup_{t\in[0,H]}\rho_k(t)\le(1+d_k)V_k$}, & \text{bounded transient regression},\\[3pt]
        \colorbox{red!15}{$\rho_k(H)\le(1-\gamma_k)V_k$}, & \text{terminal contraction},
    \end{cases}
    \label{eq:envelope}
\end{equation}
with $\gamma_k\in[0,1]$ and $d_k\ge0$. We ask the candidate motion to remain within this envelope throughout execution:
\begin{equation}
    \boxed{V(y_w^k(t))\le\rho_k(t)\qquad\text{for every }t\in[0,H].}
    \label{eq:lyapfunnel}
\end{equation}
The terminal condition controls progress at replanning times, while the transient bound limits how far the task may regress between those decisions. Any envelope that satisfies \eqref{eq:envelope} can be used, fixed or chosen anew at each iteration.

\begin{theorem}[Task convergence throughout execution]
    \label{thm:receding_horizon_contraction}
    Suppose each selected chunk $w_k^\star$ satisfies its funnel \eqref{eq:lyapfunnel} with an envelope obeying \eqref{eq:envelope}, is executed over the full horizon $H>0$ before replanning, and has an evaluated task rollout that matches execution. If the terminal contraction is uniformly positive, $\gamma_k\ge\underline\gamma>0$, and the relative transient allowance is uniformly bounded, $d_k\le\bar d<\infty$, then
    \begin{equation}
        \sup_{t\in[0,H]}V(y_{w_k^\star}^k(t))
        \le(1+\bar d)(1-\underline\gamma)^kV(y_0).
        \label{eq:continuous_task_convergence}
    \end{equation}
    Task error therefore converges to zero throughout execution. For $v_{\mathrm{goal}}>0$, execution enters $\mathcal{G}$ after finitely many chunks and remains there while these conditions hold.
\end{theorem}
The terminal contraction reduces the task error at replanning times, and the transient bound controls the error between them. Appendix~\ref{app:convergence_proof} gives the proof and the product form for iteration-dependent rates. It also extends the result to intermittent progress, history-dependent funnel parameters, a fading start allowance and bounded endpoint errors.

\paragraph{Continuous-time verification from samples.}
\label{sec:verification}

Checking \eqref{eq:lyapfunnel} directly requires the task state at every instant of the rollout, while numerical simulation supplies it only at finitely many instants. We therefore write the requirement through the residual and its worst value over the horizon,
\begin{equation}
    g_k(w,t)=V(y_w^k(t))-\rho_k(t),\qquad
    S_k(w)=\sup_{t\in[0,H]}g_k(w,t),
    \label{eq:rollout_residual}
\end{equation}
so that the funnel holds over the whole horizon exactly when $S_k(w)\le0$. Simulating $\mathcal{R}_\kappa$ gives the residual at instants $0=t_1<\cdots<t_n=H$, with spacings $\Delta_i=t_{i+1}-t_i$, samples $g_i=g_k(w,t_i)$, and largest sampled value $M_k(w)=\max_ig_i$. Requiring $M_k(w)\le0$, which we call sampled verification, checks the funnel at the samples. Verification over the whole horizon additionally accounts for the residual between consecutive samples, which the following result bounds through the regularity of the residual in physical time.

\begin{theorem}[Continuous-time verification from samples]
    \label{thm:objective_certification}
    \label{thm:lipschitz_interval_certification}
    Fix an iteration $k$ and a plan $w$, and suppose that $g_k(w,\cdot)$ is $L_i$-Lipschitz on each interval $[t_i,t_{i+1}]$. Then,
    \begin{equation}
        S_k(w)\le B_k(w)=\max_{1\le i<n}\frac{g_i+g_{i+1}+L_i\Delta_i}{2}.
        \label{eq:interval_bound}
    \end{equation}
    Consequently,
    \begin{equation}
        \boxed{B_k(w)\le0
        \quad\Longrightarrow\quad
        V(y_w^k(t))\le\rho_k(t)\quad\forall t\in[0,H].}
        \label{eq:verification_condition}
    \end{equation}
\end{theorem}
On each interval, the residual lies below the two lines of slope $L_i$ through its samples, and these lines meet at the value in \eqref{eq:interval_bound} (proof in Appendix~\ref{app:verification_proof}). Appendix~\ref{app:simulator_regularity} bounds the rates $L_i$ through the task geometry and a bounded configuration speed. In the experiments we estimate each interval's rate from the trace being judged, as the largest secant slope of the residual among the nearby intervals. Substituting these local rates $\hat L_i$ into \eqref{eq:interval_bound} gives \emph{estimated verification}, the procedure we use and report (Appendix~\ref{app:local_rates}). The guarantee of \eqref{eq:verification_condition} holds whenever the rates upper-bound the true residual rates, and estimated verification applies the same bound with sample-derived rates.

%% file: sections/background.tex
\paragraph{Flow-matching prior.}
\label{sec:flow_prior}
Given demonstrations $\mathcal{D}=\{(o_i,w_i)\}$ of observation histories and action chunks, a conditional flow-matching policy learns a vector field $v_\theta(z,\tau\mid o)$~\citep{lipmanFlowMatchingGenerative2023,lipmanFlowMatchingGuide2024}. Integrating $d\psi_\theta/d\tau=v_\theta(\psi_\theta,\tau\mid o)$ from a source seed, $\psi_\theta(0\mid z_0,o)=z_0$, to $\tau=1$ decodes the seed into an action chunk $w=\psi_\theta(1\mid z_0,o)$. Generative time $\tau$ parameterizes the construction of the entire chunk, and physical time $t$ describes the execution of the resulting motion. The framework supports Euclidean actions, trained with the \ac{OT} path, and pose-valued actions, trained with \ac{RMFM}~\citep{chenFlowMatchingGeneral2024,braunRiemannianFlowMatching2024} from a data-centered wrapped-normal rotational source (Appendix~\ref{sec:appendix_lie_group_preliminaries}). With the policy, the observation and the sampler fixed, decoding is a deterministic map from seed to chunk, and refinement searches over its input.

%% file: sections/optimization.tex
\paragraph{Search objective.}
\label{sec:optimization}
\label{sec:calibration}
The search ranks candidate plans by a smooth maximum of their sampled residuals, the unnormalized log-sum-exp
\begin{equation}
    J_k(w)=\beta^{-1}\log\textstyle\sum_{i=1}^n e^{\beta g_i}.
    \label{eq:lse_objective}
\end{equation}
It accounts for the worst residual and for the values close to it, and the temperature $\beta>0$ controls how closely it approximates the largest sampled residual $M_k$~\citep[\S3.1.5]{boydConvexOptimization2004}. This smoothing acts across the samples of one rollout and serves the ranking of candidates. The funnel verdict of a plan comes from estimated verification, which uses the regularity of the residual in physical time (\Cref{sec:task_refinement}; Appendix~\ref{app:local_rates}). Each planning call starts from the nominal proposal of the policy, keeps the lowest-$J_k$ plan among all evaluated candidates (Algorithm~\ref{alg:cmaes_refinement}), and executes it together with its verification status. The results of \Cref{sec:task_refinement} apply to the executed chunks that satisfy their funnel. Appendix~\ref{sec:appendix_cmaes_refinement} gives the stopping rules.

\paragraph{Radial quantile band.}
Seeds drawn from a high-dimensional standard Gaussian concentrate on a thin shell around a typical radius, and the flow is trained to transport seeds from this shell. Prior work fixes the radius of the seed to its typical value~\citep{stolzlePhysibleMDMEnforcingInferenceTime2026,hegdeLeWAM2026}. We extend this projection to a radial band, so that the radius of the seed varies within a prescribed quantile interval of the shell. For standard Gaussian seed coordinates $a\in\mathbb R^d$, the projection onto the centered band is
\begin{equation}
    \Psi(a)=(a-\bar a\mathbf{1})\,
    \frac{\operatorname{clip}(\|a-\bar a\mathbf{1}\|,r_{\mathrm{lo}},r_{\mathrm{hi}})}
         {\|a-\bar a\mathbf{1}\|},
    \qquad \bar a=d^{-1}\sum_{j=1}^{d}a_j,
    \label{eq:seed_shell}
\end{equation}
which shifts the entries to zero mean and clips the radius. The radii $r_{\mathrm{lo}}$ and $r_{\mathrm{hi}}$ are the quantiles of the source radius at tail probabilities $\Phi(-\zeta)$ and $\Phi(\zeta)$, with $\Phi$ the standard normal distribution function (Appendix~\ref{sec:appendix_cmaes_refinement}). A typical seed therefore keeps its radius and is centered. The band encourages the search to stay where the flow was trained, so that the seeds decode into chunks that resemble the demonstrated motions, as \Cref{sec:experiments} measures.

\paragraph{Band-constrained \acs*{CMA-ES}.}
Contact-rich rollouts are evaluated in parallel, which suits a population search. \ac{CMA-ES}~\citep{akimotoTheoreticalFoundationCMAES2012,hansenCMAEvolutionStrategy2023} samples $a_\ell\sim\mathcal N(m,\sigma^2C)$, $\ell=1,\dots,\lambda$, and ranks the samples by $F_k(a)=J_k\bigl(\psi_\theta(1\mid\mathcal C(\Psi(a)),o_k)\bigr)$. The seed chart $\mathcal C$ maps the coordinates to a seed $z_0$ of the source. It is the identity for a Euclidean flow and the exponential chart of the wrapped-normal source for a pose-valued flow, in which the source is exactly $\mathcal N(0,I_d)$ (Appendix~\ref{sec:appendix_cmaes_refinement}). As in the standardized \ac{Sep-CMA-ES} of \citet{stolzlePhysibleMDMEnforcingInferenceTime2026}, every sampled seed is projected, and we additionally project the mean, $m\leftarrow\Psi(m)$, after each update.

\begin{figure}[t]
\SetCommentSty{algcomment}
\SetKwProg{Fn}{Function}{}{}
\SetKwFunction{Cost}{Cost}\SetKwFunction{Decode}{Decode}\SetKwFunction{Refine}{Refine}
\SetKwComment{tcp}{\(\triangleright\) }{}
\begin{algorithm2e}[H]
    \scriptsize
    \caption{Task-funnel refinement of generative action chunks.}
    \label{alg:cmaes_refinement}
\end{algorithm2e}
\vspace{-8pt}
\RestyleAlgo{plain}
\begin{minipage}[t]{0.55\linewidth}
\begin{algorithm2e}[H]
    \scriptsize
    \DontPrintSemicolon
    \Fn{\Refine{}\hfill{\algcomment{\(\triangleright\) Sec.~\ref{sec:optimization}}}}{
        \For(\tcp*[f]{execution}){$k=0,1,\dots$}{
        Observe $x_k,o_k$; $V_k\leftarrow V(T(x_k))$\;
        Build and freeze \colorbox{green!15}{$\rho_k$} \tcp*[r]{Sec.~\ref{sec:funnels}}
        $w^\star\leftarrow\Decode(a^{\mathrm{nom}};o_k)$; $J^\star\leftarrow\Cost(w^\star;x_k,\rho_k)$\;
        Initialize $(m,\sigma,C)$ on the band\;
        \While(\tcp*[f]{App.~\ref{sec:appendix_cmaes_refinement}}){stopping condition not met}{
            $a_\ell\sim\mathcal N(m,\sigma^2C)$, $\ell=1,\dots,\lambda$\;
            $w_\ell\leftarrow\Decode(a_\ell;o_k)$; $J_\ell\leftarrow\Cost(w_\ell;x_k,\rho_k)$\;
            \lIf{$\min_\ell J_\ell<J^\star$}{$(w^\star,J^\star)\leftarrow$ lowest-$J$ candidate}
            Rank-update $(m,\sigma,C)$ from $J_{1:\lambda}$; $m\leftarrow\Psi(m)$\;
        }
        Execute $w^\star$ over $[0,H]$; the episode ends when the success predicate holds\;
        }
    }
\end{algorithm2e}
\end{minipage}\hfill
\begin{minipage}[t]{0.43\linewidth}
\begin{algorithm2e}[H]
    \scriptsize
    \DontPrintSemicolon
    \Fn{\Cost{$w;x_k,\rho_k$}\hfill{\algcomment{\(\triangleright\) Sec.~\ref{sec:verification}}}}{
        $x(t_i)\leftarrow$ \colorbox{orange!15}{$\mathcal R_\kappa(x_k,w;t_i)$}, $i=1,\dots,n$\;
        $g_i\leftarrow$ \colorbox{green!15}{$V(T(x(t_i)))-\rho_k(t_i)$}\;
        \KwRet{\colorbox{red!15}{$J_k(g_{1:n})$}} \tcp*[r]{Eq.~\eqref{eq:lse_objective}}
    }
    \Fn{\Decode{$a;o_k$}\hfill{\algcomment{\(\triangleright\) Sec.~\ref{sec:flow_prior}}}}{
        \KwRet{\colorbox{blue!15}{$\psi_\theta(1\mid\mathcal C(\Psi(a)),o_k)$}}\;
    }
    \vspace{2pt}{\scriptsize\colorbox{green!15}{funnel} \colorbox{red!15}{objective} \colorbox{blue!15}{flow prior} \colorbox{orange!15}{rollout}}\;
\end{algorithm2e}
\end{minipage}
\par\vspace{2pt}\hrule height 0.8pt
\end{figure}

%% file: sections/experiments.tex
\label{sec:experiments}
We compare nominal execution, the flow-matching policy without refinement, with refined execution on planar \texttt{PushT} and three-cube stacking in \texttt{robosuite}~\citep{zhuRobosuiteModularSimulation,todorovMuJoCoPhysicsEngine2012}, with stacking demonstrations from MimicGen \texttt{StackThree\_D1}~\citep{mandlekarMimicGenDataGeneration2023}. Appendix~\ref{sec:appendix_env_details} gives the environment and execution details, and Appendix~\ref{sec:additional_results} reports further ablations and two-nut assembly.

\paragraph{Task progress and evaluation.}
Refinement is guided by the task potential $V$, where lower is better: a weighted pose error on the single-stage PushT (Appendix~\ref{app:pusht_candidate}) and a phased potential on stacking, which also orders the intermediate task states (Appendix~\ref{app:phased_calibration}). We measure the physical outcome with each environment's own task score $r\in[0,1]$, where higher is better, and its success predicate. On PushT, $r$ is the coverage of the goal by the block, with success at $r>0.95$. Coverage changes continuously with the motion of the block, so the score registers progress within a chunk. On stacking, $r=\tfrac12a_{RG}(1+a_{BR})$, where $a_{XY}$ is the footprint overlap of cube $X$ on cube $Y$ times their contact, and success requires the tower to stand after the gripper releases it and retreats. The score also measures partial progress in episodes that do not succeed.

\paragraph{Policies.}
On PushT we follow Diffusion Policy~\citep{chiDiffusionPolicyVisuomotor2024} and use its one-dimensional U-Net conditioned by \ac{FiLM}, trained with the Euclidean \ac{OT} path. On stacking we use a transformer based on the \ac{DiT} architecture~\citep{peeblesScalableDiffusionModels2023} trained with \ac{RMFM}~\citep{chenFlowMatchingGeneral2024,braunRiemannianFlowMatching2024} along screw-motion paths on \ac{EE} poses (Appendix~\ref{sec:appendix_lie_group_preliminaries}). Architectures and training settings are in Appendix~\ref{app:policy_training}.

\paragraph{Refinement objectives.}
All refined runs use the same framework, with the same potential, rollouts and search, and differ in the objective (Appendix~\ref{app:adaptive_funnel}). Three objectives correspond to those of prior work. \emph{Endpoint-only} keeps the terminal contraction of the funnel and constrains $V$ at the end of the chunk, as progress- and goal-based objectives do~\citep{yanProgressVLAProgressGuidedDiffusion2026,hegdeLeWAM2026}. \emph{$\dot V\le0$} asks $V$ to decrease throughout the chunk, as stepwise Lyapunov conditions do~\citep{mizutaCoBLDiffusionDiffusionBasedConditional2024,perez-dattariLetDynamicsFlow2026a}. \emph{$\Sigma V$} sums $V$ over the chunk, as a model-predictive cost does~\citep{hamelFlowMPCImprovingFlow2026}. The \emph{plateau} funnel bounds the interior uniformly by $(1+\delta)V_k$, and the \emph{linear} funnel prescribes steady progress. The remaining funnels are fitted to demonstrated traces of $V$ and share one shape, which descends to $(1-\gamma_k)V_k$ and adds an allowance that vanishes by the end of the chunk (Appendix~\ref{app:adaptive_funnel}). \emph{Calibrated} funnels use constant parameters fitted over the whole task (\emph{global}). \emph{Learned} funnels predict the parameters from the observation with a small network, with allowances in units of $V$ (\emph{additive}) or as fractions of $V_k$ (\emph{multiplicative}). Fitting uses demonstrations only, never refined rollouts, so a fitted funnel describes the regressions and progress of demonstrated behavior. A funnel is more lenient when it requests less terminal progress or allows $V$ to rise further within the chunk. The leniency $\alpha$ of a learned funnel is a quantile level: given the observation, a fraction $\alpha$ of the demonstrated chunks reaches the requested progress and a fraction $\alpha$ stays within the allowance, so a larger $\alpha$ gives a more lenient funnel (Appendix~\ref{app:funnel_fitting}).

\paragraph{Protocol and metrics.}
An episode runs until the success predicate holds or its step budget is spent, and every chunk is executed in full. All configurations share 400 initial conditions and are paired: in each episode they use the same initial state, random seeds and initial source draw, from which refinement starts. The policy is trained on all demonstrations, and the fitted funnels on a separate split (Appendix~\ref{app:funnel_fitting}). We report success and the $\text{\acs*{CVaR}}_{10}$ of the final score, the mean over the 10\% of episodes with the lowest score~\citep{rockafellarOptimizationConditionalValueatrisk2000}. Progress retention is measured by the maximum drawdown $\max_t(\max_{s\le t}r_s-r_t)$, the largest fall of the score below its running maximum, in points ($\times100$), through its mean and its $\text{\acs*{CVaR}}_{90}$, the mean over the 10\% of episodes with the largest drawdown. For funnels with an interior bound, we report the share of executed chunks that pass estimated verification. A difference is significant when its paired-bootstrap 95\% interval excludes zero (Appendix~\ref{app:statistics}).

\begin{table}[t]
    \caption{Funnel ablation, selected configurations; best refined configuration in bold. Verified: share of complete executed chunks that pass estimated verification. Verified applies to funnels with an interior bound. Right: PushT drawdown exceedance; lower curves mean fewer episodes with large losses of achieved progress.}
    \label{tab:main_results}
    \centering
    \begin{minipage}[c]{0.57\linewidth}
    \centering
    \scriptsize
    \def\twol#1#2{\begin{tabular}[c]{@{}l@{}}#1\\\textcolor{black!65}{#2}\end{tabular}}%
    \setlength{\tabcolsep}{1.6pt}
    \def\gsep{\hspace{5pt}}%
    \setlength{\aboverulesep}{0.8pt}\setlength{\belowrulesep}{0.8pt}
    \setlength{\extrarowheight}{1pt}
    \begin{tabular}{@{}l@{\gsep}rr@{\gsep}rr@{\gsep}rr@{\gsep}rr@{\gsep}rr@{}}
        \toprule
        & \multicolumn{2}{c@{\gsep}}{Success} & \multicolumn{2}{c@{\gsep}}{Final score} & \multicolumn{4}{c@{\gsep}}{Max drawdown [pts] $\downarrow$} & \multicolumn{2}{c}{Verified}\\
        \cmidrule(lr){2-3}\cmidrule(lr){4-5}\cmidrule(lr){6-9}\cmidrule(l){10-11}
        Objective & \multicolumn{2}{c@{\gsep}}{[\%] $\uparrow$} & \multicolumn{2}{c@{\gsep}}{$\text{\acs*{CVaR}}_{10}\uparrow$} & \multicolumn{2}{c}{Mean} & \multicolumn{2}{c@{\gsep}}{$\text{\acs*{CVaR}}_{90}$} & \multicolumn{2}{c}{[\%] $\uparrow$}\\
        & {\tiny PushT} & {\tiny Stack.} & {\tiny PushT} & {\tiny Stack.} & {\tiny PushT} & {\tiny Stack.} & {\tiny PushT} & {\tiny Stack.} & {\tiny PushT} & {\tiny Stack.}\\
        \midrule
        \dswatch{armNominal}Nominal & 31.3 & 77.0 & .000 & .000 & 40.0 & 8.5 & 86.9 & 46.1 & \multicolumn{2}{c}{n/a}\\
        \addlinespace[1pt]
        \swatch[1.8pt]{black}Endpoint-only &86.8 & \textbf{86.0} & .560 & .084 & 28.3 & 6.8 & 62.8 & 46.1 & \multicolumn{2}{c}{n/a}\\
        \midrule[0.3pt]
        \dswatch{armVdot}$\dot V\le0$ & 60.5 & 85.5 & .218 & .103 & 23.4 & 5.8 & 54.2 & 41.8 & \multicolumn{2}{c}{n/a}\\
        \dswatch{armSigmaV}$\Sigma V$ & 80.0 & 82.0 & .499 & .000 & \textbf{22.6} & 6.4 & 54.4 & 43.4 & \multicolumn{2}{c}{n/a}\\
        \addlinespace[2pt]
        \swatch{armPlateau}\twol{Plateau}{$\delta=0.5$} & 73.0 & 83.8 & .494 & .075 & 23.9 & 6.9 & 56.7 & 46.2 & \textbf{73.3} & 41.9\\
        \swatch{armLinear}Linear & 54.5 & 78.8 & .194 & .000 & 24.3 & 7.6 & 54.7 & 44.3 & 0.5 & 3.3\\
        \swatch{armCalib}\twol{Calibrated}{global} & \textbf{88.5} & 80.5 & .587 & .001 & 27.4 & 7.5 & 62.9 & 46.2 & 60.5 & \textbf{61.2}\\
        \swatch{armLearnAdd}\twol{Learn.\ add.}{$\alpha=0.5$} & 84.8 & 85.0 & \textbf{.613} & \textbf{.162} & 22.9 & 7.6 & \textbf{52.3} & 47.0 & 38.0 & 44.3\\
        \swatch{armLearnMult}\twol{Learn.\ mult.}{$\alpha=0.95$} & 63.8 & 84.0 & .424 & .107 & 23.5 & \textbf{5.7} & 54.0 & \textbf{40.5} & 70.7 & 54.2\\
        \bottomrule
    \end{tabular}
    \end{minipage}\hfill
    \begin{minipage}[c]{0.41\linewidth}
    \centering
    \includegraphics[width=\linewidth]{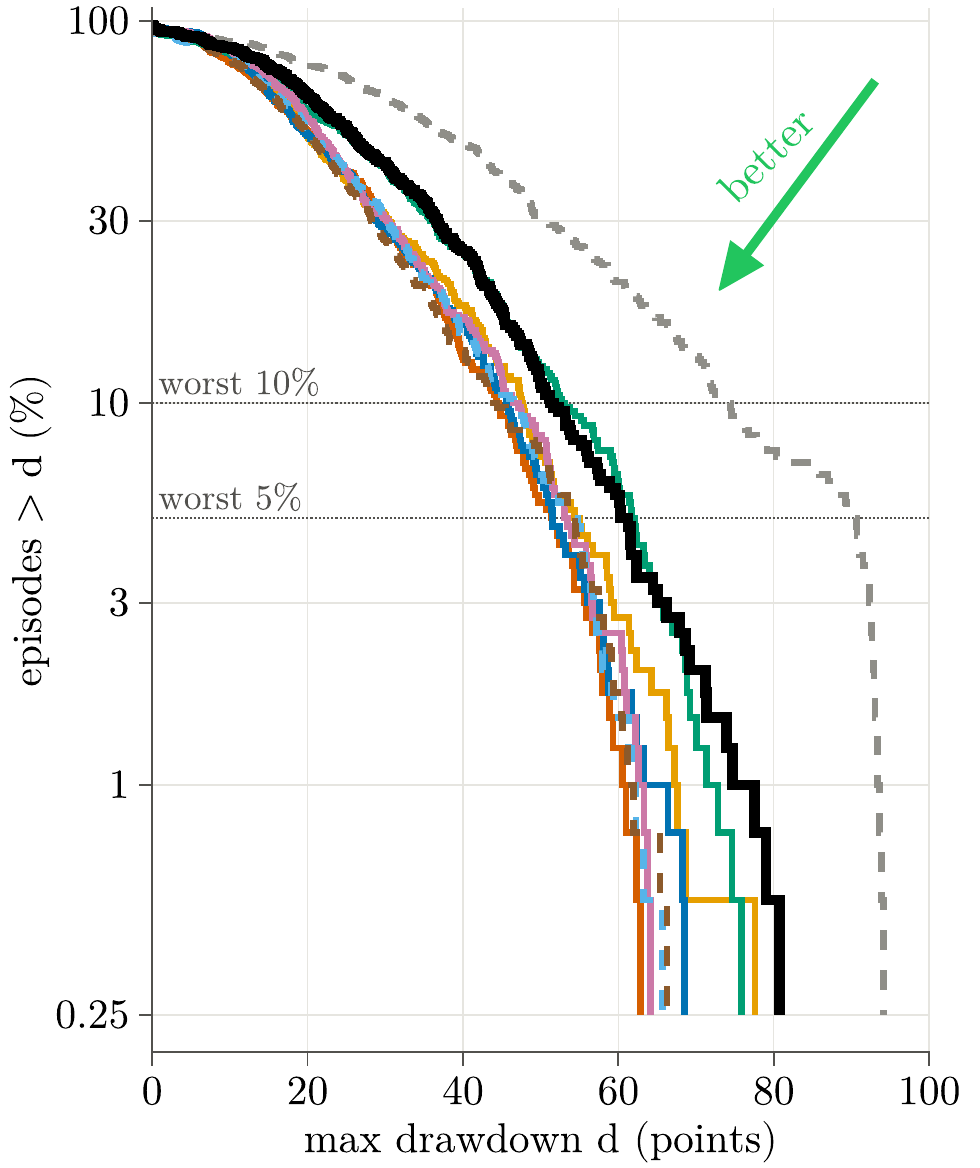}
    \end{minipage}
\end{table}

\begin{figure}[t]
    \centering
    \includegraphics[height=0.27\linewidth]{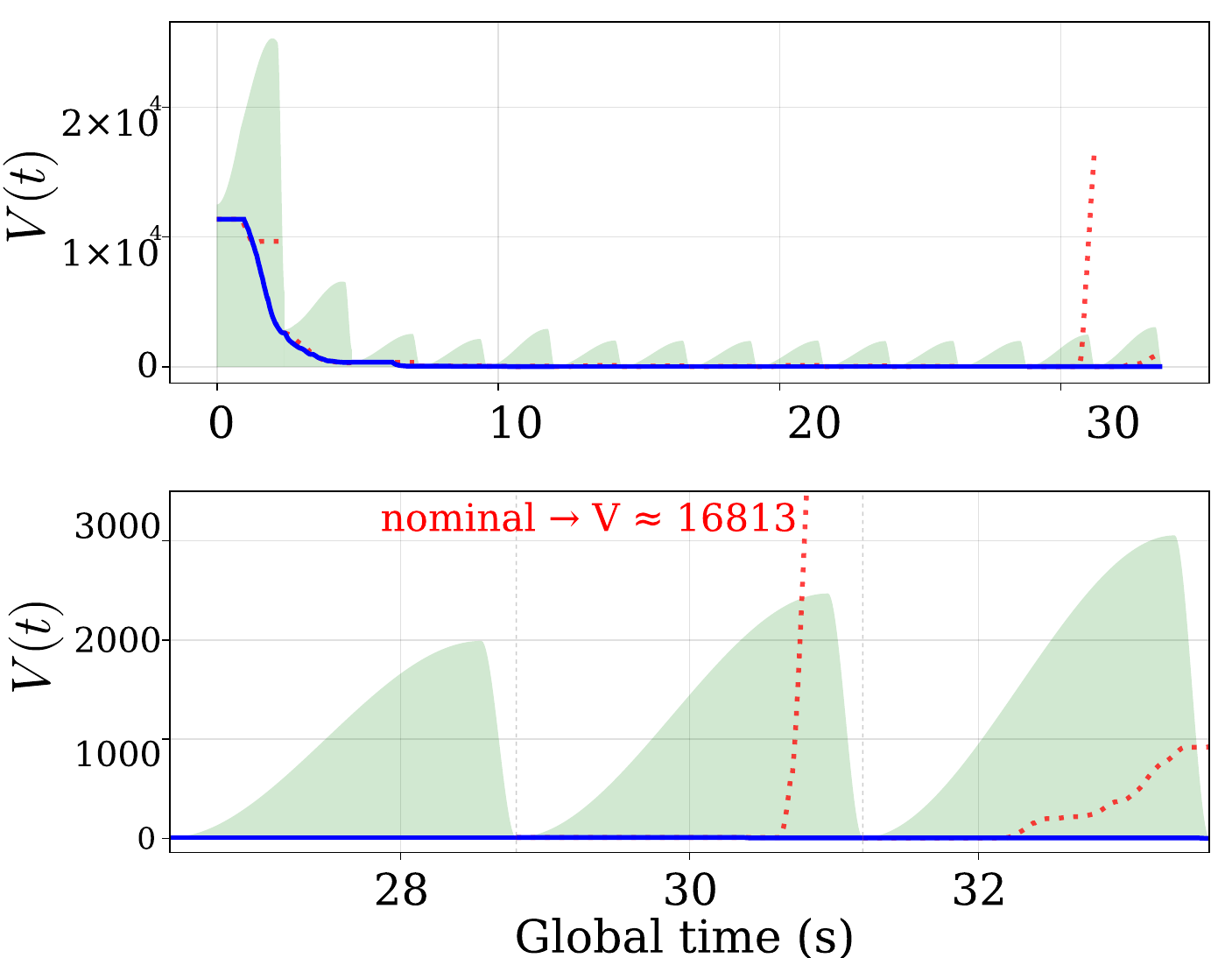}\hfill
    \includegraphics[height=0.27\linewidth]{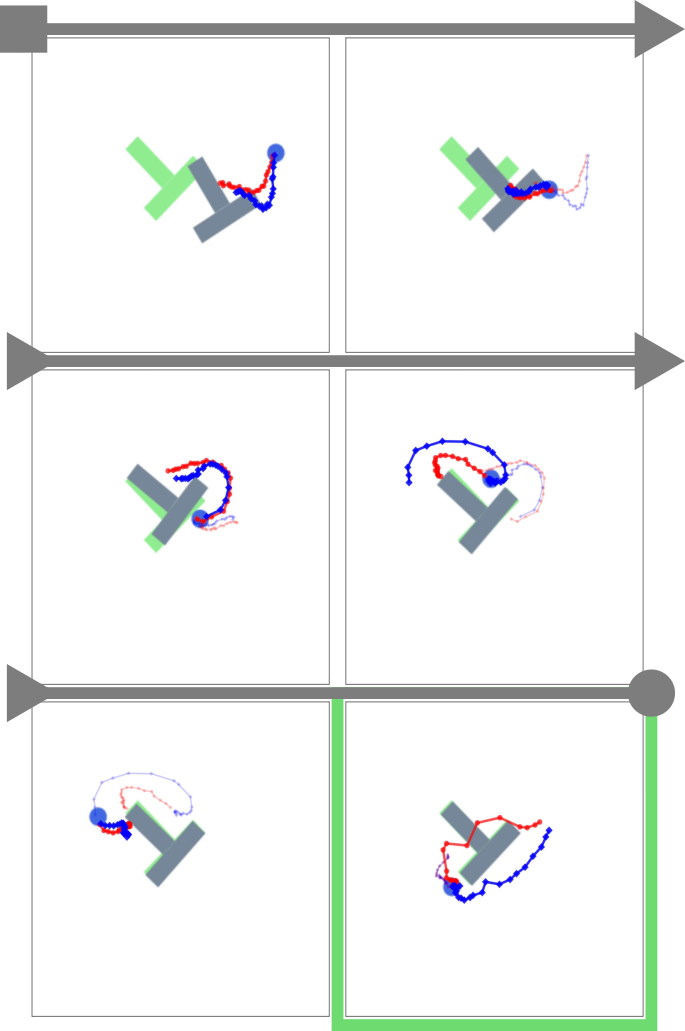}\hfill
    {\color{black!50}\vrule width 0.1pt height 0.27\linewidth}\hfill
    \includegraphics[height=0.27\linewidth]{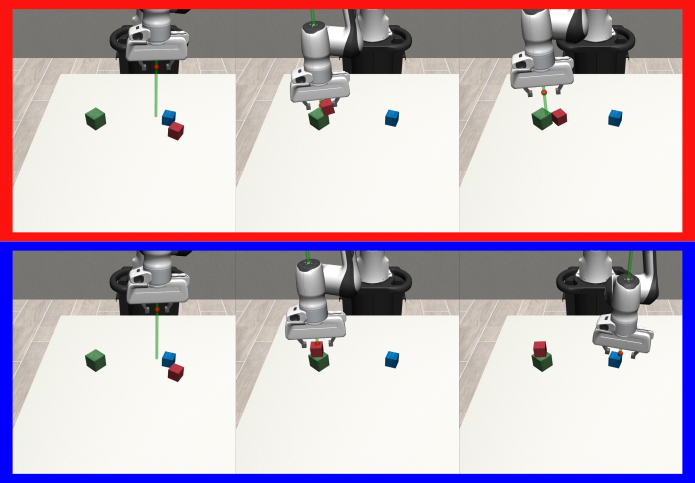}
    \caption{Representative \textcolor{blue}{refined} and \textcolor{red}{nominal} episodes on PushT (left, with $V(t)$, its funnels and a framed planning call) and stacking (right, same initial condition).}
    \label{fig:representative}
    \vspace{-0.8cm}
\end{figure}

\paragraph{Results.}
Refinement raises success above nominal execution with every objective, and the terminal contraction of the funnel already gives most of the gain. Endpoint-only refinement raises success from 31.3\% to 86.8\% on PushT and from 77.0\% to 86.0\% on stacking, and funnels that also bound the interior reach the same level, with the highest PushT success, 88.5\%, for the calibrated global funnel (Table~\ref{tab:main_results}; all configurations in Appendix~\ref{app:full_results}).

The interior bound works well when it admits the regressions that the task needs. On PushT, a useful motion often moves the T away from its target to set up a better push, so $V$ rises within the chunk. Funnels with a transient allowance accommodate this rise, and the objectives that ask $V$ to fall throughout the chunk, $\dot V\le0$ and the linear funnel, reach 54.5\% to 60.5\% success. On stacking, $V$ decreases along successful executions, and the refined configurations reach 80.5\% to 86.0\% success, except the linear funnel, which requests steady progress throughout the chunk and reaches 78.8\%.

Where the score registers progress within a chunk, the interior bound retains more of it. On PushT, every learned funnel significantly lowers the drawdown relative to endpoint-only and places the whole drawdown tail below that of nominal execution (Table~\ref{tab:main_results}, right; Appendix~\ref{app:outcome_tails}). The learned additive funnel at $\alpha=0.5$ retains the most, lowering the drawdown $\text{\acs*{CVaR}}_{90}$ from 62.8 to 52.3 at a success statistically on par with endpoint-only. On stacking, the learned multiplicative funnel at $\alpha=0.95$ significantly lowers the drawdown $\text{\acs*{CVaR}}_{90}$, from 46.1 to 40.5, at on-par success. Figure~\ref{fig:representative} shows representative refined and nominal episodes on both tasks.

Leniency trades requested progress for verification: more lenient funnels pass estimated verification on more of their executed chunks and succeed in fewer episodes (Appendix~\ref{app:full_results}). Appendix~\ref{app:potential_ablation} examines the choice of the potential.

\paragraph{Search method comparison.}
On PushT, we compare our search with other searches on 200 new initial conditions, with the same U-Net policy and the plateau funnel at $\delta=1$ as the shared objective. Ours is \ac{CMA-ES} on the band of \eqref{eq:seed_shell} at $\zeta=4$, the setting of every other experiment. \ac{CMA-ES} without the band isolates the role of the band. Best-of-$N$ decodes a single batch of 200 seeds, and \ac{CEM}~\citep{rubinsteinCrossEntropyMethod2004} is a second adaptive population search, both on the same band. \ac{L-BFGS}~\citep{nocedalNumericalOptimization2006} descends the same objective through the flow and a smoothed differentiable reimplementation of the PushT dynamics, and the population searches evaluate the objective on rollouts of the native simulator (Appendix~\ref{app:surrogate_gradient}). Each iterative search runs for 1.5\,s per chunk. We measure smoothness by the \ac{LDJ}~\citep{balasubramanianAnalysisMovementSmoothness2015} within chunks, and similarity to the demonstrations by the share of executed windows that are \ac{OOD} with respect to the training demonstrations under a nearest-neighbor test~\citep{sunOutofDistributionDetectionDeep2022}.

\begin{wraptable}{R}{0.45\linewidth}
    \vspace{-\baselineskip}
    \centering
    \caption{\raggedright Search methods on PushT. {\fboxsep=1pt\colorbox{green!25}{Better}/\colorbox{red!20}{Worse}} than {\fboxsep=1pt\colorbox{blue!10}{ours}} (significant). Significance refers to paired differences from our search. $^\dagger$Gradient.}
    \label{tab:search_ablation}
    \scriptsize
    \setlength{\tabcolsep}{2pt}
    \begin{tabular}{@{}lcccc@{}}
        \toprule
        Search & Succ.$\uparrow$ & $\text{\acs*{CVaR}}_{10}\!\uparrow$ & \acs*{LDJ}$\uparrow$ & \acs*{OOD}\%$\downarrow$\\
        \midrule
        \acs*{L-BFGS}$^\dagger$ & \sgW{70.0} & \sgW{.765} & \sgW{$-10.15$} & 12.0\\
        Best-of-$N$ & 82.5 & \sgW{.768} & $-9.87$ & 4.9\\
        \acs*{CEM} & 79.0 & .848 & \sgW{$-10.28$} & 8.0\\
        \acs*{CMA-ES}, no band & \sgB{96.0} & \sgB{.941} & \sgW{$-10.96$} & \sgW{34.1}\\
        \oursrow Ours, $\zeta{=}4$ & 86.5 & .877 & $-9.93$ & 11.3\\
        \bottomrule
    \end{tabular}
    \vspace{-\baselineskip}
\end{wraptable}
Band-constrained \ac{CMA-ES} reaches 86.5\% success, compared with 70.0\% for \ac{L-BFGS}, 82.5\% for best-of-$N$ and 79.0\% for \ac{CEM} (Table~\ref{tab:search_ablation}). Its success is significantly higher than that of \ac{L-BFGS}, and its lower-tail final score is significantly higher than those of \ac{L-BFGS} and best-of-$N$. The success differences from best-of-$N$ and \ac{CEM} have confidence intervals that contain zero. Removing the band raises success to 96.0\%, increases the share of out-of-distribution windows from 11.3\% to 34.1\%, and lowers smoothness from an \ac{LDJ} of $-9.93$ to $-10.96$. With the band, the \ac{OOD} rate lies within the range of held-out demonstrations (9.6\%, 95\% interval 5.4--15.4\%), and Table~\ref{tab:search_ablation} marks the rates above this range. The band thus trades some success for motion that stays close to the demonstrations. Appendix~\ref{app:search_ablation} reports the complete comparison, including Adam, \ac{MPPI}, the fixed-radius sphere and a sweep of the band width.

%% file: sections/conclusions.tex
We presented a task-centric inference-time refinement framework for flow matching policies, which evaluates candidate action chunks through parallel physics-based rollouts and searches the flow's source seeds for motions that satisfy sufficient conditions for task convergence. These conditions take the form of task-level performance funnels, which accommodate the temporary regressions that manipulation often requires while asking for net progress over each chunk, so that their repeated satisfaction drives task error down throughout execution and can be verified from sampled rollouts through interval-wise Lipschitz bounds. On PushT and three-cube stacking, refinement raises success far above nominal execution, as the funnel's terminal contraction steers each chunk toward net progress, and its transient bound helps retain the progress already achieved on PushT, where the task score registers it. Radial quantile bands on the source keep the refined motion close to the demonstrations.

%% file: sections/limitations.tex
\label{sec:limitations}
\paragraph{Limitations and future work.}
The quality of refinement follows the quality of the task potential $V$, which we design by hand from privileged world state. Learned progress measures are a natural route to perceptual and long-horizon tasks. Refinement also uses the simulator as its world model, so it requires the state of the scene and a simulator whose rollouts predict the executed motion. Learned world models could take this role on physical systems. The convergence results apply to executed chunks that satisfy their funnel, the search can execute a plan that does not, and estimated verification uses rates derived from the samples in place of certified bounds. Finally, planning takes seconds per chunk (Appendix~\ref{app:compute}), during which the scene has to stay still, so refinement does not yet run in real time. Appendix~\ref{app:limitations} discusses these points further.

%% file: sections/acks.tex
\acks{
This work was supported in part by the European Union’s Horizon Europe Programs from Project EMERGE (Grant No. 101070918), the Singapore MIT Alliance on Research and Technology (SMART) under the Mens, Manus, et Machina (M3S) program.

\paragraph{Use of AI.}
The authors used AI assistants to help write and debug the code, to draft and revise the text, to check the proofs, and to search for related literature. All ideas, methods, and scientific contributions are the authors' own. The authors reviewed and verified everything the assistants produced and take full responsibility for the content of this paper.
}

%% file: sections/appendix/related.tex
\section{Related Work}
\label{app:related_work}

\paragraph{Generative policies for robotics.}
Generative policies learn expressive models of multimodal robot behavior from demonstrations, representing multiple valid ways to perform a task~\citep{chiDiffusionPolicyVisuomotor2024,rouxelFlowMatchingImitation2024}. Diffusion and flow matching support conditioning on robot state, images, point clouds, and language instructions~\citep{chiDiffusionPolicyVisuomotor2024,chisariLearningRoboticManipulation2024,blackpi0VisionLanguageActionFlow2024a}, allowing motion generation to respond to both the scene and the requested task. These policies commonly generate action chunks: sequences of future commands whose initial segment is executed before replanning from updated observations in a receding-horizon loop~\citep{chiDiffusionPolicyVisuomotor2024,blackpi0VisionLanguageActionFlow2024a}. Actions often specify desired joint configurations or end-effector poses over time, so a chunk describes a robot motion sequence tracked by a lower-level controller~\citep{chiDiffusionPolicyVisuomotor2024,blackpi0VisionLanguageActionFlow2024a}. Manifold-aware formulations additionally account for the geometry of pose-valued actions~\citep{braunRiemannianFlowMatching2024,dingFastRobustVisuomotor2025}. We use a pretrained flow matching policy as a fixed prior over waypoint chunks.

\paragraph{Inference-time refinement of generative policies.}
Inference-time refinement can act on generated motions, intermediate sampling states, or source noise. Direct motion refinement uses generated action sequences to initialize trajectory optimization~\citep{qiInferenceTimeEnhancementGenerative2025,hamelFlowMPCImprovingFlow2026}, but subsequent updates can drift from the learned prior. Gradient guidance steers diffusion or flow updates using differentiable objectives~\citep{carvalhoMotionPlanningDiffusion2025,fengGuidanceFlowMatching2025}. Projection methods enforce constraints on intermediate samples or their estimated clean outputs~\citep{christopherConstrainedSynthesis2024,yuanPhysDiff2023,watanabeProjFlow2026}, requiring feasibility evaluation before generation is complete. Source-noise optimization evaluates objectives on fully generated motions: gradient-based updates differentiate through both the generator and the objective~\citep{ben-hamuDFlowDifferentiatingFlows2024,karunratanakulOptimizingDiffusionNoise2024}, while sampling-based updates use only objective evaluations~\citep{stolzlePhysibleMDMEnforcingInferenceTime2026,hegdeLeWAM2026}. We use sampling-based source optimization with radial quantile bands to restrict the search to typical source-noise radii.

\paragraph{World models for task progress evaluation.}
World models support action selection and refinement by predicting the task outcomes used to evaluate candidate actions~\citep{ebertVisualForesight2018,suhSurprisingEffectivenessLinear2021}. Distances between predicted and goal representations guide planning in learned visual spaces~\citep{zhouDINOWM2025,assranVJEPA22025,hegdeLeWAM2026}. Generative policy proposals can also be ranked or refined using predicted rewards and task objectives~\citep{qiInferenceTimeEnhancementGenerative2025,duDynaGuideSteeringDiffusion2025,hamelFlowMPCImprovingFlow2026}, or guided by learned progress estimates~\citep{yanProgressVLAProgressGuidedDiffusion2026}. We use physics-based world rollouts to evaluate task evolution under tracking and contact dynamics.

\paragraph{Task progress measures.}
Quantifying this progress is difficult when tasks involve intermediate milestones, regressions, and recovery. Beyond geometric task potentials, visual and language-conditioned representations provide learned goal proximity and reward signals~\citep{maVIPUniversalVisual2022,maLIVLanguageImageRepresentations2023}. Reward models learn completion and partial-success assessments from robot videos, incorporating failures and trajectory comparisons~\citep{leeRoboReward2026,liangRobometer2026}. Finer assessments model task stages and progress within them~\citep{chenSARMStageAwareReward2025,chenSARM22026}, or process-level advancement, regression, and recovery~\citep{maoARMAdvantageReward2026,tanRoboDopamineGeneralProcess2025,xuRoboDopamine22026}. Reward machines encode sequential milestones explicitly~\citep{camachoRewardMachinesVisionBased2021,baertRewardMachineInference2025}. We express task progress through a scalar measure of deviation from desired object relationships, combining subtask measures to account for sequential goals.

\paragraph{Safety and convergence of robot motion.}
One line of work incorporates Lyapunov or contraction conditions into training objectives or model architectures to obtain stable robot motion dynamics~\citep{perez-dattariStableMotionPrimitives2023,beik-mohammadiNeuralContractiveDynamical2024,perez-dattariLetDynamicsFlow2026a}. Another establishes closed-loop stability through online predictive optimization, using predictive control to select inputs or predictive filters to correct proposals from an existing policy~\citep{mayneConstrainedModelPredictive2000,didierPredictiveStabilityFilters2024}. Stability conditions also differ in how they treat transients. Finite-step Lyapunov conditions allow intermediate increases by requiring net decrease over several steps~\citep{norooziControlDiscretetimeNonlinear2019}. To also regulate the transient error, prescribed-performance and funnel control specify admissible envelopes~\citep{ilchmannTrackingPrescribedTransient2002,bergerFunnelControlSurvey2023a}, which can be incorporated into predictive control~\citep{grandiaNonlinearModelPredictive2020,bergerFunnelModelPredictive2022}. We apply these principles to progress of the manipulated objects, combining bounded transient regression with terminal contraction of a task potential over generated action chunks. Convergence remains conditional on finding satisfying chunks and agreement between evaluated and executed dynamics. Recursive feasibility is not established.

\paragraph{Continuous-time verification from sampled rollouts.}
Extending discrete trajectory checks to continuous-time satisfaction requires regularity assumptions and margins~\citep{fainekosRobustnessTemporalLogic2009}. Lipschitz optimization bounds function values between samples and uses local rate estimates to reduce conservatism~\citep{leraAccelerationUnivariateGlobal2013}. Our verification procedure uses bounds on how quickly the task-funnel residual can change to rule out violations between consecutive rollout samples. With valid upper bounds on this rate of change, passing the check guarantees that the task error stays within the prescribed envelope throughout the chunk. In our experiments, we estimate these rates from the sampled rollouts, and call the resulting check estimated verification.

%% file: sections/appendix/manifoldflow.tex
\section{Flow-Matching Priors and Pose-Valued Waypoints}
\label{sec:appendix_lie_group_preliminaries}
\paragraph{Conditional flow-matching training.}
The flow map $\psi_\theta$ of the learned vector field $v_\theta(z,\tau\mid o)$ transports a source distribution $p_0$ to the conditional data distribution, $(\psi_\theta(1\mid\cdot,o))_\#p_0\approx p_{\mathrm{data}}(\cdot\mid o)$, where the pushforward $f_\#p$ denotes the distribution of $f(z)$ for $z\sim p$~\citep{lipmanFlowMatchingGenerative2023,lipmanFlowMatchingGuide2024}. For Euclidean actions, $v_\theta$ regresses the velocity of the path $z_\tau=(1-\tau)z_0+\tau w$, with loss $\mathbb{E}\,\|v_\theta(z_\tau,\tau\mid o)-(w-z_0)\|_2^2$ over $(o,w)\sim\mathcal D$, $z_0\sim p_0$ and $\tau\sim\mathcal U[0,1]$. For pose-valued actions we use \ac{RMFM}~\citep{chenFlowMatchingGeneral2024,braunRiemannianFlowMatching2024} with the data-centered wrapped-normal rotational source described below. In both cases a seed $z_0$ is decoded into an action chunk by integrating $d\psi_\theta/d\tau=v_\theta(\psi_\theta,\tau\mid o)$ from $\psi_\theta(0\mid z_0,o)=z_0$ to $\tau=1$.

\paragraph{Pose notation.}
We use standard Lie-group notation for $\mathrm{SO}(3)$ and $\mathrm{SE}(3)$: exponential and logarithm maps, and the hat and vee operators identifying a body twist $\xi\in\mathbb R^6$ with its Lie-algebra matrix $\widehat\xi$~\citep{solaMicroLieTheory2021,lynchModernRoboticsMechanics2017}.

\paragraph{Interface, carrier, and velocity representations.}
The world-facing waypoint representation is eight values: position, a unit quaternion, and a gripper variable. The Euclidean (\ac{OT}) policy flows these eight coordinates directly and predicts eight coordinate velocities. Before execution, its quaternion block is normalized and converted to a rotation. Quaternions, however, are a discontinuous representation of rotations for a regression target~\citep{zhouContinuityRotationRepresentations2020}, so the Riemannian (\ac{RMFM}) policy flows a continuous carrier instead: the nine rotation-matrix entries, three translation coordinates, and the gripper channel. This 13-dimensional carrier represents a state in $\mathrm{SE}(3)\times\mathbb R$, and the group operations act on it directly as matrix products, which keeps the pose kernels of the flow simple. The model reads the 13 carrier channels and outputs seven tangent channels, a body twist $(v,\omega)$ and a gripper rate. Quaternions appear only in the world-facing interface. The \ac{RMFM} flow evolves the 13-dimensional carrier, and the refinement search operates on chart coordinates of its source (Appendix~\ref{sec:appendix_cmaes_refinement}). Rotation observations use the distinct six-dimensional representation of \citet{zhouContinuityRotationRepresentations2020}.

\paragraph{Coupled pose interpolation and integration.}
Write $G\in\mathrm{SE}(3)$ for the pose and $r\in\mathbb R$ for the gripper variable. For source and target poses, let $\xi=\log(G_0^{-1}G_1)^\vee$. The training path is
\begin{equation}
    G_\tau=G_0\exp(\tau\widehat\xi),\qquad
    r_\tau=(1-\tau)r_0+\tau r_1,
\end{equation}
a screw motion with constant body velocity $\xi$, which couples translation and rotation through the $\mathrm{SE}(3)$ exponential, while the gripper moves at constant rate $r_1-r_0$. Applied waypoint-wise, this gives the flow-matching target~\citep{braunRiemannianFlowMatching2024,chenFlowMatchingGeneral2024}. Sampling uses 10 Lie--Euler steps,
\begin{equation}
    G\leftarrow G\exp\bigl(\Delta\tau\,\widehat v_\theta^{\mathrm{pose}}\bigr),\qquad
    r\leftarrow r+\Delta\tau\,v_\theta^{\mathrm{grip}},
\end{equation}
so the rotation stays on $\mathrm{SO}(3)$ without re-orthonormalization. The final carrier is converted back to the eight-value interface for execution. %

\paragraph{Data-centered rotational source.}
The source samples translation and gripper coordinates from standard Gaussians in normalized units, and rotation from a wrapped normal centered on the data,
\begin{equation}
    R_0=\bar R\exp(\sigma_{\mathrm{rot}}[\epsilon_R]_\times),
    \qquad \epsilon_R\sim\mathcal{N}(0,I_3),\qquad
    \sigma_{\mathrm{rot}}=0.5\ \mathrm{rad},
    \label{eq:rotation_prior}
\end{equation}
which is then converted to the 13-dimensional carrier. Here $[\cdot]_\times$ is the skew-symmetric matrix representation. The center $\bar R$ is computed from demonstration rotations: its unit quaternion is a dominant eigenvector of $\sum_j q_jq_j^\top$, the sign-invariant chordal quaternion average~\citep{markleyAveragingQuaternions2007}, computed by power iteration. The prior center is saved with the training statistics. Centering the source on the demonstrated orientations reduces the frequency of source--target pairs near the cut locus of the rotational logarithm, where a relative angle near $\pi$ makes the interpolation target bimodal. For concentrated demonstrated orientations, almost all pairs lie well inside the cut locus. A Haar-uniform source~\citep{chirikjianStochasticModelsInformation2012a} places a fixed fraction of pairs near the cut locus for any data.

Training and inference use the same Gaussian-to-pose source map. The source samples translation and rotation as separate factors, and the flow couples them through the learned pose velocity and its integration with the $\mathrm{SE}(3)$ exponential. The map $\epsilon_R\mapsto R_0$ is injective for rotation angles below $\pi$, where the principal logarithm recovers $\epsilon_R$~\citep{hartleyRotationAveraging2013}.

%% file: sections/appendix/phasedlyapunov.tex
\section{Task Progress Measures and Task Potentials}
\label{app:phased_lyapunov}
\label{sec:potential_construction}
\label{sec:instantiation}
This appendix constructs the task potential $V$ of \Cref{sec:task_refinement}, a task-level Lyapunov candidate, so that $V$ does not increase while the task progresses as intended: the subtasks are completed in order, each by first meeting a prerequisite, such as lifting a cube, and then completing it, such as placing the cube on its support. Executed motions can leave this progression, through robot-only motions that leave $V$ constant, contacts that push, drop or topple objects, and detours away from the goal, so $V$ remains a candidate, and the funnel imposes its decrease on each executed chunk. All constructions are stated on a bounded operating domain $\Omega$ of task states, set by the task definition and the training data: the physical limits of the scene, such as the table extent, together with the range of task states covered by the demonstrations, with a margin.

This appendix has two parts. The first is a construction procedure that can be reused for other tasks with identifiable subtasks: the requirements that a potential should meet, a way to blend the costs of one subtask into a smooth residual without ridges (Appendix~\ref{app:smooth_residual}), and a way to compose the residuals of successive subtasks so that completing one never increases the potential (Appendix~\ref{app:subtask_composition}, \Cref{prop:phased_composition}). The second part consists of the potentials of our three tasks (Appendices~\ref{app:pusht_candidate}--\ref{app:nut_candidate}). Their coordinates, costs, gates and constants are designed by hand for each task. They are instances of the procedure and make no claim of generality.

\paragraph{Requirements.}
The method places four requirements on the potential over $\Omega$.
\begin{itemize}
    \item \emph{Zero set.} $V$ is nonnegative and vanishes exactly on the completed task, so that the terminal contraction of \Cref{thm:receding_horizon_contraction} drives execution toward completion.
    \item \emph{Regularity.} $V$ is Lipschitz along rollouts, as the verification bound requires (\Cref{lem:residual_regularity}). We construct it $C^1$, which gives a Lipschitz constant on the bounded domain $\Omega$ and keeps kinks out of the trace that the funnel evaluates.
    \item \emph{No ridges.} A motion that advances the current subtask never increases $V$, so the funnel does not penalize progress on that subtask.
    \item \emph{Ordering.} Completing a subtask never increases $V$, and $V$ does not depend on a later subtask until the earlier ones approach completion.
\end{itemize}
For a single-stage task, a weighted distance to the desired state meets these requirements, as in PushT (Appendix~\ref{app:pusht_candidate}). For tasks with identifiable subtasks and prerequisites, we give a construction procedure. A subtask residual $\phi$ is a nonnegative scalar function of the task state that vanishes exactly when its subtask is complete. The procedure blends the costs of a subtask into a smooth residual (Appendix~\ref{app:smooth_residual}) and composes the residuals of successive subtasks, so that activating the next subtask preserves the desired ordering of task states (Appendix~\ref{app:subtask_composition}). In stacking, for example, the first subtask places the red cube on the green one and the second places the blue cube on the red one. The prerequisite of each placement is lifting the cube from the table, and its completion is the cube centred and resting on its support. Each task supplies its own geometric subtask costs and completion gates, with a few constants calibrated on demonstrations. We apply the full procedure to stacking (Appendix~\ref{app:phased_calibration}) and to two-nut assembly (Appendix~\ref{app:nut_candidate}). Learning the potential from demonstrated progress, in place of the hand-designed instances, is left to future work.

\subsection{Designing a Smooth Subtask Residual}
\label{app:smooth_residual}
\paragraph{Choosing coordinates and regimes.}
Select coordinates $\xi$ that describe the prerequisite and completion of a subtask, and define nonnegative costs $U_0(\xi)$ and $Q(\xi)$ for these two regimes. A prerequisite can represent clearance before transport or alignment before insertion, and completion specifies the desired task relation. From these two costs we build the subtask residual $\phi(\xi)$ defined above. It should vanish at the intended geometric goal and decrease along the intermediate motions that lead to it. In stacking, $\xi=(d,e)$ collects the horizontal distance and the vertical gap between a cube and its support, $U_0$ is smallest when the cube is carried at a clearance above its support, and $Q$ when it is centred over the support and lowered onto it (Appendix~\ref{app:phased_calibration}).

\paragraph{Connecting regimes smoothly.}
With an offset $\bar D\ge0$, let $U(\xi)=U_0(\xi)+\bar D$, and blend the two regimes through a gate $\ell(\xi)\in[0,1]$,
\begin{equation}
    \phi(\xi)=[1-\ell(\xi)]U(\xi)+\ell(\xi)Q(\xi).
    \label{eq:stage_residual_blend}
\end{equation}
The gate should equal $0$ before the prerequisite is started and $1$ once it is met, so that each regime is exact away from the transition. It should also be monotone in the prerequisite's progress, and $C^1$ with zero slope at both ends of the transition, so that the blend has no kinks. The blend is then nonnegative, and if $U>0$ and $Q$ vanishes exactly at the goal, where $\ell=1$, $\phi$ has the desired zero set. When a scalar coordinate $a(\xi)$ increases with prerequisite progress, the clamped cubic step is the lowest-degree polynomial with these properties,
\begin{equation}
    \ell=s\!\left(\frac{a-a_{\mathrm{lo}}}{a_{\mathrm{hi}}-a_{\mathrm{lo}}}\right),\qquad
    s(u)=\begin{cases}0,&u\le0,\\3u^2-2u^3,&0<u<1,\\1,&u\ge1,\end{cases}
    \label{eq:clamped_smoothstep}
\end{equation}
where $[a_{\mathrm{lo}},a_{\mathrm{hi}}]$ is the band over which the emphasis moves from one regime to the other.

\paragraph{Avoiding ridges at activation.}
A smooth gate avoids discontinuities in the measurement, but it can still create a cost ridge: while the gate rises, the residual can increase while the robot performs the prerequisite motion. A motion that improves the prerequisite should not increase $\phi$. For a coordinate $\xi_a$ that increases with the desired prerequisite motion,
\begin{equation}
    \partial_{\xi_a}\phi=(1-\ell)\partial_{\xi_a}U
        +\ell\partial_{\xi_a}Q+(\partial_{\xi_a}\ell)(Q-U).
    \label{eq:micro_derivative}
\end{equation}
Suppose that both regime costs decrease along this motion, $\partial_{\xi_a}U_0\le0$ and $\partial_{\xi_a}Q\le0$, and that the gate rises along it, $\partial_{\xi_a}\ell\ge0$, on the transition band $\Omega_{\mathrm{tr}}\subseteq\Omega$ where $0<\ell<1$. The first two terms are then nonpositive, and the last one is nonpositive whenever $Q\le U$. A sufficient offset is therefore
\begin{equation}
    \bar D\ge\max\left\{0,\sup_{\xi\in\Omega_{\mathrm{tr}}}
        [Q(\xi)-U_0(\xi)]\right\},
    \label{eq:micro_offset}
\end{equation}
which gives $\partial_{\xi_a}\phi\le0$ on the band, the no-ridge requirement. The offset is computed once from $\Omega$ and involves no tuning on trajectories.

\subsection{Composing Subtasks}
\label{app:subtask_composition}
For two successive subtasks with residuals $\phi_1$ and $\phi_2$, choose a $C^1$ completion gate $\chi:[0,\infty)\to[0,1]$ with $\chi'\le0$ and $\chi(0)=1$, which exposes the second residual as the first subtask is completed. Choose a reserve $\bar B\ge\sup_{y\in\Omega}\phi_2(y)$ and define
\begin{equation}
    V(y)=\phi_1(y)+\bigl[1-\chi\bigl(\phi_1^{\mathrm g}(y)\bigr)\bigr]\bar B+\chi\bigl(\phi_1^{\mathrm g}(y)\bigr)\,\phi_2(y).
    \label{eq:phased_potential}
\end{equation}
While the first subtask is incomplete, $\chi=0$ and the second subtask enters only through the constant $\bar B$, so $V$ does not depend on $\phi_2$: in stacking, moving the blue cube before the red one is placed leaves $V$ unchanged, so $V$ decreases only through the current subtask. As the first subtask is completed, $\bar B$ is exchanged for the actual second residual, which is no larger on $\Omega$, so the exchange cannot itself increase $V$.

The gate reads a gate residual $\phi_1^{\mathrm g}$: a nonnegative $C^1$ residual of the first subtask that vanishes on the same set as $\phi_1$. The simplest choice is $\phi_1^{\mathrm g}=\phi_1$. A separate gate residual is useful when $\phi_1$ contains terms that keep refining an already completed subtask, such as fine centring: the gate can then hand over to the next subtask as soon as the first is complete, without waiting for that refinement. One suitable completion gate is
\begin{equation}
    \chi(\phi)=s\!\left(\frac{b_{\mathrm{hi}}-\phi}{b_{\mathrm{hi}}-b_{\mathrm{lo}}}\right),
    \qquad 0<b_{\mathrm{lo}}<b_{\mathrm{hi}},
    \label{eq:completion_gate}
\end{equation}
with $\chi=1$ for $\phi\le b_{\mathrm{lo}}$ and $\chi=0$ for $\phi\ge b_{\mathrm{hi}}$: $[b_{\mathrm{lo}},b_{\mathrm{hi}}]$ is the band of first-stage gate residual over which the second subtask is activated.

\begin{proposition}[Smooth sequential composition]
    \label{prop:phased_composition}
    Let $\phi_1$, $\phi_2$, and $\phi_1^{\mathrm g}$ be nonnegative $C^1$ residuals with $\{\phi_1=0\}=\{\phi_1^{\mathrm g}=0\}$, and let $\chi$ and $\bar B$ be as above. Then $V$ is nonnegative and $C^1$ on $\Omega$, with $V(y)=0$ exactly when $\phi_1(y)=\phi_2(y)=0$. In residual coordinates,
    \begin{equation}
        \frac{\partial V}{\partial \phi_1}=1,
        \qquad \frac{\partial V}{\partial \phi_1^{\mathrm g}}=\chi'(\phi_1^{\mathrm g})\,(\phi_2-\bar B)\ge0,
        \qquad \frac{\partial V}{\partial \phi_2}=\chi(\phi_1^{\mathrm g})\in[0,1].
        \label{eq:phase_priority_derivatives}
    \end{equation}
\end{proposition}
\begin{proof}
Every term in \eqref{eq:phased_potential} is nonnegative. If $V=0$, then $\phi_1=0$, hence $\phi_1^{\mathrm g}=0$, $\chi=1$, and $\phi_2=0$; the converse follows by substitution. Products and compositions preserve $C^1$ regularity. Differentiation gives \eqref{eq:phase_priority_derivatives}, whose signs follow from $\chi'\le0$ and $\phi_2\le\bar B$.
\end{proof}
Reducing either first-stage residual at a fixed second residual therefore never increases $V$, including during activation, and the first subtask has full weight while the second has weight $\chi\le1$. Together with the zero set and the $C^1$ regularity of the proposition, this meets the ordering requirement. Along a trajectory, all residuals change together,
\[
    \dot V=\dot\phi_1+\chi'(\phi_1^{\mathrm g})(\phi_2-\bar B)\,\dot \phi_1^{\mathrm g}+\chi(\phi_1^{\mathrm g})\,\dot\phi_2,
\]
so the construction does not fix the sign of $\dot V$, and the funnel evaluates how $V$ evolves there. Longer sequences follow by applying the composition recursively.

\subsection{Parameter Selection and Calibration}
\label{app:phased_calibration_procedure}
The constants of a potential come from three sources, and each task uses only the groups its potential needs. Tables~\ref{tab:phased_parameters} and~\ref{tab:nut_parameters} list every constant of the stacking and two-nut potentials with its source. The PushT pose error has a single angular weight and no stage constants, and its weight equates an angular and a positional tolerance (Appendix~\ref{app:pusht_candidate}).

\paragraph{Geometry and physical tolerances.}
Error coordinates come from the task's object relations and are scaled by object dimensions or acceptable completion errors. In stacking, for instance, the horizontal and vertical error scales are the half-height of the supporting cube, and the carry clearance is one cube height plus one centimetre. Relative weights then set how the normalized errors are combined. We set these weights by hand and use the demonstrations to check that the tolerances accommodate the contact noise of successful configurations.

\paragraph{Analytic composition constants.}
Offsets and reserves are computed from bounds on the operating domain $\Omega$. The lift offset $\bar D$ is the bound \eqref{eq:micro_offset} evaluated on $\Omega$. The stage reserve $\bar B$ is the largest value of the second-stage residual on $\Omega$, which keeps $V$ from rising at the hand-over in \eqref{eq:phased_potential}. The composition properties hold on $\Omega$, and the potential remains defined and finite outside it.

\paragraph{Demonstration-informed thresholds.}
The remaining hand-over thresholds and scales come from training states. In stacking, the completion band starts ten percent above the largest gate residual among training placements that are geometrically complete, and the alignment scale is close to the typical seat offset of demonstrated placements. Placements outside the intended completion geometry are not used to widen the band. All potential parameters are fixed before the funnel's training labels are generated.

\subsection{PushT}
\label{app:pusht_candidate}
For block position $p_T$, target position $p_G$, and orientations $\theta_T,\theta_G$, use
\begin{equation}
    V_{\mathrm{PushT}}(y)=\|p_T-p_G\|_2^2+
    \lambda_\theta\operatorname{wrap}(\theta_T-\theta_G)^2,
    \label{eq:pusht_potential}
\end{equation}
where $\operatorname{wrap}(u)=\operatorname{atan2}(\sin u,\cos u)\in(-\pi,\pi]$, positions are in the pixel units of the $512\times512$ workspace, angles are in radians, and $\lambda_\theta=10^4$, so that an angular error of $0.1$\,rad weighs as much as a position error of $10$ pixels. For $\lambda_\theta>0$, the potential is nonnegative and vanishes exactly at the target pose. The squared wrapped-angle term is locally Lipschitz, including at the angular cut, though not differentiable there.

\subsection{Three-Cube Stacking}
\label{app:phased_calibration}
\begin{table}[t]
    \centering
    \caption{Phased potential parameters for three-cube stacking, grouped by source, for red on green (R on G) and blue on red (B on R). Lengths are in meters; costs and weights are dimensionless.}
    \label{tab:phased_parameters}
    \footnotesize
    \setlength{\tabcolsep}{3pt}
    \begin{tabular}{@{}llccl@{}}
        \toprule
        Symbol & Meaning & R on G & B on R & Rule\\
        \midrule
        \multicolumn{5}{@{}l}{\textit{Geometry and physical tolerances}}\\
        $h_X,h_Y$ & half-heights & $0.020/0.025$ & $0.020/0.020$ & model geometry\\
        $\tau_{xy}=\tau_z$ & error scales & $0.025$ & $0.020$ & $h_Y$\\
        $\lambda_z$ & vertical weight & $1$ & $1$ & design\\
        $d_{\mathrm{lo}}/d_{\mathrm{hi}}$ & lowering band & $0.010/0.045$ & $0.010/0.040$ & $d_{\mathrm{lo}}$ design; $d_{\mathrm{hi}}=h_X+h_Y$\\
        $c$ & carry clearance & $0.050$ & $0.050$ & $2h_X+0.01$\\
        $a_{\mathrm{lo}}/a_{\mathrm{hi}}$ & lift band & $-0.027/-0.012$ & $-0.027/-0.012$ & $\le12$\,mm above support, $15$\,mm band\\
        $\lambda_{\mathrm{pre}}$ & pre-positioning & $0.01$ & $0.01$ & design\\
        $\lambda_{\mathrm{align}}$ & alignment & $10$ & $10$ & above jitter, below a stage drop\\
        $\phi^{\mathrm g}_{RG}$ bands & gate residual & \multicolumn{2}{c}{$d_{\mathrm{lo}}=0.025$, $a_{\mathrm{lo}}/a_{\mathrm{hi}}=-0.045/-0.030$} & wider, no alignment basin\\
        \midrule
        \multicolumn{5}{@{}l}{\textit{Analytic composition constants (on $\Omega$)}}\\
        $\bar D$ & lift offset & $559.65$ & $868.83$ & \eqref{eq:unlifted_offset}\\
        $\bar D^{\mathrm g}$ & gate-residual offset & \multicolumn{2}{c}{$549.65$} & \eqref{eq:unlifted_offset}, $\lambda_{\mathrm{align}}=0$\\
        $\bar B$ & second-stage reserve & \multicolumn{2}{c}{$1139.36$} & $\ge\sup_\Omega \phi_{BR}$\\
        $\delta_z$ & no-ridge margin & \multicolumn{2}{c}{$0.0045$} & lowering-band condition\\
        \midrule
        \multicolumn{5}{@{}l}{\textit{Demonstration-informed}}\\
        $\Omega$ & operating domain & \multicolumn{2}{c}{$p_{X,xy}\in[-0.2,0.2]^2$ at reset; $d\le0.59$} & MimicGen reset range + margin\\
        $\sigma_{\mathrm{align}}$ & alignment scale & $0.010$ & $0.010$ & typical demo seat offset\\
        $b_{\mathrm{lo}}/b_{\mathrm{hi}}$ & completion-gate band & \multicolumn{2}{c}{$1.093/2.186$} & $1.1\times$ largest completed; $b_{\mathrm{hi}}=2b_{\mathrm{lo}}$\\
        \bottomrule
    \end{tabular}
\end{table}

We apply the procedure to three-cube stacking, which places the red cube $R$ on the green cube $G$ and then the blue cube $B$ on the red one. Each placement is a subtask with a lift prerequisite, and the two placements are composed in this order. Let $p_X=(p_{X,xy},z_X)$ and $h_X$ be cube $X$'s center and half-height. The task map extracts $y=(p_R,p_G,p_B)$. For each placement of $X$ on $Y$,
\begin{equation}
    d_{XY}=\|p_{X,xy}-p_{Y,xy}\|_2,
    \qquad e_{XY}=z_X-z_Y-(h_X+h_Y),
    \label{eq:placement_geometry}
\end{equation}
where $d_{XY}$ is the horizontal distance between the two centers and $e_{XY}$ the vertical gap between the bottom face of $X$ and the top face of $Y$, assuming upright faces.

\paragraph{Placement residual.}
Dropping the placement subscripts, the lift gate $\ell(e)=s((e-a_{\mathrm{lo}})/(a_{\mathrm{hi}}-a_{\mathrm{lo}}))$ measures how far the cube has been raised relative to its support, and $A(d)=1-s((d-d_{\mathrm{lo}})/(d_{\mathrm{hi}}-d_{\mathrm{lo}}))$ measures horizontal alignment. The vertical target $z_\star(d)=c[1-A(d)]$ equals the carry clearance $c$ far from alignment and approaches zero near it, a curved valley that rewards carrying before lowering. With error scales $\tau_{xy},\tau_z$, a vertical weight $\lambda_z$, and a small pre-positioning $\lambda_{\mathrm{pre}}\in[0,1)$,
\begin{align}
    E_{xy}(d)&=d^2/\tau_{xy}^2,& U_0(d,e)&=\lambda_z(e-c)^2/\tau_z^2+\lambda_{\mathrm{pre}}E_{xy}(d),\notag\\
    Q(d,e)&=E_{xy}(d)+\lambda_z[e-z_\star(d)]^2/\tau_z^2,
    &U&=U_0+\bar D.
    \label{eq:placement_regimes}
\end{align}
Here $E_{xy}$ is the horizontal misalignment in units of the error scale. The lift cost $U_0$ penalizes the deviation of the gap from the carry clearance $c$, and its small horizontal term moves the cube toward its support while it is lifted. The placement cost $Q$ penalizes the horizontal misalignment and the deviation of the gap from the valley $z_\star(d)$, so that the cube is lowered as it aligns. The error scales make both costs dimensionless, and $\lambda_z$ weighs vertical against horizontal error.
Once the cube is lifted, an alignment basin $f_{\mathrm{align}}(d)=d^2/(d^2+\sigma_{\mathrm{align}}^2)$ ranks completed placements by centring, and the residual is
\begin{equation}
    \phi(d,e)=[1-\ell(e)]U(d,e)+\ell(e)\bigl[Q(d,e)+\lambda_{\mathrm{align}}f_{\mathrm{align}}(d)\bigr].
    \label{eq:placement_residual}
\end{equation}
In the transition band, $e<0$ and $0\le z_\star\le c$ give $Q+\lambda_{\mathrm{align}}f_{\mathrm{align}}-U_0\le(1-\lambda_{\mathrm{pre}})E_{xy}+\lambda_{\mathrm{align}}$, so the rule in \eqref{eq:micro_offset} is met by
\begin{equation}
    \bar D=(1-\lambda_{\mathrm{pre}})\sup_{(d,e)\in\Omega}E_{xy}(d)+\lambda_{\mathrm{align}}.
    \label{eq:unlifted_offset}
\end{equation}
The gate residual $\phi^{\mathrm g}_{RG}$ has the same form without the alignment basin and with wider bands, so that the hand-over to the second placement does not wait for fine centering.

\paragraph{Potential.}
With the two placement residuals,
\begin{equation}
    V_{\mathrm{stack}}(y)=\underbrace{\phi_{RG}(y)}_{\text{\scriptsize first placement}}
        +\underbrace{[1-\chi(\phi^{\mathrm g}_{RG}(y))]\bar B}_{\substack{\text{\scriptsize reserve while the second}\\\text{\scriptsize placement is inactive}}}
        +\underbrace{\chi(\phi^{\mathrm g}_{RG}(y))\phi_{BR}(y)}_{\substack{\text{\scriptsize second placement,}\\\text{\scriptsize once activated}}},\label{eq:stack_potential}
\end{equation}
which by \Cref{prop:phased_composition} is $C^1$ on $\Omega$ and vanishes exactly at the geometric stack $d_{RG}=e_{RG}=d_{BR}=e_{BR}=0$: each residual vanishes only where $\ell=1$, $Q=0$, and $f_{\mathrm{align}}=0$, that is, at $d=e=0$. The potential is invariant to common translations and measures geometric placement, not orientation, release, or stable support. Completion is judged by the task's success predicate. The main experiments use this potential, which builds both placements from the residual $\phi$ of \eqref{eq:placement_residual}. Appendix~\ref{app:potential_ablation} compares it with a potential that builds both placements from the gate residual $\phi^{\mathrm g}$. Table~\ref{tab:phased_parameters} lists the constants, grouped by their source as in Appendix~\ref{app:phased_calibration_procedure}.

\subsection{Two-Nut Assembly}
\label{app:nut_candidate}
The task threads the square nut onto its peg, then the round nut onto the other (Appendix~\ref{app:nut_results}). The potential reads the two nuts only, each relative to its own peg. For a nut, let $p_\perp\in\mathbb R^2$ be the horizontal offset of its center from the peg axis, $R\in SO(3)$ its orientation and $z$ its height. The task map extracts $y=(p_\perp,R,z)$ for both nuts.

\paragraph{Alignment energy.}
With the native lateral tolerance $\tau_r$ and $u=\|p_\perp\|/\tau_r$, the lateral error is quadratic inside the native tolerance and linear outside it, $h(u)=u^2$ for $u\le1$ and $h(u)=2u-1$ otherwise, so that it stays $C^1$ and grows slowly far from the peg. Tilt is measured by $e_{\mathrm{tilt}}=1-n_z^2$ with $n=Re_z$, the nut's axis. The square hole also needs yaw alignment up to its fourfold symmetry, measured with $a=Re_x$ by
\begin{equation}
    e_{\mathrm{yaw}}=\frac{8a_x^2a_y^2}{(a_x^2+a_y^2)^2+\varepsilon},
    \qquad
    q=h(u)+\lambda_{\mathrm{tilt}}\,e_{\mathrm{tilt}}+\lambda_{\mathrm{yaw}}\,e_{\mathrm{yaw}},
    \label{eq:nut_alignment}
\end{equation}
which vanishes when an edge of the nut is parallel to an axis of the peg. The round nut has $\lambda_{\mathrm{yaw}}=0$.

\paragraph{Nut residual.}
An alignment valve $v(q)=s((q_{\mathrm{hi}}-q)/(q_{\mathrm{hi}}-q_{\mathrm{lo}}))$, with $s$ the clamped cubic step of \eqref{eq:clamped_smoothstep}, opens as the nut aligns. The height cost of an aligned nut, $H(z)=\bar H\,s(|z-z_{\mathrm{seat}}|/(z_{\mathrm{app}}-z_{\mathrm{seat}}))\le\bar H$, vanishes at the seated height $z_{\mathrm{seat}}$ and saturates at the approach height $z_{\mathrm{app}}$ above the peg. A nut counts as resting on the table when it is laterally misaligned and not lifted: with the lift indicator $\mu(z)=s((z-z_{\mathrm{lift}})/h_{\mathrm{lift}})$ and the lateral valve $v_{\mathrm{lat}}=s((q_{\mathrm{hi}}-h(u))/(q_{\mathrm{hi}}-q_{\mathrm{lo}}))$, its resting indicator is $\omega=(1-v_{\mathrm{lat}})(1-\mu(z))$. The residual is
\begin{equation}
    \phi=\underbrace{(1-\omega)\,q}_{\substack{\text{\scriptsize alignment (lifted}\\\text{\scriptsize or over the peg)}}}
        +\underbrace{\omega\bigl[m(q)+L_{\mathrm{lift}}\bigr]}_{\substack{\text{\scriptsize resting on}\\\text{\scriptsize the table}}}
        +\underbrace{v(q)\,H(z)}_{\substack{\text{\scriptsize height,}\\\text{\scriptsize once aligned}}}
        +\underbrace{\bigl[1-v(q)\bigr]\bar H}_{\substack{\text{\scriptsize height reserve}\\\text{\scriptsize while misaligned}}},
    \label{eq:nut_residual}
\end{equation}
where $m$ is a nondecreasing $C^1$ smoothing of $\max(q,q_{\mathrm{rest}})$ with a quadratic corner of width $\delta_{\mathrm{rest}}$, and $q_{\mathrm{rest}}$ bounds the alignment energy on $\Omega$.

\paragraph{Properties.}
On $\Omega$, aligning a nut at fixed height never raises $\phi$: for a lifted nut or one over its peg, $\omega=0$ and $\partial\phi/\partial q=1+v'(q)[H(z)-\bar H]\ge1$, since $v'\le0$ and $H\le\bar H$, and for a resting nut $m$ is nondecreasing. A misaligned nut has height cost $\bar H$ at any height and an aligned one has $H(z)$, so lowering an aligned nut to its seat decreases $\phi$. A nut resting away from its peg ($\omega=1$) costs $q_{\mathrm{rest}}+L_{\mathrm{lift}}+\bar H$ wherever it lies: pushing or turning it on the table leaves $\phi$ unchanged, lifting it decreases $\phi$, and dropping a carried nut increases $\phi$. Since $z_{\mathrm{lift}}$ is the highest spawn height, the reset's drop of the nuts onto the table does not count as a lift.

\paragraph{Potential.}
With the square-nut residual $\phi_{\mathrm{sq}}$ as first stage, also used as gate residual, and the round-nut residual $\phi_{\mathrm{rd}}$ as second,
\begin{equation}
    V_{\mathrm{nut}}(y)=\underbrace{\phi_{\mathrm{sq}}(y)}_{\text{\scriptsize first placement}}
        +\underbrace{\bigl[1-\chi(\phi_{\mathrm{sq}}(y))\bigr]\bar B}_{\substack{\text{\scriptsize reserve while the second}\\\text{\scriptsize placement is inactive}}}
        +\underbrace{\chi(\phi_{\mathrm{sq}}(y))\,\phi_{\mathrm{rd}}(y)}_{\substack{\text{\scriptsize second placement,}\\\text{\scriptsize once activated}}},
    \label{eq:nut_potential}
\end{equation}
with the completion gate of \eqref{eq:completion_gate} and $\bar B\ge\sup_\Omega \phi_{\mathrm{rd}}$, so that by \Cref{prop:phased_composition} $V_{\mathrm{nut}}$ is $C^1$ on $\Omega$ and completing the first placement never increases it. It vanishes only when both nuts are centred on their pegs, level and seated, with the square nut's yaw a multiple of $90^\circ$, a set much smaller than the tolerances of the native success check. Flipped nuts are not excluded. Outside $\Omega$ the potential remains $C^1$ and finite, but its bounds are not claimed there. Table~\ref{tab:nut_parameters} lists the constants, grouped by source as in Appendix~\ref{app:phased_calibration_procedure}.

\begin{table}[t]
    \centering
    \caption{Phased potential parameters for two-nut assembly, grouped by source, for the square and the round nut. Lengths are in meters; costs and weights are dimensionless.}
    \label{tab:nut_parameters}
    \footnotesize
    \setlength{\tabcolsep}{3pt}
    \begin{tabular}{@{}llccl@{}}
        \toprule
        Symbol & Meaning & Square & Round & Rule\\
        \midrule
        \multicolumn{5}{@{}l}{\textit{Geometry and physical tolerances}}\\
        $\tau_r$ & lateral tolerance & $0.03$ & $0.03$ & half-width of the native success box\\
        $z_{\mathrm{seat}}/z_{\mathrm{app}}$ & seated / approach height & $0.83/0.97$ & $0.83/0.97$ & model geometry; peg top plus nut thickness\\
        $q_{\mathrm{lo}}$ & aligned threshold & $0.101$ & $0.163$ & $(\text{seated offset}/\tau_r)^2$\\
        $\lambda_{\mathrm{tilt}}$ & tilt weight & $5.13$ & $3.11$ & $1/\sin^2$ of the threading tilt limit\\
        $\lambda_{\mathrm{yaw}}$ & yaw weight & $0.0127$ & $0$ & $(\text{clearance}/\tau_r)^2/4$; none for the round nut\\
        $q_{\mathrm{hi}}$ & valve closure & $1$ & $1$ & edge of the native box\\
        $\bar H$ & height reserve & $5.04$ & $5.04$ & $\lambda_z/s(\tau_z/(z_{\mathrm{app}}-z_{\mathrm{seat}}))$, $\tau_z=0.04$, $\lambda_z=1$\\
        $L_{\mathrm{lift}}=\delta_{\mathrm{rest}}$ & lift credit, corner width & $5.04$ & $5.04$ & $\bar H$\\
        $h_{\mathrm{lift}}$ & lift band & $0.02$ & $0.02$ & $\tau_z/2$\\
        \midrule
        \multicolumn{5}{@{}l}{\textit{Analytic composition constants (on $\Omega$)}}\\
        $\bar B$ & second-stage reserve & \multicolumn{2}{c}{$46.9$} & $\ge\sup_\Omega \phi_{\mathrm{rd}}$\\
        \midrule
        \multicolumn{5}{@{}l}{\textit{Demonstration-informed}}\\
        $z_{\mathrm{lift}}$ & lift height & $0.89$ & $0.89$ & highest spawn height of the reset\\
        $q_{\mathrm{rest}}$ & alignment bound on $\Omega$ & $28.35$ & $36.78$ & $1.1\times$ largest observed\\
        $b_{\mathrm{lo}}/b_{\mathrm{hi}}$ & completion-gate band & \multicolumn{2}{c}{$0.136/0.271$} & $1.1\times$ largest $\phi_{\mathrm{sq}}$ over flat final seats; $b_{\mathrm{hi}}=2b_{\mathrm{lo}}$\\
        \bottomrule
    \end{tabular}
\end{table}

%% file: sections/appendix/funnels.tex
\section{Performance Funnels}
\label{app:adaptive_funnel}
\label{sec:adaptive_funnel}
\label{app:fixed_funnel}
\begin{figure}[t]
    \centering
    \includegraphics[width=\linewidth]{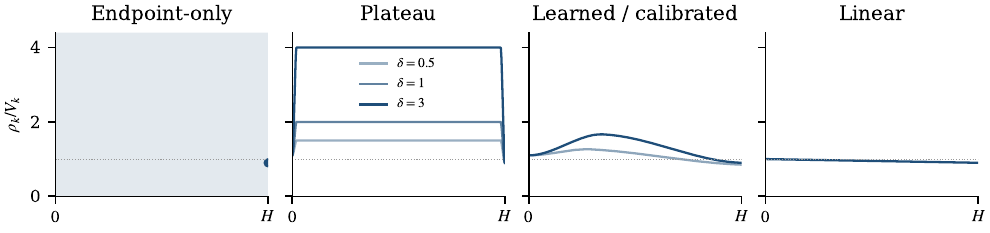}
    \caption{Shapes of the compared funnels, relative to the current value $V_k$ (dotted). Endpoint-only bounds only the last sample, the plateau adds a uniform bound $(1+\delta)V_k$, the linear funnel asks for steady progress, and the calibrated and learned funnels follow \eqref{eq:adaptive_envelope} (two illustrative parameter sets).}
    \label{fig:funnel_shapes}
\end{figure}

The funnel is built for a fixed task potential $V$ (Appendix~\ref{app:phased_lyapunov}) and is frozen at the start of each planning call. The results of \Cref{sec:task_refinement} place four requirements on it.
\begin{itemize}
    \item \emph{Anchoring.} It starts at the current value $V_k$, possibly raised by a start allowance that fades over the chunk and makes the first samples verifiable (Appendix~\ref{app:initial_lift}).
    \item \emph{Terminal contraction.} It ends at the requested value $(1-\gamma_k)V_k$, which drives the convergence of \Cref{thm:receding_horizon_contraction}.
    \item \emph{Bounded transient.} It stays below $(1+d_k)V_k$, which bounds the task error between replanning times.
    \item \emph{Lipschitz in time.} Its Lipschitz constant enters the rate of the residual, as the verification bound requires (\Cref{lem:residual_regularity}).
\end{itemize}
Every funnel we compare meets the terminal requirement, and the funnels differ in how much they constrain the interior of the chunk (Figure~\ref{fig:funnel_shapes}). We compare five types:
\begin{itemize}
    \item \emph{Endpoint-only} constrains the last sample and leaves the interior free.
    \item \emph{Plateau} adds a uniform transient bound $(1+\delta)V_k$ to the endpoint-only funnel.
    \item \emph{Linear} decreases at a constant rate from $V_k$ to the requested value over the chunk.
    \item \emph{Calibrated} funnels use the fitted family of Appendix~\ref{app:funnel_families} with constant parameters, calibrated on the demonstrations over the whole task (\emph{global}) or for each stage of a sequential task (\emph{per stage}).
    \item \emph{Learned} funnels, additive or multiplicative, use the same family and predict its parameters from the observation history at leniency $\alpha$ (Appendix~\ref{app:funnel_fitting}).
\end{itemize}
We also compare two objectives without an envelope, which the same search minimizes in place of the funnel loss. On the samples $t_i$ of the chunk, judged from the end of the first control step as for the funnels, the \emph{$\dot V\le0$} objective is the smooth maximum $\frac1\beta\log\sum_i\exp(\beta\,\dot V_i)$ with $\beta=10$, where $\dot V_i$ is the kinematic time derivative of $V$ on PushT and the difference $V(t_{i})-V(t_{i-1})$ on stacking. It penalizes any increase of $V$ within the chunk. The \emph{$\Sigma V$} objective is the running cost $\sum_iV(t_i)$, the cost of a model-predictive controller on $V$. Neither defines an envelope, so their searches use a single patience $P_{\mathrm u}$ on the improvement of the objective, together with the generation cap and numerical stopping.

\subsection{Funnel Families}
\label{app:funnel_families}
\paragraph{Fixed shapes.}
With normalized chunk time $s=t/H$, the endpoint-only, plateau and linear funnels have fixed parameters,
\begin{equation}
    \begin{gathered}
    \rho^{\mathrm{end}}_k(t)=\begin{cases}\bar V,&s<1,\\(1-\gamma)V_k,&s=1,\end{cases}
    \qquad
    \rho^{\mathrm{lin}}_k(t)=V_k(1-\gamma s),\\[3pt]
    \rho^{\mathrm{pl}}_k(t)=V_k\Bigl[1+\bigl(\eta+(\delta-\eta)\,S(s/s_0)\bigr)\bigl(1-F(s)\bigr)-\gamma F(s)\Bigr],
    \end{gathered}
    \label{eq:fixed_funnels}
\end{equation}
where $\bar V$ is a constant far above any reachable $V$, so that endpoint-only leaves the interior samples free, and $S(u)=v^2(3-2v)$ with $v=\operatorname{clip}(u,0,1)$ is the clamped cubic step of \eqref{eq:clamped_smoothstep}. In the plateau, $F(s)=S((s-1+s_0)/s_0)$ moves the envelope to its terminal value at the end of the chunk. The plateau rises from $(1+\eta)V_k$ to $(1+\delta)V_k$ over the first fraction $s_0$ of the chunk, holds it, and falls to $(1-\gamma)V_k$ over the last, with $\eta=\gamma=0.1$ and $\delta\in\{0.5,1,3\}$. The ramps make the envelope continuous, and therefore Lipschitz, at both ends of the chunk. We set $s_0=0.02$, an arbitrary small value, so that the plateau holds over almost the whole chunk. For a fixed $\gamma>0$, the plateau and linear funnels, like the multiplicative family below, meet the terminal and transient requirements directly, and both are Lipschitz in time. The endpoint-only funnel jumps from $\bar V$ to the terminal value at the last sample, so it has no interior bound to verify.

\paragraph{Additive and multiplicative families.}
The fitted funnels need a parameterization that meets the four requirements for every value of its parameters, so that fitting cannot break them. Beyond these, the parameters should be few and interpretable, and the transient allowance should have an adjustable size and timing, since demonstrated regressions differ in how large they are and when they occur. The requested descent should also be able to start late in the chunk, after a preparatory motion. We build the family from the simplest functions with these properties, a linear ramp for the descent and cubic steps for the allowances. The learned and calibrated funnels use
\begin{equation}
    \rho_k(t)=V_k-\gamma_kV_kD_{\sigma_k}(s)+\delta_k\,b_{\tau_k}(s)+\eta_k\,c_0(s),
    \label{eq:adaptive_envelope}
\end{equation}
with the profiles
\begin{equation}
    D_\sigma(s)=\max\Bigl(0,\tfrac{s-\sigma}{1-\sigma}\Bigr),\quad
    b_\tau(s)=\begin{cases}
        S(s/\tau),&s\le\tau,\\
        S((1-s)/(1-\tau)),&s>\tau,
    \end{cases}
    \qquad c_0(s)=1-S(s),
    \label{eq:adaptive_profiles}
\end{equation}
where $S$ is the clamped cubic step defined above. The descent $D_\sigma$ ramps from $0$ at the onset $s=\sigma$ to $1$ at $s=1$, the bump $b_\tau$ peaks at $s=\tau$, and $c_0$ decays from $1$ to $0$. Here $\gamma_k\in[0,1]$ is the requested relative progress, $\delta_k$ and $\eta_k$ are the transient and start allowances, $\tau_k$ is the bump time and $\sigma_k$ the descent onset. In the \emph{additive} family the allowances are in units of $V$, scaled by the median training value of $V_k$. The \emph{multiplicative} family takes them as fractions of the current value,
\begin{equation}
    \rho_k(t)=V_k\bigl[1-\gamma_kD_{\sigma_k}(s)+\delta_k\,b_{\tau_k}(s)+\eta_k\,c_0(s)\bigr],
    \label{eq:multiplicative_envelope}
\end{equation}
so that the funnel shrinks with $V_k$ as the task progresses.

\paragraph{How the families meet the requirements.}
Each requirement follows from the values and slopes of the three profiles, for every admissible parameter value.
\begin{itemize}
    \item \emph{Anchoring.} At $s=0$, $D_\sigma=b_\tau=0$ and $c_0=1$, so $\rho_k(0)=V_k+\eta_k$ in the additive family and $(1+\eta_k)V_k$ in the multiplicative one. The funnel starts at $V_k$ raised by the start allowance, which we floor at $0.1\,V_k$ (Appendix~\ref{app:initial_lift}), and $c_0$ fades this allowance over the chunk.
    \item \emph{Terminal contraction.} At $s=1$, $D_\sigma=1$ and $b_\tau=c_0=0$, so $\rho_k(H)=(1-\gamma_k)V_k$ for any allowances, bump time and onset. The descent carries the requested progress, and both allowance profiles vanish at the endpoint.
    \item \emph{Bounded transient.} All three profiles take values in $[0,1]$, and the descent term only lowers the envelope. The multiplicative family therefore stays below $(1+\delta_k+\eta_k)V_k$, which gives $d_k=\delta_k+\eta_k$. The additive family stays below $V_k+\delta_k+\eta_k$, which gives $d_k=(\delta_k+\eta_k)/V_k$, a relative bound that grows as $V_k$ decreases (Appendix~\ref{app:theorem_notes}).
    \item \emph{Lipschitz in time.} The clamped cubic step is $C^1$ with largest slope $3/2$. The bump rises over $[0,\tau]$ and falls over $[\tau,1]$, with zero slope at its peak, so its slope is at most $3/\bigl(2\min(\tau,1-\tau)\bigr)$, and $c_0$ has slope at most $3/2$. The ramp is Lipschitz with slope $1/(1-\sigma)$, including at its onset. With the bump time in $[0.1,0.9]$ and the onset before the last sample, every fitted envelope is therefore Lipschitz in $t$ with constant at most
    \begin{equation}
        L_{\rho_k}\le\frac1H\Bigl(\frac{\gamma_kV_k}{1-\sigma_k}+\frac{3\,\delta_k}{2\min(\tau_k,1-\tau_k)}+\frac{3\,\eta_k}{2}\Bigr),
        \label{eq:envelope_lipschitz}
    \end{equation}
    with $\delta_k$ and $\eta_k$ multiplied by $V_k$ in the multiplicative family.
\end{itemize}
The remaining parameters fit the shape to the demonstrations: $\delta_k$ sets the size of the transient allowance and $\tau_k$ its timing, and $\sigma_k$ delays the requested descent until after a preparatory motion.

\subsection{Fitting Funnels to Demonstrations}
\label{app:funnel_fitting}
The learned and calibrated funnels take their parameters from how the task potential evolves along demonstrated chunks. The labels come from demonstrations only, measured with the fixed task potential, and not from refined rollouts or their outcomes. A fitted funnel therefore describes how far the potential rises within a chunk and how much progress it makes along demonstrated behavior, and refinement requires proposed chunks to stay within that envelope. The fitted parameters are fixed before refinement, and every evaluated leniency level is reported in Appendix~\ref{app:full_results}.

\paragraph{Demonstration labels.}
The policy is trained on the full demonstration dataset. Funnel fitting uses a separate three-way split of the same demonstrations into training, validation and test groups (70/15/15\%), made at the level of complete episodes before windows are extracted. Task outcomes are evaluated in simulator episodes from the sampled initial conditions. Each window of horizon $H$ starting at boundary $k$ gives the initial value $V_k$, the clipped relative endpoint progress $p_k=\operatorname{clip}(1-V_{k+1}/V_k,0,1)$, and the interior trace $V(t_i)$, $1<i<n$. We use the demonstrated prefix up to first success and, for sequential tasks, exclude windows that undo a completed prerequisite according to the task's completion gates.

\paragraph{Calibrated funnels.}
Calibrated funnels use the additive family of \eqref{eq:adaptive_envelope} with constant parameters, fitted directly to the demonstration labels. The training and calibration episodes are disjoint. The bump time and onset are fixed to $\tau=0.5$ and $\sigma=0$, and the progress $\gamma$ is the 10\% quantile of the progress labels $p_k$ of the training chunks. The two allowances share one scale, $\delta=\eta=c$. Since the envelope is affine in $(\delta,\eta)$ at fixed $\gamma$, each calibration chunk has a smallest scale $c_k$ for which the envelope contains its samples before the endpoint, the largest ratio between the room it needs above the descent and the room a unit allowance gives. The calibration score of an episode is the largest $c_k$ over its chunks, so an episode is contained at scale $c$ when all its included interior samples lie inside the funnel. With $n$ calibration episodes, we set $c$ to the $\lceil(n+1)\cdot0.9\rceil$-th smallest episode score, the finite-sample split-conformal upper quantile at coverage $0.9$~\citep{angelopoulosGentleIntroductionConformal2022b}. Under exchangeability of the calibration episodes and a new demonstration episode, the new episode has all its included interior samples inside the funnel with probability at least $0.9$. The global funnel fits one parameter set to all chunks, and the per-stage funnel one set to the chunks of each stage of a sequential task, falling back to the global set for a stage with fewer than 10 training or calibration episodes, or when the order statistic above exceeds the number of episodes.

\paragraph{Learned funnels.}
Learned funnels let the parameters of the family depend on the observation, which tests whether the allowance and the requested progress benefit from adapting to the state. The predictor $\pi_\vartheta$ reads the same observation history as the policy and is trained with the pinball loss $\ell_q(u)=\max(qu,(q-1)u)$, whose minimizer is the $q$-quantile of $u$~\citep{koenkerRegressionQuantiles1978}. At leniency $\alpha$ the three heads minimize
\begin{align}
    \mathcal L_\gamma&=\textstyle\sum_k\ell_{1-\alpha}\bigl(p_k-\gamma_\vartheta(o_k)\bigr),\notag\\
    \mathcal L_{\mathrm{tr}}&=\textstyle\sum_k\frac{1}{\max(V_k,1)}\,\ell_\alpha\Bigl(\max_{1<i<n}\bigl[V(t_i)-\rho_k(t_i;\operatorname{sg}[\gamma_k])\bigr]\Bigr),\label{eq:funnel_losses}\\
    \mathcal L_\sigma&=\textstyle\sum_{k:\,p_k>0}\ell_\alpha\bigl(\sigma^{\mathrm{lab}}_k-\sigma_\vartheta(o_k)\bigr),\notag
\end{align}
Their minimizers satisfy the conditional quantile conditions
\begin{equation}
    \Pr\bigl(p_k\ge\gamma_k\mid o_k\bigr)\approx\alpha,
    \qquad
    \Pr\Bigl(\max_{1<i<n}\bigl[V(t_i)-\rho_k(t_i)\bigr]\le0\;\Big|\;o_k\Bigr)\approx\alpha,
    \label{eq:funnel_quantiles}
\end{equation}
These are two separate marginal targets over demonstrated chunks. The first states that the requested endpoint progress is met on a fraction $\alpha$ of the chunks, and the second that the interior allowance contains the trace on a fraction $\alpha$ of the chunks. A larger $\alpha$ lowers the requested progress and widens the allowance, which gives a more permissive funnel. We clip the demonstrated relative progress to $[0,1]$, so a chunk that regresses at its endpoint supplies a zero-progress target, and the predicted $\gamma_k$ lies in the same range. The smallest progress a learned funnel requests is therefore zero. The transient allowance is fitted for the current progress prediction, which is held fixed ($\operatorname{sg}$) in $\mathcal L_{\mathrm{tr}}$ within each epoch. The initial sample is covered by the start allowance and is therefore left out of the transient target. The onset labels $\sigma^{\mathrm{lab}}_k$ are the delays before descent in progressing windows. These quantile targets concern demonstrated chunks, the calibrated coverage concerns demonstration episodes, and the verification rates of Appendix~\ref{app:verification_details} count executed chunks, so the three quantities summarize different populations.

The predictor has two hidden layers of 64 \acp{SiLU} and outputs parameter tuples at four leniency levels, $\alpha\in\{0.5,0.8,0.9,0.95\}$. Its parameterization makes $\gamma$ nonincreasing and $\delta$, $\eta$, $\sigma$ nondecreasing in $\alpha$, with a shared $\tau\in[0.1,0.9]$ and onset no later than the last interior sample, so higher levels give nested, more permissive envelopes. Training uses AdamW (learning rate $10^{-3}$, weight decay $10^{-4}$, batch size 512) for up to 200 epochs.

%% file: sections/appendix/implementationdetails.tex
\section{Refinement and Implementation}

\subsection{Typical-Set-Constrained \acs*{CMA-ES} Refinement}
\label{sec:appendix_cmaes_refinement}
Every planning call runs a fresh search, shares the frozen task potential and funnel across all evaluations, and evaluates the nominal policy seed first. Candidate decoding is deterministic for fixed conditioning and numerical integration settings.

\paragraph{Search distribution and band projection.}
For a flattened seed $a\in\mathbb{R}^d$, \ac{CMA-ES} samples $a_\ell=m+\sigma BD\epsilon_\ell$ with $\epsilon_\ell\sim\mathcal{N}(0,I_d)$, where $m$ is the distribution mean, $\sigma$ the step size, and $C=BD^2B^\top$ the eigendecomposition of the search covariance $C$. Every evaluated seed is $\Psi(a_\ell)$, with $\Psi$ from \eqref{eq:seed_shell}, and degenerate or non-finite inputs are rejected. The projection first centers the seed, subtracting the mean $\bar a$ of its $d$ entries so that they average to zero, and then clips the radius of the centered seed to the band. After each \ac{CMA-ES} update the mean is projected onto the band, $m\leftarrow\Psi(m)$. The raw population, covariance, and evolution paths are not modified, so the update itself is standard \ac{CMA-ES}. All horizon and channel entries form one global group, so $\bar a$ is taken over the whole seed, and a seed with zero-mean entries and a radius inside the band is unchanged. Without the band, a \ac{CMA-ES} initialized at a seed drawn from the source can drift radially away from the origin over the generations, and the decoded chunks then come from seeds the flow was never trained to transport.

For a standard Gaussian, the centered radius satisfies $\|a-\bar a\mathbf{1}\|^2\sim\chi^2_{d-1}$. We place the band endpoints at tail probabilities $\Phi(-\zeta)$ and $\Phi(\zeta)$ using the Wilson--Hilferty approximation~\citep{wilsonDistributionChiSquare1931}. With $\nu=d-1$ and $b=2/(9\nu)$,
\begin{equation}
    r_{\mathrm{lo}}=\sqrt{\nu(1-b-\zeta\sqrt b)^3},\qquad
    r_{\mathrm{hi}}=\sqrt{\nu(1-b+\zeta\sqrt b)^3}.
    \label{eq:source_shell_radii}
\end{equation}
The approximation is used for $\nu\ge30$. Fixing both radii to $\sqrt{\nu}$ recovers the single-sphere projection of standardized \ac{CMA-ES}~\citep{stolzlePhysibleMDMEnforcingInferenceTime2026}. We use $\zeta=4$: a source draw then lies outside the band with probability about $6\times10^{-5}$, so in almost every call the projection centers the nominal seed and keeps its radius. The first evaluated plan is decoded from this centered seed. This differs slightly from the plan of nominal execution, which decodes the uncentered draw. The band constrains the seed's mean and radius and leaves its direction, channel-wise structure and correlations free.

\paragraph{Euclidean flows.}
The search runs on the flow's Gaussian seed, of dimension $d$ equal to the chunk length times the width of the action representation (for stacking, $32\times8=256$). The initial mean is the projected nominal seed, which decodes to the first evaluated plan.

\paragraph{Riemannian flows.}
The search runs Euclidean \ac{CMA-ES} and \eqref{eq:seed_shell} in chart coordinates of the source at its center. Each waypoint has seven coordinates, a translation $a_p\in\mathbb R^3$, a rotation vector $a_R\in\mathbb R^3$, and a gripper coordinate $a_r\in\mathbb R$, so $d=7\times32=224$ for stacking. They are mapped waypoint-wise to carriers by the source's own sampling map, the exponential chart at the source center with the per-axis source scales $S=\operatorname{diag}(1,1,1,\sigma_{\mathrm{rot}},\sigma_{\mathrm{rot}},\sigma_{\mathrm{rot}},1)$,
\begin{equation}
    z_0=\mathcal C(\Psi(a)),\qquad
    \mathcal C(a_p,a_R,a_r)
    =\bigl(a_p,\bar R\exp(\sigma_{\mathrm{rot}}[a_R]_\times),a_r\bigr),
    \label{eq:chart_seed_decode}
\end{equation}
which matches \eqref{eq:rotation_prior}. Then $z_0$ is written in the 13-dimensional carrier that the flow evolves. In these coordinates the source is exactly $a\sim\mathcal N(0,I_d)$, so the band constrains the Gaussian the flow was trained to transport, with no approximation. The nominal seed is expressed in these coordinates and projected, and, as for Euclidean flows, it is the initial search mean.

Searching in chart coordinates keeps the search distribution Gaussian in the coordinates in which the source is Gaussian, so the typical-set band applies exactly. A Riemannian \ac{CMA-ES}~\citep{coluttoCMAESRiemannianManifolds2010} is an alternative that searches on the carriers themselves, with tangent-space sampling at the current mean, a weighted Fréchet mean update and parallel transport of the adapted covariance.

\paragraph{Best plan and stopping.}
\ac{CMA-ES} updates its distribution from the ranks of the unnormalized log-sum-exp $J_k$ of \Cref{sec:optimization}, with a fixed grid and temperature throughout each search. The first population contains the projected nominal seed, whose plan becomes the initial best plan (Algorithm~\ref{alg:cmaes_detailed}). After each generation, the candidate with the lowest objective replaces the best plan when its objective is strictly lower, so the nominal plan is returned when it has the lowest objective of the search. The best plan carries the estimated-verification verdict $v^\star\in\{0,1\}$ of its trace, which is recomputed whenever the best plan changes.

Patience is an optional stopping mechanism, enabled in the funnel ablation. A generation counts as progress when the best objective satisfies $J^\star<J_{\mathrm{ref}}-T$ with
\begin{equation}
    T=\max\bigl(\epsilon_{\mathrm{abs}},\,\epsilon_{\mathrm{rel}}(J_{\mathrm{nom}}-J_{\mathrm{ref}})\bigr),
    \label{eq:patience_threshold}
\end{equation}
where $J_{\mathrm{nom}}$ is the objective of the projected nominal plan and $J_{\mathrm{ref}}$ is the best objective at the last generation that counted as progress. $J_{\mathrm{ref}}$ is then reset to $J^\star$. It starts at $+\infty$, so the first generation always counts as progress. Small gains accumulate until they clear $T$, and scaling $\epsilon_{\mathrm{rel}}$ by the improvement already achieved makes the rule invariant to the offset of the objective. $J_{\mathrm{ref}}$ is used by this stopping rule and does not enter the objective. The search stops after $P$ generations without progress, with $P=P_{\mathrm v}$ when the best plan passes estimated verification and $P=P_{\mathrm u}$ otherwise. The shorter $P_{\mathrm v}$ gives a verified plan a few further generations to improve. The search also stops at the generation cap and on numerical degeneracy (Table~\ref{tab:search_settings}). The search comparison of Appendix~\ref{app:search_ablation} disables patience and the generation cap and stops every iterative search on a time budget of 1.5\,s per chunk.

\begin{figure}[t]
\SetCommentSty{algcomment}
\SetKwProg{Fn}{Function}{}{}
\SetKwFunction{Cost}{Cost}\SetKwFunction{Decode}{Decode}\SetKwFunction{Verify}{Verify}
\SetKwComment{tcp}{\(\triangleright\) }{}
\begin{algorithm2e}[H]
    \footnotesize
    \DontPrintSemicolon
    \caption{Seed search within one planning call (detailed).}
    \label{alg:cmaes_detailed}
    \KwIn{$x_k$, $o_k$, frozen $\rho_k$, nominal source coordinates $a^{\mathrm{nom}}$}
    $J_{\mathrm{ref}}\leftarrow+\infty$; $p\leftarrow0$; initialize $(m,\sigma,C)$ on the band\;
    \For{$g=1,2,\dots$}{
        $a_1\leftarrow a^{\mathrm{nom}}$ if $g=1$; otherwise $a_1\sim\mathcal N(m,\sigma^2C)$; $a_\ell\sim\mathcal N(m,\sigma^2C)$, $\ell=2,\dots,\lambda$\;
        $w_\ell\leftarrow\Decode(a_\ell;o_k)$; $J_\ell\leftarrow\Cost(w_\ell;x_k,\rho_k)$, $\ell=1,\dots,\lambda$ \tcp*[r]{in parallel}
        \lIf{$g=1$}{$(w^\star,J^\star)\leftarrow(w_1,J_1)$; $J_{\mathrm{nom}}\leftarrow J_1$ \tcp*[f]{projected nominal plan}}
        Rank-update $(m,\sigma,C)$ from $J_{1:\lambda}$; $m\leftarrow\Psi(m)$\;
        $\ell'\leftarrow\arg\min_\ell J_\ell$; \lIf{$J_{\ell'}<J^\star$}{$(w^\star,J^\star)\leftarrow(w_{\ell'},J_{\ell'})$}
        \lIf{$J^\star<J_{\mathrm{ref}}-T$}{$p\leftarrow0$; $J_{\mathrm{ref}}\leftarrow J^\star$ \textbf{else} $p\leftarrow p+1$ \tcp*[f]{threshold $T$ of Eq.~\eqref{eq:patience_threshold}}}
        \lIf{$w^\star$ changed}{$v^\star\leftarrow\Verify(w^\star)$}
        \lIf{$p\ge P(v^\star)$ (patience, when enabled) or the time budget is spent}{stop}
        \lIf{numerical degeneracy or generation cap}{stop}
    }
    \KwRet{$(w^\star,J^\star,v^\star)$}\;
    \Fn{\Verify{$w$}}{
        $g_i\leftarrow V(T(x(t_i)))-\rho_k(t_i)$ on the stored rollout of $w$\;
        $\hat L_i\leftarrow$ local rates from $g_{1:n}$ \tcp*[r]{Appendix~\ref{app:local_rates}}
        \KwRet{$[\hat B_k\le0]$, with $\hat B_k$ from \eqref{eq:interval_bound} at rates $\hat L_i$}\;
    }
    $P(v)=P_{\mathrm v}$ if $v=1$, else $P_{\mathrm u}$; \Cost{} and \Decode{} as in Algorithm~\ref{alg:cmaes_refinement}\;
\end{algorithm2e}
\end{figure}

\begin{table}[t]
    \centering
    \caption{Refinement settings used in all reported refined runs. Population and parents follow the \acs*{CMA-ES} defaults, with the standard log-linear recombination weights~\citep{hansenCMAEvolutionStrategy2023}.}
    \label{tab:search_settings}
    \small
    \begin{tabular}{@{}ll@{}}
        \toprule
        Parameter & Setting\\
        \midrule
        Population / parents & $20/10$ ($\lambda=4+\lfloor3\ln d\rfloor$)\\
        Initial step size / generation cap & $1/30$\\
        Covariance eigendecomposition & Every generation\\
        Seed region & Band with $\zeta=4$, one global group; mean projected after each update\\
        Meaningful improvement & $\epsilon_{\mathrm{abs}}=10^{-2}$, $\epsilon_{\mathrm{rel}}=5\times10^{-3}$ (relative to achieved improvement)\\
        Patience & $P_{\mathrm u}=6$ / $P_{\mathrm v}=4$ generations (best plan not verified / verified)\\
        Verification & Estimated, window $w_L=0.02$\,s, on the physics-step grid\\
        Start allowance & $\eta_k\ge0.1\,V_k$\\
        Ranked samples & From the end of the first control step\\
        Objective temperature & $\beta=10$ (PushT), $\beta=50$ (stacking)\\
        Numerical stopping & $\sigma\sqrt{\lambda_{\max}(C)}<10^{-12}$ or $\operatorname{cond}(C)>10^{14}$\\
        \bottomrule
    \end{tabular}
\end{table}

\subsection{Gradient Search on a Differentiable Surrogate}
\label{app:surrogate_gradient}
The gradient searches of Table~\ref{tab:search_ablation_all} optimize the seed $a$ by descending $F_k(a)=J_k(\psi_\theta(1\mid a,o_k))$ of Section~\ref{sec:optimization}, with the rollout $\mathcal R_\kappa$ replaced by a differentiable model of the PushT dynamics and without the band projection.

\paragraph{Quasi-static pushing model.}
The model is the discrete step of the native PushT environment, our Chipmunk2D~\citep{slembckeChipmunk2D} reimplementation of the Diffusion Policy environment~\citep{chiDiffusionPolicyVisuomotor2024} (Appendix~\ref{sec:appendix_env_details}), written as a smooth function with the same constants. Only the contact switch and the impulse clamp are smoothed. The state $x=(p_a,v_a,p_b,\theta_b,\xi_b)\in\mathbb R^{10}$ holds the agent position and velocity, the pose of the block's center of mass and its twist over the last step. Each physics substep ($\Delta t=0.01$\,s) follows Chipmunk's order. The agent's kinematic \ac{PD} controller updates $v_a\leftarrow v_a+(k_p(u-p_a)-k_vv_a)\Delta t$ with the held target $u$, $k_p=225$ and $k_v=30$. The block starts every substep at rest, since the native space has zero damping, which in Chipmunk removes all body velocity at each step: the block moves only while it is pushed, and the pushing is quasi-static. Contacts are frictionless and inelastic, as in the native environment, whose shapes have zero friction, so they transmit only normal impulses. They are resolved at the positions at the start of the substep by four passes of sequential impulses, between the agent disc and each convex part of the T and between every block vertex and the four walls. For a contact that requires the block point to move along $n$, with arm $r$ from the center of mass and block inverse inertia $M^{-1}=\operatorname{diag}(1/m,1/m,1/I)$, the impulse along $c=(n,r\times n)$ is $\lambda=\Delta/(c^\top M^{-1}c)$ for the missing normal speed $\Delta$, and accumulated impulses are kept nonnegative. A separate bias channel removes penetration beyond a slop of 0.1\,px, as in Chipmunk. Positions are then integrated with the resolved velocities. A waypoint target is held for 10 substeps, one control step.

\paragraph{Smoothing.}
For the agent disc and a convex part of the T, the signed distance is the differentiable conservative separating-axis distance of \citet{wongContactAwareSafetySoft2025},
\begin{equation}
    h(p_a)=\frac{1}{\beta_{\mathrm c}}\log\sum_{j=1}^{m}e^{\beta_{\mathrm c} s_j(p_a)}-\frac{0.4\log m}{\beta_{\mathrm c}},
    \qquad \hat n=\frac{\nabla_{p_a}h}{\|\nabla_{p_a}h\|},
    \label{eq:dcsat_distance}
\end{equation}
where $s_j$ is the separation of the disc and the part along the $j$-th of $m$ axes, the edge normals of the part and the directions from the disc center to its vertices, and $\beta_{\mathrm c}=80$. The log-sum-exp is a smooth maximum over the axes, and the shift keeps it from overestimating the true separation. The contact normal $\hat n$ is its normalized gradient, and the two parts of the T are combined by a smooth minimum. Distances to the walls are exact. Each constraint is gated by $g=\operatorname{sigmoid}(-h/w_c)$, $w_c=0.1$\,px, and its impulse is updated by
\begin{equation}
    \Delta\lambda=\frac{g\,\Delta}{c^\top M^{-1}c}-(1-g)\,\lambda,
    \qquad
    \lambda\leftarrow g\,\operatorname{sp}_{w_i}(\lambda),\quad
    \operatorname{sp}_w(x)=w\log\bigl(1+e^{x/w}\bigr),
    \label{eq:soft_contact}
\end{equation}
a soft constraint whose fixed point does not depend on the number of passes, a hard contact for $g=1$ and no constraint for $g=0$. The softplus of width $w_i=1$\,px/s replaces the clamp $\max(\lambda,0)$, and the gate in front of it keeps inactive contacts from transmitting any impulse, and the bias channel uses the same gate. Contacts farther than $4.1$\,px, where gate and bias are both below $e^{-40}$, are skipped, which changes neither values nor derivatives. On recorded chunks of the nominal policy, the final block position of the model is within 0.16\,px of the native simulator in median and 0.5\,px at the 90th percentile, in a $512\times512$\,px workspace. The model is differentiated in double precision.

\paragraph{Gradient through the flow.}
One evaluation decodes the seed by the 8-step Euler sampler of the flow, maps the result to held targets anchored at the agent's observed position, rolls them out in the model from the observed start state $x_k$, and evaluates $J_k$ of \Cref{sec:optimization} on the model's trace, with the same frozen funnel, grid and temperature as the population searches. Its gradient with respect to the seed is a single reverse-mode chain: Enzyme~\citep{mosesInsteadRewritingForeign2020} differentiates the rollout and the objective and then the decoding, and the vector--Jacobian product of the whole sampler is computed by Enzyme on the sampler compiled with Reactant.jl\footnote{\url{https://github.com/EnzymeAD/Reactant.jl}}. The network weights and the conditioning are fixed, and the seed is the optimization variable. Since $V$ depends only on the block pose and the plan reaches the block only through agent--block contact, a plan whose agent stays out of the contact range for a whole chunk has an identically zero gradient. The search then stops and keeps the nominal plan.

\paragraph{Optimizers.}
Both searches start from the nominal seed. Adam uses a learning rate of $0.05$ and default moments, so that each seed coordinate moves by at most about $0.05$ per iteration. \ac{L-BFGS} keeps 10 curvature pairs, stores a pair only if $s^\top y>10^{-10}\|s\|\|y\|$, restarts along $-\nabla F_k$ when the direction is not a descent direction, and selects the step by Armijo backtracking ($c_1=10^{-4}$, halving, at most 10 backtracks) from a unit trial step, with every trial step capped at norm 1 in seed space in place of a trust region. Both stop at the 1.5\,s budget, at a stationary point ($\|\nabla F_k\|\le10^{-6}(1+|F_k|)$), on non-finite values, or, for \ac{L-BFGS}, when the line search fails. The search returns the iterate with the lowest loss on the model, the nominal seed unless some iterate is strictly better.

\paragraph{Planning and execution models.}
All searches optimize the same funnel objective. The population searches evaluate it on rollouts of the native simulator. The gradient searches evaluate it on rollouts of the smoothed differentiable reimplementation, starting from the observed start state of the chunk, and differentiate through that rollout and the flow decoder. Every loss, gradient, line-search trial and stopping decision of a gradient search comes from this model, which agrees with the native simulator to within 0.16\,px in median on recorded chunks. The selected plans of all searches are executed in the native environment.

\subsection{Policy Architecture and Training}
\label{app:policy_training}
\paragraph{Observations.}
PushT observes the agent position, the block position, and the block angle as its cosine and sine, six features per step. Stacking observes the end-effector pose, the cube poses, their relative vectors, the measured gripper aperture and last command, and proximity and gripper features, with rotations in the six-dimensional representation of \citet{zhouContinuityRotationRepresentations2020}, 74 features per step. Two-nut assembly observes the end-effector pose, the world pose of each nut, its pose relative to the hand and its offset from its peg, and the measured gripper aperture and last command, 53 features per step.

\paragraph{Demonstration windows and normalization.}
We train on dense, stride-one windows of the demonstrations, padding at episode boundaries by repeating the first or last sample, and normalize features to $[-1,1]$ with min--max statistics from all policy-training demonstrations. Two choices are specific to our setting. For stacking, the gripper target is the signed command at each control step, while the measured aperture and the previous command are observations. Decoding executes the command through its sign. Rotation blocks are left unnormalized, since they already lie in $[-1,1]$, before conversion to the \ac{RMFM} carrier. The policy is trained on all demonstrations. The funnels are fitted on a separate three-way split of the same demonstrations, described in Appendix~\ref{app:funnel_fitting}.

\paragraph{Velocity-regression objective.}
For each window, draw one source sample and one independent flow time uniformly on $[0,1]$. The model regresses the corresponding path velocity with normalized weighted squared error,
\begin{equation}
    \mathcal L(\theta)=\mathbb E\!\left[
    \frac{\sum_{i,j}\omega_j\bigl(u_{\theta,ij}-u^\star_{ij}\bigr)^2}
         {\sum_{i,j}\omega_j}\right],
    \label{eq:weighted_fm_loss}
\end{equation}
where $i$ indexes waypoint tokens and $j$ velocity channels. The action space is a product of a continuous pose space and a binary-like gripper command, which gives two kinds of target: the pose velocity varies smoothly over every window, while the gripper target is nearly constant except at sparse switching events, where its error is large. With uniform weights these rare but large gripper errors can dominate the gradient. Pose-valued policies therefore weight each pose channel by 10 and the gripper channel by 1, so that the pose dominates the loss while the gripper term still contributes. PushT has no gripper and uses uniform weights. The targets are Euclidean path velocities or body twists and gripper rates (Appendix~\ref{sec:appendix_lie_group_preliminaries}).

\paragraph{Architecture and optimization.}
On stacking, the conditional vector field is a transformer based on the \ac{DiT} architecture~\citep{peeblesScalableDiffusionModels2023} with one action token per waypoint, learned positional embeddings, and cross-attention to observation tokens. It uses 6 blocks, width 384, 8 heads, \ac{MLP} ratio 4, and \ac{GELU}. Each block has self-attention, cross-attention, and an \ac{MLP}, with \ac{adaLN-Zero} modulation driven by a 64-dimensional sinusoidal flow-time embedding followed by a Dense--Mish--Dense map. We use no dropout. Two-nut assembly uses the same architecture, trained on its own demonstrations. On PushT, the vector field is the conditional one-dimensional U-Net of Diffusion Policy~\citep{chiDiffusionPolicyVisuomotor2024} over the 24 waypoints, with three levels of 128, 256 and 512 channels, two residual blocks of kernel 5 per level, GroupNorm with 8 groups, and Mish. The flow time is embedded by a 256-dimensional sinusoidal embedding and a Dense--Mish--Dense map and concatenated with the normalized observations of the two history steps, and every residual block applies a \ac{FiLM} scale and shift predicted from this vector. As in Diffusion Policy, the up path concatenates the outputs of the two coarser levels. We replace the strided convolution by a stride-one convolution followed by subsampling, and the transposed convolution by nearest-neighbour upsampling followed by a convolution, since the strided convolution miscompiles under Reactant. The U-Net has 17.6M parameters. Training uses AdamW with learning rate $10^{-4}$ and weight decay $10^{-6}$, linear warmup followed by cosine decay to $10^{-6}$, and \ac{EMA} weights with decay 0.999 for evaluation. Table~\ref{tab:task_choices} gives task-specific settings.
Training uses Julia~\citep{Julia-2017}, Lux.jl~\citep{palLuxExplicitParameterization2025}, and Reactant.jl.

\begin{table}[t]
    \centering
    \caption{Task-specific training, execution and evaluation choices. Every chunk is executed in full before replanning.}
    \label{tab:task_choices}
    \label{tab:policy_io}
    \label{tab:task_candidates}
    \label{tab:task_evaluation}
    \footnotesize
    \setlength{\tabcolsep}{3pt}
    \begin{tabular}{@{}llll@{}}
        \toprule
        & PushT & Three-cube stacking & Two-nut assembly\\
        \midrule
        Demonstrations & 206 & 1000 & 1000\\
        Training iterations & 30k & 100k & 100k\\
        Batch size / warmup & 128 / 500 steps & 128 / 1000 steps & 128 / 1000 steps\\
        Prediction / history & 24 waypoints / 2 steps & 32 waypoints / 5 steps & 32 waypoints / 2 steps\\
        Execution horizon & 24 steps & 32 steps & 32 steps\\
        State features $\times$ history & $6\times2=12$ & $74\times5=370$ & $53\times2=106$\\
        Network & U-Net (\acs*{FiLM}) & \acs*{DiT} & \acs*{DiT}\\
        Flow & \acs*{OT} & \acs*{RMFM} & \acs*{RMFM}\\
        Waypoint target & Agent position & \acs*{EE} pose + gripper command & \acs*{EE} pose + gripper command\\
        Network channels (in/out) & $2/2$ & $13/7$ & $13/7$\\
        Flow integration & 8 Euler steps & 10 Lie--Euler steps & 10 Lie--Euler steps\\
        Control / physics rate & 10\,Hz / 100\,Hz & 20\,Hz / 500\,Hz & 20\,Hz / 500\,Hz\\
        Step budget & 400 & 400 & 600\\
        Potential & Pose error (App.~\ref{app:pusht_candidate}) & Phased (App.~\ref{app:phased_calibration}) & Phased (App.~\ref{app:nut_candidate})\\
        Funnel split (fraction) & 70/15/15\,\% & 70/15/15\,\% & 70/15/15\,\%\\
        \midrule
        Score $r\in[0,1]$ & $\mathrm{Coverage}(\mathrm{block},\mathrm{goal})$ & $\tfrac12\,a_{RG}(1+a_{BR})$ & $V$\\
        Success & $r>0.95$ & $\mathrm{Stacked}\wedge\mathrm{Open}$ & $\mathrm{Seated}_{\mathrm{sq}}\wedge\mathrm{Seated}_{\mathrm{rd}}\wedge\mathrm{Clear}$\\
        \bottomrule
    \end{tabular}
    \par\vspace{2pt}{\scriptsize $a_{XY}=\mathrm{Overlap}(X,Y)\cdot\mathrm{Contact}(X,Y)$, the projected footprint overlap of $X$ on $Y$ times their contact.}
\end{table}

\subsection{Environment, Controller, and Execution}
\label{sec:appendix_env_details}
\label{app:runtime}
Experiments use the initialization conventions in Table~\ref{tab:environment_settings}. Waypoint targets are held over a control period, while tracking and simulation advance at physics substeps. Rollout evaluation includes the initial state and all subsequent physics steps. For stacking, each waypoint is tracked by \ac{DLS} \ac{IK} followed by joint-space \ac{PD}, with gains (Table~\ref{tab:controller_gains}) calibrated once and then fixed: \ac{CMA-ES} searches the logarithms of the gains (population 12, 15 generations), scoring each candidate by the task success obtained when the controller replays the demonstrated waypoints of 50 demonstrations spread over the dataset. Two-nut assembly uses the same controller on the Sawyer arm, with gains calibrated in the same way on its own demonstrations (Table~\ref{tab:controller_gains}). PushT tracks agent waypoints with \ac{PD} ($k_p=225$, $k_v=30$). The PushT environment reproduces exactly the one of Diffusion Policy~\citep{chiDiffusionPolicyVisuomotor2024}, with the same geometry, physical constants, controller, reward and success predicate. That environment steps Chipmunk2D~\citep{slembckeChipmunk2D} through the Python binding pymunk. We call Chipmunk2D directly from Julia, which removes the Python interpreter from the rollout and lets the candidate rollouts of a generation run in parallel on threads.

\paragraph{Stacking success.}
Stacking uses the MimicGen \texttt{StackThree\_D1} success predicate. At its first success the robot opens the gripper (0.3\,s), retreats 5\,cm vertically (0.5\,s) and holds still (0.2\,s). The episode counts as a success only if the predicate still holds after this sequence, so the tower must stand without support.

\paragraph{Two-nut assembly success.}
Two-nut assembly uses robosuite's own check: each nut lies within 30\,mm of its peg axis in $x$ and in $y$ and below table height plus 50\,mm, and the gripper meets the native clearance condition. The check is evaluated on the current state, so a placement is not latched, and since it already requires the gripper to be clear, the episode ends at its first success without a release sequence.

\begin{table}[t]
    \centering
    \caption{Environment initialization and tracking.}
    \label{tab:environment_settings}
    \small
    \begin{tabular}{@{}>{\raggedright\arraybackslash}p{0.14\linewidth}>{\raggedright\arraybackslash}p{0.57\linewidth}>{\raggedright\arraybackslash}p{0.21\linewidth}@{}}
        \toprule
        Task & Initialization & Tracking\\
        \midrule
        PushT & Hard reset mode: block in $[100,400]^2$, agent in $[50,450]^2$ & Position \acs*{PD}\\[4pt]
        Cube Stacking & Each cube placed uniformly in $x,y\in[-0.2,0.2]$\,m with uniform yaw, overlaps rejected (the robosuite placement sampler~\citep{zhuRobosuiteModularSimulation})& \acs*{IK} with joint-space \acs*{PD}\\[4pt]
        Two-Nut Assembly & MimicGen \texttt{NutAssembly\_D0} reset: both nuts placed on the table with uniform yaw, the square nut at $x\in[-0.115,-0.110]$, $y\in[0.110,0.225]$ and the round nut at $x\in[-0.115,-0.110]$, $y\in[-0.225,-0.110]$, robot at the dataset's initial joint configuration & \acs*{IK} with joint-space \acs*{PD}\\
        \bottomrule
    \end{tabular}
\end{table}
\begin{table}[t]
    \centering
    \caption{\acs*{IK}-\acs*{PD} controller gains for stacking and for two-nut assembly (Sawyer). Twists are clipped to the limits.}
    \label{tab:controller_gains}
    \small
    \begin{tabular}{@{}lll@{}}
        \toprule
        Gain & Stacking & Two-nut assembly\\
        \midrule
        Task $k_p$ (position / rotation) & $22.88$ / $6.24$ & $44$ / $28$\\
        Task $k_d$ (position / rotation) & $1.43$ / $1.53$ & $0.81$ / $0.39$\\
        Joint $k_p$ / $k_d$ & $91.18$ / $85.28$ & $94.52$ / $88$\\
        Posture $k_p$ / $k_d$ & $1.40$ / $0.37$ & $3.50$ / $0.42$\\
        \acs*{DLS} damping & $8\times10^{-4}$ & $8\times10^{-4}$\\
        Twist limit (linear / angular) & $0.75$\,m/s / $1.5$\,rad/s & $0.75$\,m/s / $1.5$\,rad/s\\
        \bottomrule
    \end{tabular}
\end{table}

%% file: sections/appendix/proofs.tex
\section{Regularity, Verification and Convergence Results}
\label{sec:appendix_proofs}
\label{rem:boundary_relaxation}
This appendix supports the results of \Cref{sec:task_refinement}. Appendix~\ref{app:convergence_proof} states the convergence result in its product form and proves it. Appendix~\ref{app:simulator_regularity} gives conditions under which the task residual is Lipschitz, as the verification bound of \Cref{thm:objective_certification} requires, and Appendix~\ref{app:verification_proof} proves that bound. The remaining subsections extend \Cref{thm:receding_horizon_contraction} in three directions. A fading start allowance $\widetilde\rho_k(t)=\rho_k(t)+a_kc_0(t/H)$, with $a_k\ge0$, $c_0(0)=1$ and $c_0(1)=0$, keeps the terminal condition and enters the transient bound, and it enables verification from samples at the start of a chunk (Appendix~\ref{app:initial_lift}). Endpoint errors of at most $\epsilon_k\le\bar\epsilon$, from terminal violation of the funnel or from a mismatch between predicted and executed motion, give the recursion $V(y_{k+1})\le(1-\gamma_k)V(y_k)+\epsilon_k$, whose endpoint values approach a neighborhood of size $\bar\epsilon/\underline\gamma$ (Appendix~\ref{app:practical_certification}). Progress accumulated over windows of chunks admits preparatory chunks with $\gamma_k=0$ (Appendix~\ref{app:theorem_notes}).

\subsection{Convergence: Product Form and Proof of \Cref{thm:receding_horizon_contraction}}
\label{app:convergence_proof}
\begin{proposition}[Task contraction with iteration-dependent rates]
    \label{prop:product_contraction}
    Suppose each selected chunk $w_k^\star$ is executed for the full horizon $H>0$ before replanning, its evaluated task rollout matches execution, and it satisfies \eqref{eq:lyapfunnel} with the terminal and transient bounds of \eqref{eq:envelope} for some $\gamma_k\in[0,1]$ and $d_k\ge0$. Then, for every $m\ge1$,
    \begin{equation}
        V(y_{k+m})\le\left[\prod_{j=k}^{k+m-1}(1-\gamma_j)\right]V(y_k),
        \label{eq:iterated_contraction}
    \end{equation}
    and $\sup_{t\in[0,H]}V(y_{w_k^\star}^k(t))\le(1+d_k)V(y_k)$.
\end{proposition}
\begin{proof}
    At $t=H$, funnel satisfaction gives $V(y_{k+1})\le\rho_k(H)\le(1-\gamma_k)V(y_k)$, and iteration yields the product bound. Within iteration $k$, the funnel and the transient bound give $V(y_{w_k^\star}^k(t))\le\rho_k(t)\le(1+d_k)V(y_k)$.
\end{proof}
\begin{proof}[Proof of \Cref{thm:receding_horizon_contraction}]
    Substituting the uniform bounds $\gamma_k\ge\underline\gamma$ and $d_k\le\bar d$ into \Cref{prop:product_contraction} gives \eqref{eq:continuous_task_convergence}. Iterations have a fixed positive duration, so the bound tends to zero with physical time and eventually remains below any positive goal threshold.
\end{proof}

\subsection{Residual Regularity}
\label{app:simulator_regularity}
The verification bound in \Cref{thm:objective_certification} uses regularity of the task residual rather than smoothness of the full contact dynamics. The dynamics $f$ in \eqref{eq:overview_optimization} can represent stiff compliant-contact dynamics, or flows supplemented by velocity resets in an impulsive model. In either case, the following configuration-level assumptions suffice.

\begin{lemma}[Configuration-based residual regularity]
    \label{lem:residual_regularity}
    Let $q(t)$ denote the configuration of the world along the rollout $x_w^k$, that is, the generalized coordinates of the robot and the objects, without their velocities. Suppose that, for every candidate plan, $q$ is absolutely continuous on $[0,H]$ with speed bounded almost everywhere by a common $v_{\max}$. Let $F(q)=V(T(q))$ be Lipschitz on the reachable set with constant $L_F$, and let every envelope $\rho_k$ be Lipschitz with constant at most $L_\rho$. Then, uniformly over plans, the task residual is Lipschitz on every sampling interval with
    \begin{equation}
        L_i\le L_Fv_{\max}+L_\rho.
        \label{eq:residual_regularity_bound}
    \end{equation}
\end{lemma}
\begin{proof}
    Bounded speed gives $d(q(t),q(s))\le v_{\max}|t-s|$. Consequently,
    \[
        |F(q(t))-F(q(s))|\le L_Fv_{\max}|t-s|.
    \]
    Adding the envelope's Lipschitz bound yields the result.
\end{proof}
Configuration trajectories can satisfy these assumptions despite velocity jumps. For a $C^1$ composition $F$, a bound on its gradient over a suitable compact neighborhood provides $L_F$, and Appendix~\ref{app:phased_lyapunov} establishes this regularity for the phased construction. For the learned envelope, $L_\rho$ is bounded because $V_k$ is bounded on $\Omega$ and the predicted parameters are bounded. Numerical sample error and prediction mismatch require separate allowances if the samples are not exact values of the trajectory being verified.

\paragraph{Initial boundary of the funnel.}
\label{app:initial_lift}
With exact anchoring, $\rho_k(0)=V_k$, every plan starts with a zero residual at the first sample, $g_1=0$. The first interval then contributes $(g_2+L_1\Delta_1)/2$ to \eqref{eq:interval_bound}, which is positive unless the residual falls at least as fast as the rate allowed for it, so almost no chunk can be verified, regardless of the rest of its trace. The start allowance $\eta_kc_0(s)$ in \eqref{eq:adaptive_envelope} lowers the initial residual to $-\eta_k$ and vanishes at $s=1$, so terminal contraction is unchanged. It is part of the frozen envelope and enters the transient bound through $d_k$. A learned predictor may output $\eta_k\approx0$, since demonstrated chunks start exactly on their anchor. We therefore floor the allowance at a fixed fraction of $V_k$ (Appendix~\ref{sec:appendix_env_details}).

\subsection{Proof of \Cref{thm:objective_certification}}
\label{app:verification_proof}
On $[t_i,t_{i+1}]$ the Lipschitz bound from each endpoint gives $g_k(w,t)\le\min\{g_i+L_i(t-t_i),\,g_{i+1}+L_i(t_{i+1}-t)\}$. The first term increases and the second decreases in $t$, so their minimum is largest where they are equal, at the value $(g_i+g_{i+1}+L_i\Delta_i)/2$. Taking the maximum over the intervals gives \eqref{eq:interval_bound}, and $B_k(w)\le0$ is funnel satisfaction throughout $[0,H]$ by the definition of the residual.\hfill$\square$

\subsection{Convergence with a Terminal Error Allowance}
\label{app:practical_certification}
\begin{corollary}[Practical receding-horizon contraction]
    \label{cor:practical_contraction}
    Suppose executed chunks satisfy
    \[
        V(y_{k+1})\le(1-\gamma_k)V(y_k)+\epsilon_k,
        \qquad \gamma_k\in[\underline\gamma,1],\quad
        0\le\epsilon_k\le\bar\epsilon,
    \]
    where $\underline\gamma>0$ and $\bar\epsilon$ are uniform bounds. Then
    \begin{equation}
        V(y_{k+m})\le(1-\underline\gamma)^mV(y_k)
        +\bar\epsilon\sum_{j=0}^{m-1}(1-\underline\gamma)^j,
        \label{eq:practical_iterated_contraction}
    \end{equation}
    and $\limsup_{m\to\infty}V(y_{k+m})\le\bar\epsilon/\underline\gamma$.
\end{corollary}
\begin{proof}
    Nonnegativity of $V$ permits replacing $\gamma_k$ and $\epsilon_k$ by their uniform bounds. Iteration gives the geometric sum in \eqref{eq:practical_iterated_contraction}, whose limit yields the stated bound.
\end{proof}
The allowance can combine terminal residual violation and terminal prediction mismatch. Endpoint contraction needs only the last sample, which sampled verification already covers under matched full-chunk execution. Continuous-time satisfaction follows from \eqref{eq:verification_condition}. A whole-trajectory practical bound must also account for transient error.

\subsection{Notes on the Theorems}
\label{app:theorem_notes}
\paragraph{Learned additive envelopes and \Cref{thm:receding_horizon_contraction}.}
All profiles in \eqref{eq:adaptive_envelope} take values in $[0,1]$, so the learned envelope satisfies the transient bound of \eqref{eq:envelope} with
\begin{equation}
    \sup_{t\in[0,H]}\rho_k(t)\le V_k+\delta_k+\eta_k,
    \qquad d_k=\frac{\delta_k+\eta_k}{V_k}.
    \label{eq:learned_transient_bound}
\end{equation}
The predictor is a continuous network on a bounded feature domain with bounded allowance outputs, so $\delta_k+\eta_k\le\bar a$ for some constant $\bar a$, and
\begin{equation}
    d_k\le\bar d=\bar a/v_{\mathrm{goal}}\qquad\text{whenever }V_k\ge v_{\mathrm{goal}}>0.
    \label{eq:learned_dbar}
\end{equation}
An additive allowance $\delta_k+\eta_k$ corresponds to the relative allowance $(\delta_k+\eta_k)/V_k$, which grows as $V_k$ decreases. For example, a fixed allowance of 0.1 is 10\% of $V_k=1$ and 1000\% of $V_k=0.01$. The bound $\bar d=\bar a/v_{\mathrm{goal}}$ therefore holds on the region $V_k\ge v_{\mathrm{goal}}$, and \Cref{thm:receding_horizon_contraction} applies to additive funnels until $V$ reaches the positive level $v_{\mathrm{goal}}$. This scope matches our evaluation, which judges completion by the native success predicate of each task. The multiplicative family has a relative allowance by construction and satisfies the bounded-allowance condition at every level of $V$. If, in addition, $V(y)\ge\varphi_{\mathcal G}(\operatorname{dist}(y,\mathcal G))$ for a continuous, strictly increasing $\varphi_{\mathcal G}$ with $\varphi_{\mathcal G}(0)=0$, convergence of $V$ implies convergence in distance to $\mathcal G$.

\paragraph{Progress without a uniform rate.}
A learned funnel may request no endpoint progress for some chunks, $\gamma_k=0$, for instance while the robot regrasps or moves to the next object. The rate in \eqref{eq:continuous_task_convergence} then does not apply, because it assumes $\gamma_k\ge\underline\gamma>0$ at every iteration. The product bound in \eqref{eq:iterated_contraction} still holds for any $\gamma_k\in[0,1]$. The following proposition shows that convergence still holds when requested progress accumulates, either in total or within windows of $m$ chunks.
\begin{proposition}[Progress over windows]
    \label{prop:window_progress}
    Suppose every executed chunk satisfies the funnel \eqref{eq:lyapfunnel} with the envelope conditions \eqref{eq:envelope}, under the execution assumptions of \Cref{thm:receding_horizon_contraction}, and $d_k\le\bar d$ for all $k$.
    \begin{enumerate}
        \item For every $k$, $\sup_{t\in[0,H]}V(y^k_{w_k^\star}(t))\le(1+\bar d)\exp\bigl(-\sum_{j<k}\gamma_j\bigr)V(y_0)$. Hence, for any tolerance $\varepsilon\in(0,V(y_0))$, $V(y_k)\le\varepsilon$ as soon as $\sum_{j<k}\gamma_j\ge\log(V(y_0)/\varepsilon)$.
        \item If there are $m\ge1$ and $\Gamma>0$ such that $\sum_{j=k}^{k+m-1}\gamma_jV(y_j)\ge\Gamma$ whenever $V(y_k)>v_{\mathrm{goal}}$, then $V(y_k)\le v_{\mathrm{goal}}$ for all $k\ge m\lceil(V(y_0)-v_{\mathrm{goal}})/\Gamma\rceil$.
    \end{enumerate}
\end{proposition}
\begin{proof}
    (i) Since $1-\gamma\le e^{-\gamma}$, \eqref{eq:iterated_contraction} gives $V(y_k)\le\exp(-\sum_{j<k}\gamma_j)V(y_0)$, and within chunk $k$ the transient bound adds at most a factor $1+\bar d$.
    (ii) Terminal contraction gives $V(y_j)-V(y_{j+1})\ge\gamma_jV(y_j)\ge0$, so $V(y_k)$ is nonincreasing. Summing over a window that starts outside the goal set gives $V(y_k)-V(y_{k+m})\ge\Gamma$. At most $\lceil(V(y_0)-v_{\mathrm{goal}})/\Gamma\rceil$ such windows fit before the boundary value falls below $v_{\mathrm{goal}}$, and monotonicity keeps it there.
\end{proof}
Statement (i) needs no lower bound on the individual rates, only that their sum eventually exceeds the stated level. Statement (ii) admits chunks with $\gamma_k=0$, provided some progress is requested in every window of $m$ chunks. The predictor does not enforce this condition, and it can be checked along executed episodes. Both statements concern chunk boundaries. Within chunk $k$ the task potential stays below $(1+\bar d)V(y_k)$.

\paragraph{Local rates and their estimates.}
\label{app:local_rates}
\label{app:lse_excess}
In \Cref{thm:objective_certification}, $L_i$ bounds the rate of the residual on interval $i$. We estimate the rate of each interval from the largest nearby secant slope. The local rate $\hat L_i$ is the largest secant slope $|g_{j+1}-g_j|/\Delta_j$ among the intervals $j$ whose midpoints lie within a window $w_L$ of the midpoint of interval $i$, with the adjacent intervals always included. Substituting $\hat L_i$ into \eqref{eq:interval_bound} gives estimated verification, the verdict that the search uses for its patience and that the main text reports. The neighbouring intervals supply the rate information: the secant of interval $i$ alone returns $\max(g_i,g_{i+1})$ in \eqref{eq:interval_bound}, which is sampled verification. Since $\hat L_i$ is at least the secant of its own interval, a chunk that passes estimated verification also passes sampled verification. The estimate assigns a fast change of the residual to the intervals near where it occurs, and its resolution is set by the sample spacing and the window $w_L$. The guarantee of \Cref{thm:objective_certification} applies when the rates upper-bound the true residual rate on each interval, for instance with $L_Fv_{\max}+L_\rho$ from \Cref{lem:residual_regularity}.

\paragraph{Search objective and verification.}
The log-sum-exp $J_k$ of \Cref{sec:optimization} lies between $M_k$ and $M_k+\log(n)/\beta$, so $J_k\le0$ implies sampled verification. $J_k$ and the verdict serve separate purposes. $J_k$ aggregates the sampled residuals smoothly and ranks the candidates. The verdict of each plan is computed from its samples through \eqref{eq:interval_bound} with the local rates, which concern the regularity of the residual over physical time.

%% file: sections/appendix/additionalresults.tex
\section{Evaluation Statistics and Additional Results}
\label{sec:additional_results}

\subsection{Paired Bootstrap Intervals}
\label{app:search_statistics}
\label{app:statistics}
Each comparison uses matched episodes. For a given episode index, all compared configurations share the environment initial condition, the random seeds of the environment and the policy, and the initial flow-matching source draw. The funnel ablation uses $N=400$ episodes and the search comparison $N=200$.

For each of 4000 bootstrap replicates, we sample $N$ episode indices with replacement and use the same indices for both configurations. A draw such as $[2,2,5,1,\dots]$ selects those episodes, duplicates included, from both configurations. We recompute the metric for each configuration on the sampled episodes and take the difference from the reference configuration, which is endpoint-only in the funnel ablation and our search in the search comparison. The 2.5th and 97.5th percentiles of these differences give the 95\% percentile-bootstrap interval. We report it together with the point estimate, the difference of the metric between the two configurations on the original paired episodes, and call a difference significant when its interval excludes zero. Pairing retains the correspondence between starting conditions, so the variation in episode difficulty is shared by the two configurations. The intervals describe the uncertainty across the evaluated initial conditions for the reported policy and execution setup. Figure~\ref{fig:tail_tradeoff} and Table~\ref{tab:search_paired} report these intervals, and the shading of Tables~\ref{tab:search_ablation} and~\ref{tab:search_ablation_all} follows them, except for the \ac{OOD} column, which is shaded when the rate lies above the 95\% interval of the held-out demonstrations (Appendix~\ref{app:search_ablation}).

For tail metrics, each replicate identifies the lowest-scoring or largest-drawdown fraction of the resampled episodes separately for each configuration and averages that fraction before taking the difference. An episode can therefore belong to the tail of one configuration and to the bulk of the other, while both resampled datasets are built from the same episode indices.

\subsection{Metric Definitions}
\label{app:metrics}
\paragraph{Final-score $\text{\acs*{CVaR}}_{10}$.} The mean final score over the lowest-scoring 10\% of episodes. Higher is better.

\paragraph{Maximum drawdown and its $\text{\acs*{CVaR}}_{90}$.} The maximum drawdown of an episode is the largest fall of the score below its earlier peak, $\max_t(\max_{s\le t}r_s-r_t)$, in points ($\times100$). The trace $0.2,0.7,0.4,0.8$ has a drawdown of $0.3$, or 30 points. $\text{\acs*{CVaR}}_{90}$ is the mean drawdown over the 10\% of episodes with the largest drawdown, and $\text{\acs*{CVaR}}_{95}$ over the largest 5\%. Lower is better. The subscript is the quantile level at which the tail starts, so $\text{\acs*{CVaR}}_{10}$ of a score and $\text{\acs*{CVaR}}_{90}$ of a drawdown both average the worst 10\% of episodes.

\paragraph{\acs*{KL}.} $\text{\acs*{KL}}(p_{\max}\,\|\,p_{\mathrm{final}})$ compares the distribution of the peak score with that of the final score across episodes. Both are histograms with 20 bins over the joint range of the two scores and a pseudocount of 0.5. A lower value means that the episodes end close to their peak. \ac{KL} complements the drawdown at the level of distributions: it compares the peak and final scores across episodes, and the drawdown measures the loss within each episode.

\paragraph{Verification rate.} The share of complete executed chunks that pass estimated verification against the funnel of their configuration. The denominator counts the chunks executed over their full horizon $H$.

\paragraph{V-score.} On two-nut assembly, the V-score is $1-V/V_0$, with $V_0=85.3$ the potential at the reset state (Appendix~\ref{app:nut_results}).

\paragraph{Smoothness and similarity to the demonstrations.} Smoothness is the \ac{LDJ}~\citep{balasubramanianAnalysisMovementSmoothness2015} of the executed pusher path within chunks, where higher is smoother. Similarity to the demonstrations is the share of executed 24-step windows that are out of distribution under a nearest-neighbor test against the training demonstrations~\citep{sunOutofDistributionDetectionDeep2022}, with the settings of Appendix~\ref{app:search_ablation}.

\subsection{Complete Funnel Ablation}
\label{app:full_results}
Table~\ref{tab:full_results} extends Table~\ref{tab:main_results} to every configuration of the funnel ablation, with the mean final score and the verification of the nominal plans of the same planning calls. Plateaus at $\delta\in\{1,3\}$ allow $V$ to at least double inside a chunk, a level above the values that rollouts reach. Their results closely follow those of endpoint-only, and their equal sampled and estimated verification shares show that the endpoint is the binding condition (Appendix~\ref{app:verification_details}).

\begin{table}[t]
    \centering
    \caption{Funnel ablation, all configurations, 400 episodes per configuration. Shading: {\fboxsep=1pt\colorbox{green!40}{\textbf{best}}, \colorbox{green!22}{second} and \colorbox{green!9}{third}} refined configuration in each column. Verified: share of complete executed chunks (Executed) and of planning calls whose nominal plan (Nominal) passes estimated verification (Appendix~\ref{app:verification_details}).}
    \label{tab:full_results}
    \scriptsize
    \setlength{\tabcolsep}{1.6pt}
    \def\gsep{\hspace{5pt}}%
    \setlength{\aboverulesep}{0.8pt}\setlength{\belowrulesep}{0.8pt}
    \setlength{\extrarowheight}{1pt}
    \begin{tabular}{@{}ll@{\gsep}rr@{\gsep}rr@{\gsep}rr@{\gsep}rr@{\gsep}rr@{\gsep}rr@{\gsep}rr@{\gsep}rr@{}}
        \toprule
        & & \multicolumn{2}{c@{\gsep}}{Success} & \multicolumn{4}{c@{\gsep}}{Final score $\uparrow$} & \multicolumn{4}{c@{\gsep}}{Max drawdown [pts] $\downarrow$} & \multicolumn{2}{c@{\gsep}}{\acs*{KL}} & \multicolumn{4}{c}{Verified [\%]}\\
        \cmidrule(lr){3-4}\cmidrule(lr){5-8}\cmidrule(lr){9-12}\cmidrule(lr){13-14}\cmidrule(l){15-18}
        Funnel & & \multicolumn{2}{c@{\gsep}}{[\%] $\uparrow$} & \multicolumn{2}{c}{Mean} & \multicolumn{2}{c@{\gsep}}{$\text{\acs*{CVaR}}_{10}$} & \multicolumn{2}{c}{Mean} & \multicolumn{2}{c@{\gsep}}{$\text{\acs*{CVaR}}_{90}$} & \multicolumn{2}{c@{\gsep}}{$\downarrow$} & \multicolumn{2}{c}{Executed $\uparrow$} & \multicolumn{2}{c}{Nominal}\\
        & & {\tiny PushT} & {\tiny Stack.} & {\tiny PushT} & {\tiny Stack.} & {\tiny PushT} & {\tiny Stack.} & {\tiny PushT} & {\tiny Stack.} & {\tiny PushT} & {\tiny Stack.} & {\tiny PushT} & {\tiny Stack.} & {\tiny PushT} & {\tiny Stack.} & {\tiny PushT} & {\tiny Stack.}\\
        \midrule
        Nominal &  & 31.3 & 77.0 & .638 & .637 & .000 & .000 & 40.0 & 8.5 & 86.9 & 46.1 & .472 & .181 & \multicolumn{2}{c}{n/a} & \multicolumn{2}{c}{n/a}\\
        \addlinespace[1pt]
        Endpoint-only &  & \rkB{86.8} & \rkB{86.0} & .923 & \rkB{.804} & .560 & .084 & 28.3 & 6.8 & 62.8 & 46.1 & .053 & .060 & \multicolumn{2}{c}{n/a} & \multicolumn{2}{c}{n/a}\\
        \midrule[0.3pt]
        \multirow{2}{*}{No envelope} & $\dot V\le0$ & 60.5 & 85.5 & .855 & .744 & .218 & \rkC{.103} & 23.4 & \rkB{5.8} & 54.2 & \rkB{41.8} & .153 & .065 & \multicolumn{2}{c}{n/a} & \multicolumn{2}{c}{n/a}\\
         & $\Sigma V$ & 80.0 & 82.0 & .910 & .715 & .499 & .000 & \rkA{22.6} & \rkC{6.4} & 54.4 & \rkC{43.4} & \rkC{.052} & .084 & \multicolumn{2}{c}{n/a} & \multicolumn{2}{c}{n/a}\\
        \addlinespace[1.5pt]
        \multirow{3}{*}{Plateau} & $\delta=0.5$ & 73.0 & 83.8 & .905 & .771 & .494 & .075 & 23.9 & 6.9 & 56.7 & 46.2 & .098 & .065 & \rkA{73.3} & 41.9 & 48.7 & 40.0\\
         & $\delta=1$ & \rkC{86.5} & \rkA{86.5} & .920 & \rkC{.803} & .529 & .091 & 27.9 & 7.1 & 62.9 & 46.8 & .060 & \rkA{.045} & 68.8 & 40.9 & 45.8 & 35.4\\
         & $\delta=3$ & \rkB{86.8} & \rkC{85.8} & \rkC{.924} & \rkA{.805} & \rkC{.568} & .097 & 28.1 & 6.7 & 62.4 & 45.7 & .053 & \rkB{.054} & \rkC{69.0} & 40.6 & 46.2 & 35.4\\
        \addlinespace[1.5pt]
        Linear &  & 54.5 & 78.8 & .839 & .662 & .194 & .000 & 24.3 & 7.6 & 54.7 & 44.3 & .219 & .108 & 0.5 & 3.3 & 5.5 & 9.7\\
        \addlinespace[1.5pt]
        \multirow{2}{*}{Calibrated} & per stage & -- & 82.8 & -- & .792 & -- & .101 & -- & 7.6 & -- & 48.5 & -- & .102 & -- & \rkB{57.6} & -- & 47.2\\
         & global & \rkA{88.5} & 80.5 & \rkA{.927} & .717 & \rkB{.587} & .001 & 27.4 & 7.5 & 62.9 & 46.2 & \rkB{.038} & .107 & 60.5 & \rkA{61.2} & 15.4 & 52.5\\
        \addlinespace[1.5pt]
        \multirow{3}{*}{Learned add.} & $\alpha=0.5$ & 84.8 & 85.0 & \rkB{.925} & .799 & \rkA{.613} & \rkA{.162} & \rkB{22.9} & 7.6 & \rkB{52.3} & 47.0 & \rkA{.029} & \rkC{.056} & 38.0 & 44.3 & 7.5 & 23.6\\
         & $\alpha=0.8$ & 71.8 & 82.3 & .900 & .775 & .480 & .065 & \rkC{23.1} & 6.6 & 53.7 & 44.8 & .126 & .058 & 56.2 & 50.8 & 18.6 & 35.4\\
         & $\alpha=0.95$ & 74.3 & 82.3 & .902 & .767 & .474 & .025 & \rkA{22.6} & 7.0 & 52.6 & 44.8 & .126 & .101 & 62.9 & \rkC{54.3} & 24.6 & 41.9\\
        \addlinespace[1.5pt]
        \multirow{3}{*}{Learned mult.} & $\alpha=0.5$ & 76.5 & \rkC{85.8} & .910 & .796 & .542 & .095 & \rkC{23.1} & 7.0 & \rkC{52.5} & 46.3 & .169 & .060 & 40.5 & 42.1 & 12.2 & 23.5\\
         & $\alpha=0.8$ & 71.8 & 82.0 & .906 & .771 & .529 & .067 & 23.2 & 8.2 & \rkA{52.0} & 47.6 & .091 & .083 & 62.6 & 49.2 & 28.6 & 34.2\\
         & $\alpha=0.95$ & 63.8 & 84.0 & .887 & .775 & .424 & \rkB{.107} & 23.5 & \rkA{5.7} & 54.0 & \rkA{40.5} & .145 & \rkC{.056} & \rkB{70.7} & 54.2 & 33.1 & 43.6\\
        \bottomrule
    \end{tabular}
\end{table}

\input{sections/appendix/nutassembly.tex}

\FloatBarrier
\subsection{Lower Tails of the Outcome}
\label{app:outcome_tails}
\begin{figure}[!htbp]
    \centering
    \includegraphics[width=\linewidth]{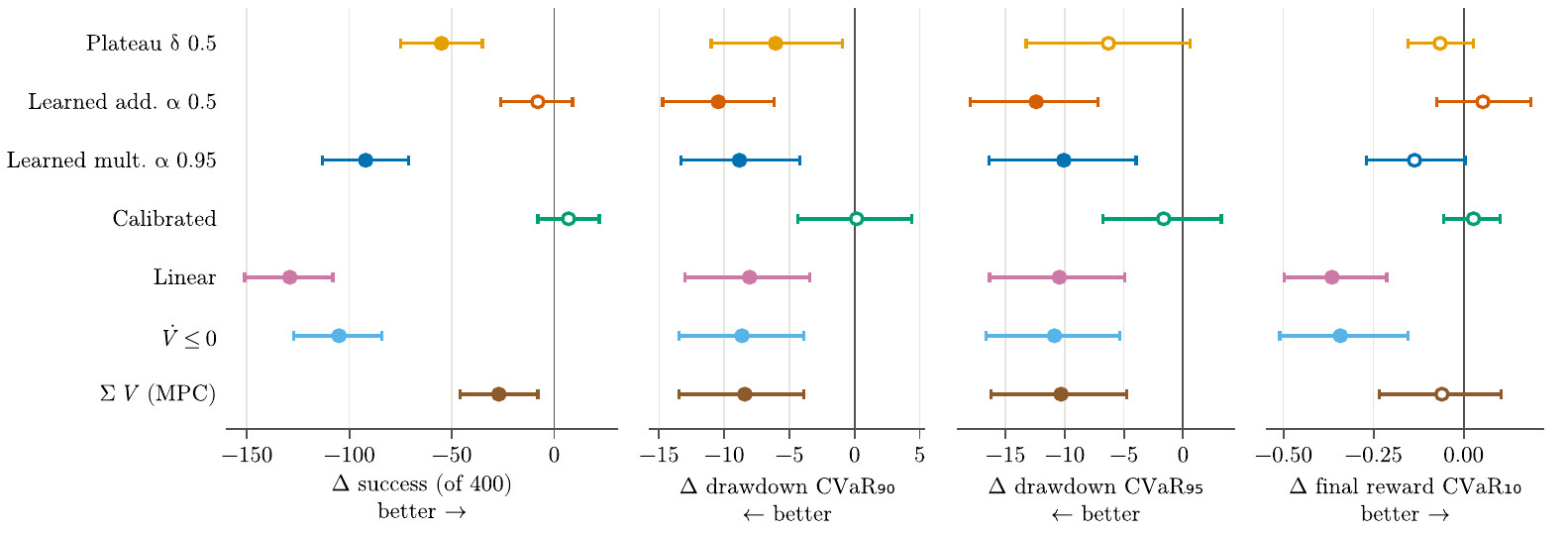}\\[2pt]
    {\footnotesize (a) PushT}\\[6pt]
    \includegraphics[width=\linewidth]{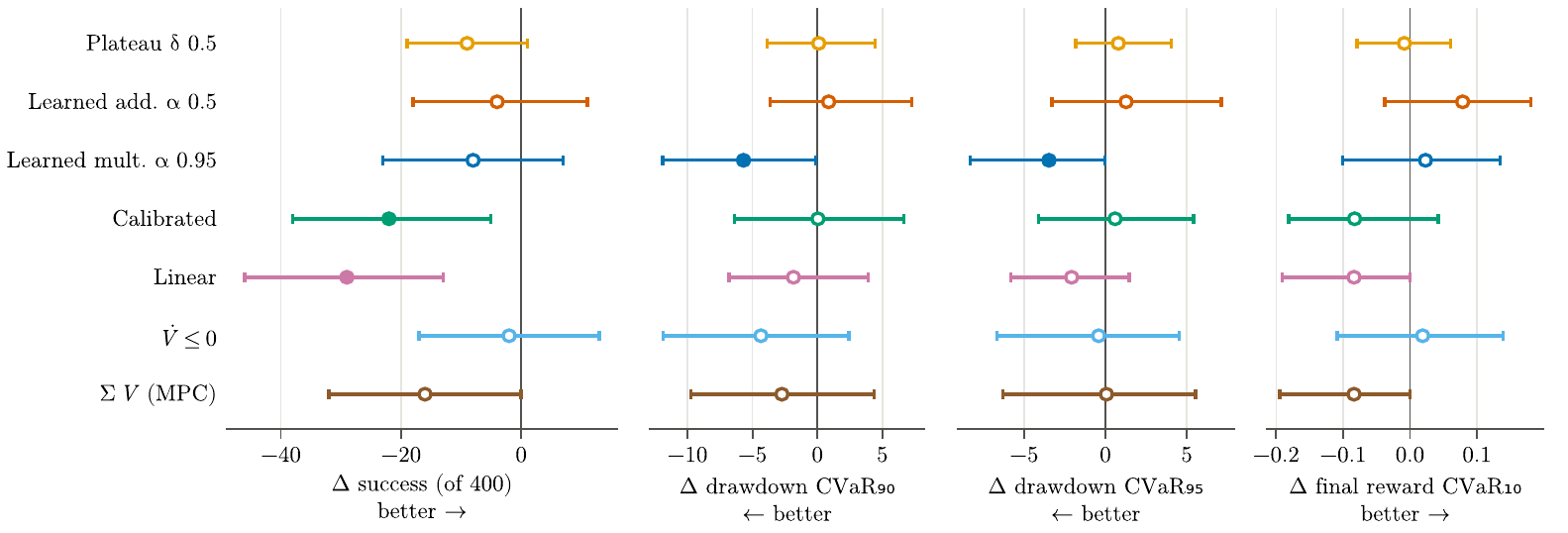}\\[2pt]
    {\footnotesize (b) Stacking}\\[6pt]
    \includegraphics[width=\linewidth]{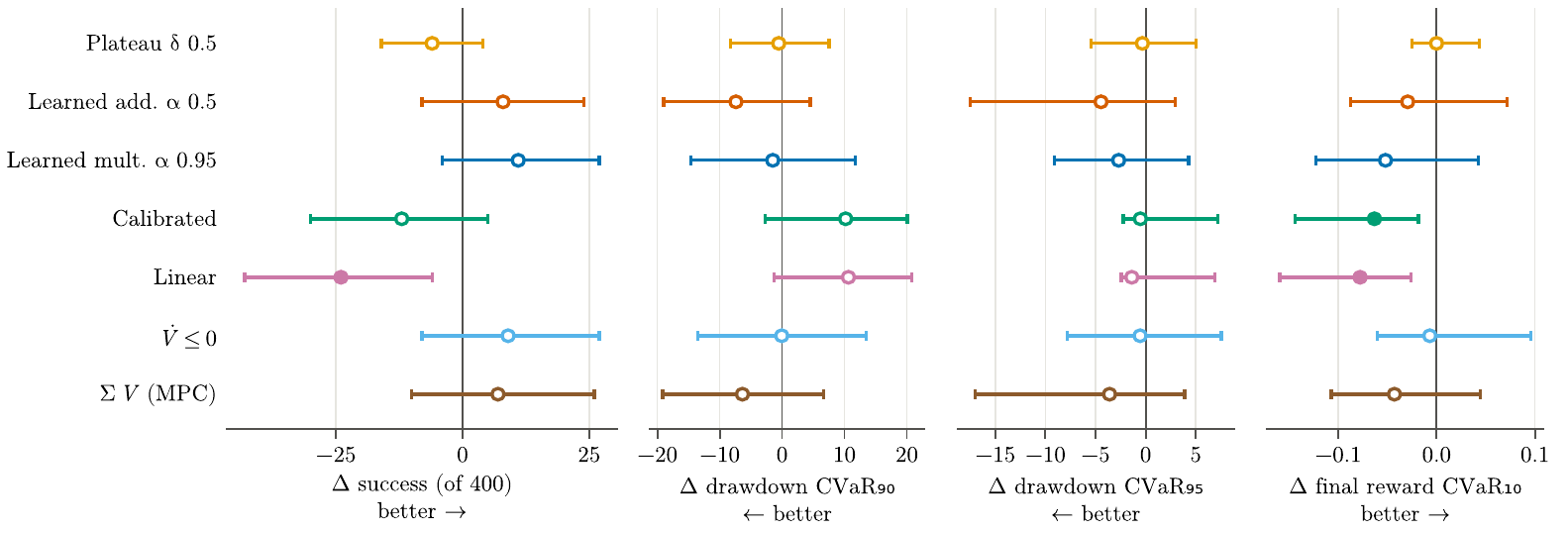}\\[2pt]
    {\footnotesize (c) Two-nut assembly}
    \caption{Paired differences from endpoint-only on the same 400 initial conditions: point estimate on the original paired episodes and 95\% percentile bootstrap interval; filled markers exclude zero. Success in episodes, drawdown in score points; on two-nut assembly, drawdown and final score are those of the V-score (Appendix~\ref{app:nut_results}). In the legends, Endpoint is endpoint-only, Calibrated the calibrated global funnel and $\Sigma V$ (\acs*{MPC}) the $\Sigma V$ objective; final reward is the final score.}
    \label{fig:tail_tradeoff}
\end{figure}

Figure~\ref{fig:tail_tradeoff} reports, for each objective of Table~\ref{tab:main_results}, the paired difference from endpoint-only on the same 400 initial conditions in success, in the $\text{\acs*{CVaR}}_{90}$ and $\text{\acs*{CVaR}}_{95}$ of the maximum drawdown, and in the $\text{\acs*{CVaR}}_{10}$ of the final score. On PushT, every objective except the calibrated funnel lowers the drawdown $\text{\acs*{CVaR}}_{90}$ significantly, by 6 to 10 points, and every one except the learned additive funnel at $\alpha=0.5$ also loses successes significantly, from 27 episodes for $\Sigma V$ to 129 for the linear funnel. The learned additive funnel at $\alpha=0.5$ lowers the $\text{\acs*{CVaR}}_{90}$ by 10.4 points and the $\text{\acs*{CVaR}}_{95}$ by 12, at a success statistically on par with endpoint-only. Lowering the drawdown tail is thus common to the objectives that constrain the interior, and this funnel combines it with the success of endpoint-only. On stacking, only the learned multiplicative funnel at $\alpha=0.95$ lowers the drawdown tail significantly, by 5.7 points in $\text{\acs*{CVaR}}_{90}$ and 3.5 in $\text{\acs*{CVaR}}_{95}$, at a success statistically on par with endpoint-only, while the calibrated global funnel and the linear funnel lose successes significantly. On two-nut assembly, the confidence intervals for the V-score drawdown tail against endpoint-only all contain zero. The learned additive funnel at $\alpha=0.5$ and $\Sigma V$ have the largest reductions of its $\text{\acs*{CVaR}}_{90}$, 7.4 and 6.4 points, and the calibrated funnel and the linear funnel the largest increases, about 10 points. Both of the latter lower the $\text{\acs*{CVaR}}_{10}$ of the final V-score significantly, and the linear funnel also loses 24 episodes. Among the configurations not shown, only the plateau at $\delta=3$ gains successes significantly (Appendix~\ref{app:nut_results}). Refinement leaves the motion as smooth as endpoint-only does, with the log dimensionless jerk of executed chunks within $0.2$ of endpoint-only's for every funnel on PushT and stacking, and within $0.13$ on two-nut assembly.

\begin{figure}[!htbp]
    \centering
    \begin{minipage}[b]{0.40\linewidth}
        \centering
        \includegraphics[height=0.95\linewidth]{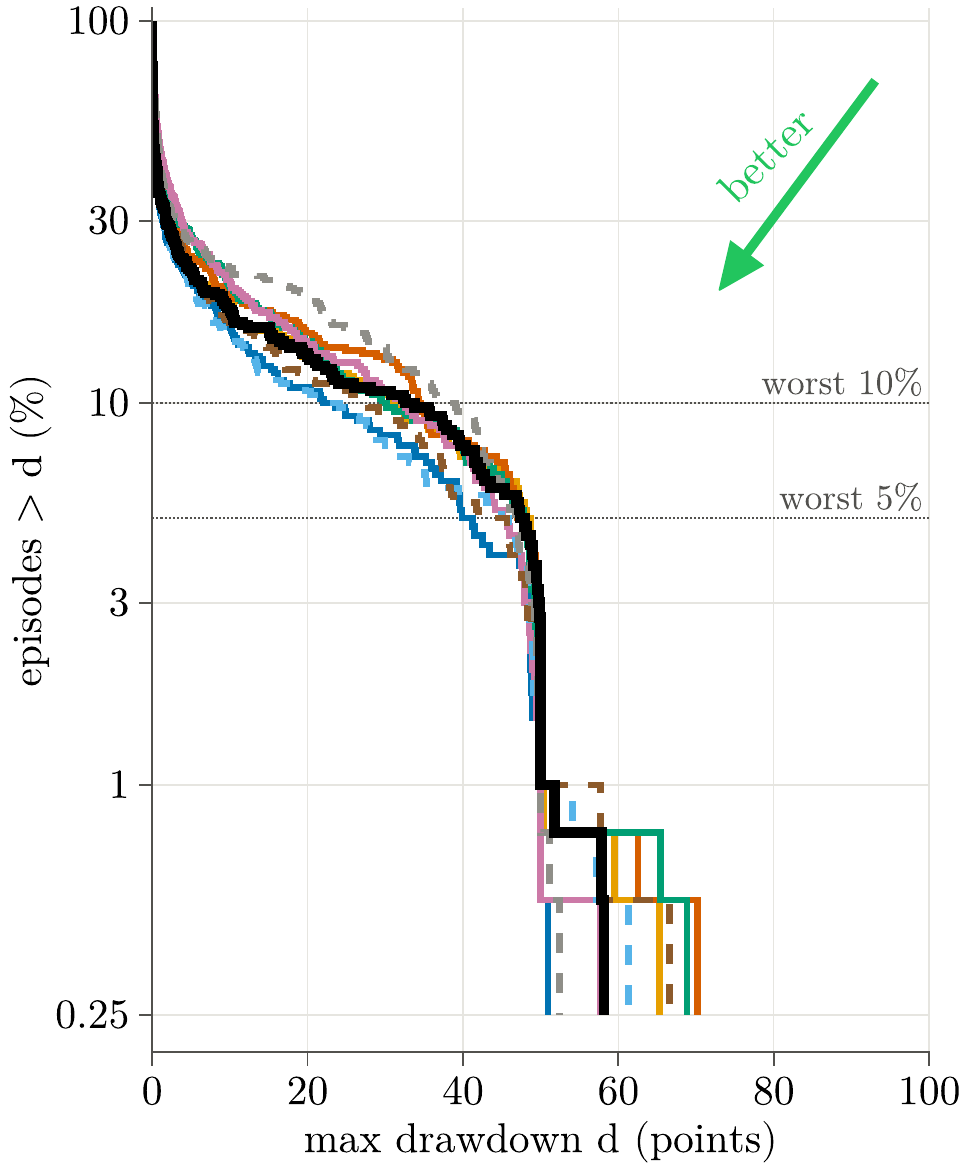}
        \caption{Stacking: share of episodes whose maximum drawdown exceeds $d$, as for PushT in Table~\ref{tab:main_results} (right).}
        \label{fig:drawdown_exceedance_stack}
    \end{minipage}\hfill
    \begin{minipage}[b]{0.54\linewidth}
        \centering
        \includegraphics[height=0.7037\linewidth]{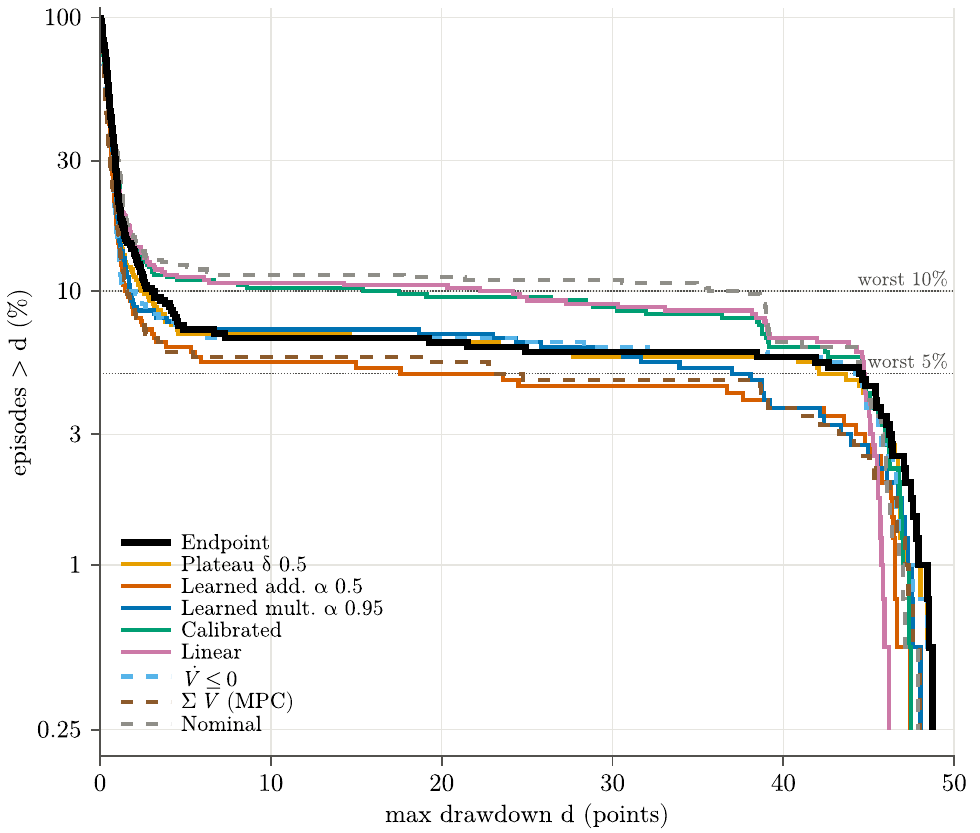}
        \caption{Two-nut assembly: share of episodes whose maximum V-score drawdown exceeds $d$.}
        \label{fig:drawdown_exceedance_nut}
    \end{minipage}
\end{figure}

\begin{figure}[!htbp]
    \centering
    \begin{minipage}[b]{0.49\linewidth}
        \centering
        \includegraphics[width=\linewidth]{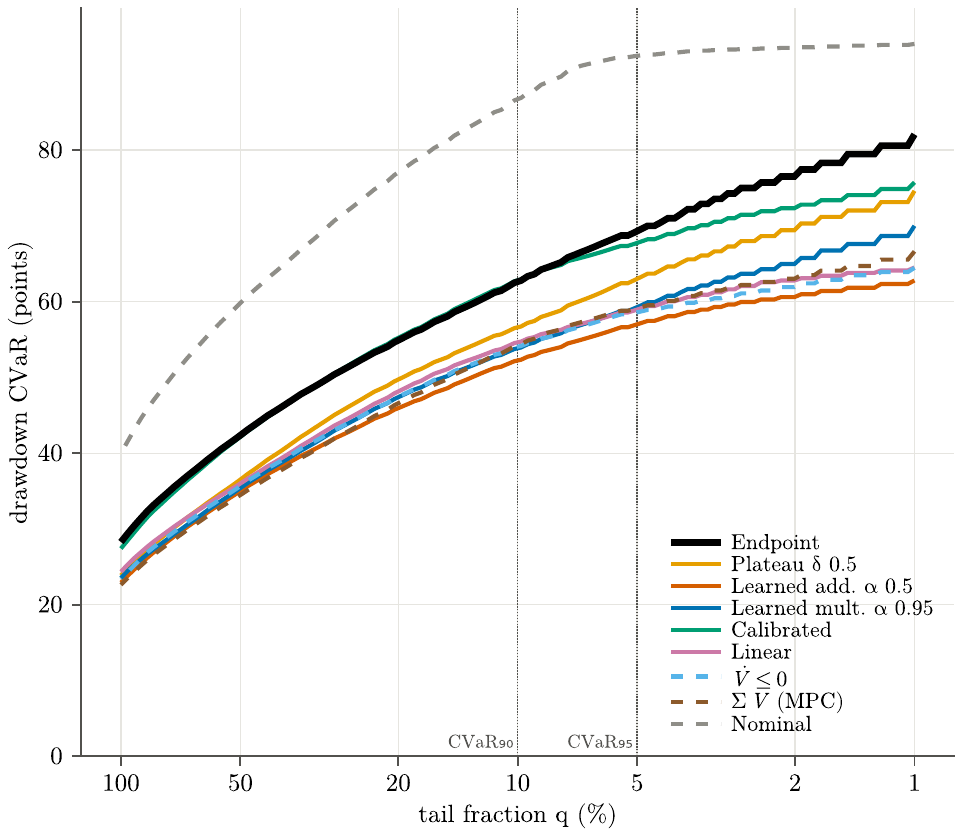}\\[2pt]
        {\footnotesize (a) PushT}
    \end{minipage}\hfill
    \begin{minipage}[b]{0.49\linewidth}
        \centering
        \includegraphics[width=\linewidth]{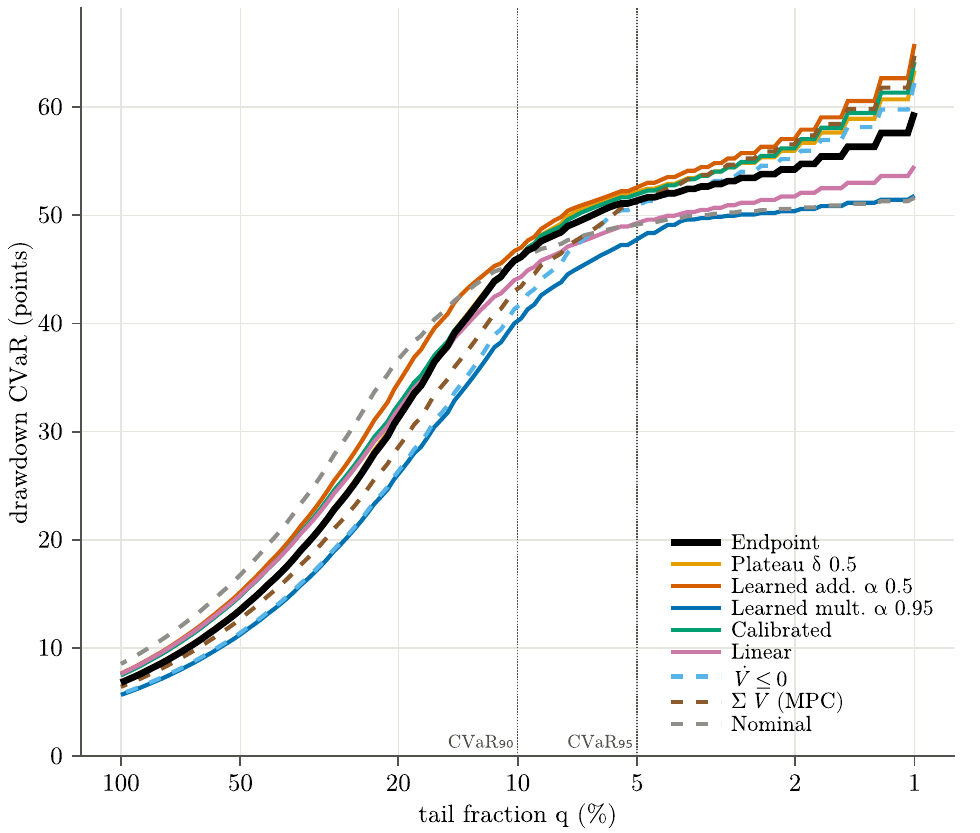}\\[2pt]
        {\footnotesize (b) Stacking}
    \end{minipage}\\[6pt]
    \begin{minipage}[b]{0.49\linewidth}
        \centering
        \includegraphics[width=\linewidth]{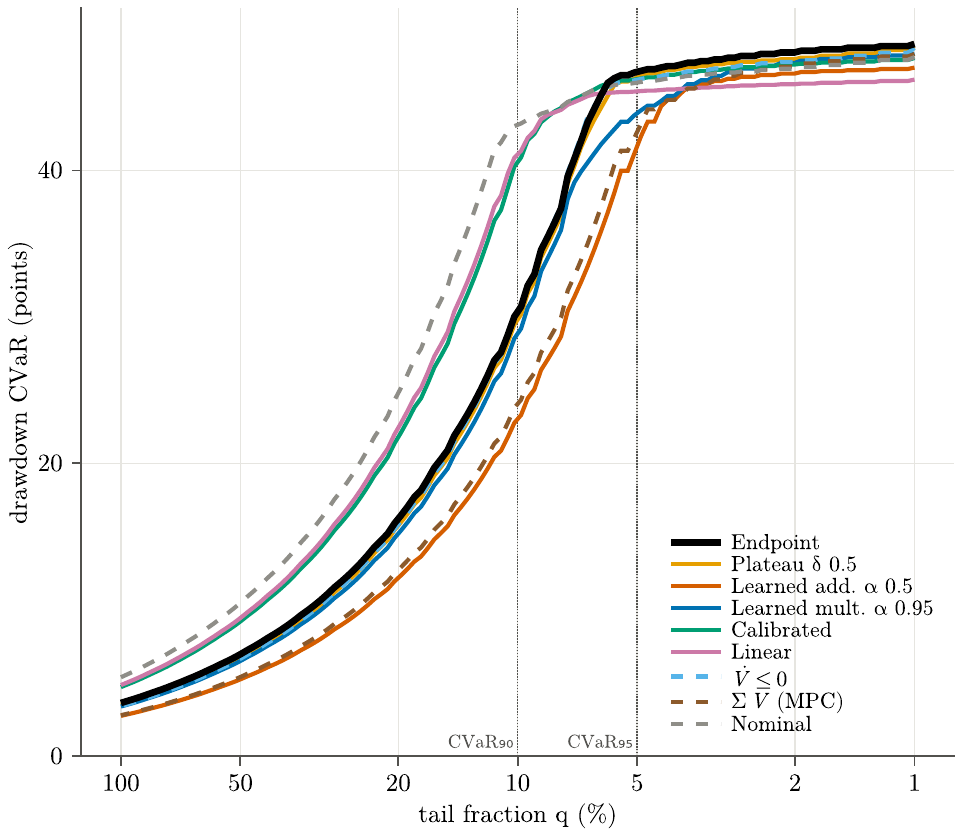}\\[2pt]
        {\footnotesize (c) Two-nut assembly}
    \end{minipage}
    \caption{Mean maximum drawdown of the worst $q$\% of episodes, $\text{\acs*{CVaR}}$ at tail fraction $q$, for the objectives of Table~\ref{tab:main_results}; the dotted lines mark $\text{\acs*{CVaR}}_{90}$ and $\text{\acs*{CVaR}}_{95}$. On two-nut assembly (c) the drawdown is that of the V-score. In the legends, Endpoint is endpoint-only, Calibrated the calibrated global funnel and $\Sigma V$ (\acs*{MPC}) the $\Sigma V$ objective.}
    \label{fig:drawdown_tails}
\end{figure}

Figures~\ref{fig:drawdown_exceedance_stack}--\ref{fig:drawdown_tails} show the drawdown at every tail fraction, of which Table~\ref{tab:main_results} reports only the mean and $\text{\acs*{CVaR}}_{90}$. On stacking (Figure~\ref{fig:drawdown_exceedance_stack}), most episodes give back almost nothing, and every configuration, nominal execution included, has a step just below 50 points, the drawdown of an episode that loses the top placement after reaching it. Refinement changes the number of such episodes and leaves the size of the loss unchanged. The learned multiplicative funnel and $\dot V\le0$ lie lowest between about 10 and 45 points, which is where their lower $\text{\acs*{CVaR}}_{90}$ comes from. On PushT (Figure~\ref{fig:drawdown_tails}a), nominal execution lies far above every refined configuration at every tail fraction. Endpoint-only is the highest refined curve throughout, the calibrated global funnel nearly coincides with it down to the worst 10\%, and every objective that constrains the interior of the chunk sits about 8 to 10 points below it over the whole tail, with the learned additive funnel lowest in the extreme tail. On stacking (Figure~\ref{fig:drawdown_tails}b), the configurations differ by a few points. The learned multiplicative funnel and $\dot V\le0$ stay lowest from the worst half of the episodes down to about the worst 5\%. Beyond that, each curve rests on a handful of episodes (4 at $q=1\%$), too few to order the configurations. On two-nut assembly (Figures~\ref{fig:drawdown_exceedance_nut} and~\ref{fig:drawdown_tails}c), most episodes give back only a few points, and the configurations separate by how many episodes later lose a large part of their progress: about 10\% under nominal execution, the calibrated funnel and the linear funnel, about 7\% under endpoint-only and most other objectives, and 5 to 6\% under the learned additive funnel at $\alpha=0.5$ and $\Sigma V$, which lie lowest from the worst 50\% down to about the worst 5\% (Figure~\ref{fig:drawdown_tails}c). Beyond that, every curve rests on drawdowns of 40 to 48 points, close to the 45 points of $V_0$ that the square nut accounts for at rest, and the curves lie too close to order the configurations.

\subsection{Choice of the Potential}
\label{app:potential_ablation}
We ask how much task structure the potential needs. On stacking we refine against three potentials, each with a fixed additive funnel and with its own learned additive funnel at $\alpha=0.95$ of matched size, under the search of Table~\ref{tab:search_settings}. The terminal-tolerance potential is a simple potential that scores the final geometry of the two placements. The two phased potentials follow the construction of Appendix~\ref{app:phased_lyapunov} and guide the lift, the carry and the transition between the placements. The first is the potential of the main experiments, \eqref{eq:stack_potential}, which builds both placements from the residual $\phi$ of \eqref{eq:placement_residual} with its finer lowering bands and alignment basin. The second builds both placements from the gate residual $\phi^{\mathrm g}$, which has wider bands and no alignment basin. The ablation uses the 400 initial conditions and seeds of the main funnel ablation, and the learned-funnel configuration of the residual-$\phi$ potential is the one reported in Table~\ref{tab:full_results}.
\begin{table}[h]
    \centering
    \caption{Potential ablation on three-cube stacking, \acs*{RMFM} policy, with a fixed additive funnel and a learned additive funnel at $\alpha=0.95$ of matched size, on the 400 initial conditions and seeds of the main funnel ablation. The phased residual-$\phi$ potential is the potential of the main experiments, and its learned-funnel entries are those of Table~\ref{tab:full_results}. Success in percent of initial conditions. Drawdown statistics, in score points, are those of the learned-funnel configurations.}
    \label{tab:potential_ablation}
    \footnotesize
    \begin{tabular}{@{}lccccc@{}}
        \toprule
        & \multicolumn{2}{c}{Success [\%] $\uparrow$} & \multicolumn{3}{c}{Max drawdown [pts], learned funnel $\downarrow$}\\
        \cmidrule(lr){2-3}\cmidrule(lr){4-6}
        Potential & Fixed funnel & Learned funnel & Mean & Median & $\text{\acs*{CVaR}}_{90}$\\
        \midrule
        Nominal & \multicolumn{2}{c}{$77.0$} & $8.5$ & $0.75$ & $46.1$\\
        Terminal tolerance & $63.0$ & $50.0$ & $13.4$ & $6.4$ & $42.9$\\
        Phased, gate residual $\phi^{\mathrm g}$ & $80.5$ & $86.0$ & $6.0$ & $0.31$ & $43.0$\\
        Phased, residual $\phi$ (main) & $74.5$ & $82.3$ & $7.0$ & $0.51$ & $44.8$\\
        \bottomrule
    \end{tabular}
\end{table}

The simple potential alone leaves refinement below nominal execution. With the terminal-tolerance potential, refinement reaches 63.0\% success with the fixed funnel and 50.0\% with the learned funnel, compared with 77.0\% for nominal execution, and its mean drawdown under the learned funnel is 13.4 points, compared with 8.5 (Table~\ref{tab:potential_ablation}). With both phased potentials, refinement improves on nominal execution. With the learned funnel, the residual-$\phi$ potential of the main experiments reaches 82.3\% success and the gate-residual potential 86.0\%, with mean drawdowns of 7.0 and 6.0 points. The learned funnel adds 7.8 and 5.5 points of success over the fixed funnel for the two potentials. These results support encoding the intermediate task structure in the potential: one that follows the task along the way lets the search select motions that advance it. Within the phased construction, the gate residual, with wider bands and no alignment basin, performs at least as well as the residual $\phi$, so this coarser residual is sufficient on this task.

\subsection{Verification Details}
\label{app:verification_details}
A funnel with an interior envelope gives a condition on the transient that can be checked on each chunk. Verification applies to the configurations with an interior envelope, and endpoint-only is covered by its endpoint condition. For these configurations, Table~\ref{tab:verification_full} reports the share of complete executed chunks that pass sampled verification and estimated verification. As a reference, it also reports the share of planning calls whose nominal plan passes estimated verification on its predicted trace. Estimated verification uses the window $w_L=0.02$\,s on the physics-step grid of the rollout.

Estimated verification is stricter than sampled verification by up to 16 points on PushT and 27 points on stacking, which shows how much the bound between samples contributes to the verdict, and the difference grows with the leniency of the learned funnels. For the plateaus the two shares agree to within a point, since their envelope lies far above $V$ inside the chunk and the endpoint is the binding condition. Under the learned funnels, the executed chunks pass estimated verification more often than the nominal plans of the same calls, by 28 to 38 points on PushT and by 11 to 21 points on stacking. Under the linear funnel, few candidates meet the constraint, and the executed share is 0.5\% on PushT and 3.3\% on stacking. The nominal share counts every planning call on its predicted trace, including the last chunk of each episode, and the executed share counts complete executed chunks.

Two-nut assembly follows the pattern of stacking. The difference between sampled and estimated verification reaches 53 points for the calibrated global funnel, and under the learned funnels the executed chunks pass estimated verification more often than the nominal plans, by 5 to 17 points. The sampled and estimated shares of the plateaus coincide, as on the other tasks.

\begin{table}[h]
    \centering
    \caption{Verification against each configuration's own funnel, in percent of chunks, U-Net-\acs*{OT} on PushT and \acs*{DiT}-\acs*{RMFM} on stacking and two-nut assembly. For the plateaus only the endpoint binds, so their sampled and estimated shares coincide. Executed: complete executed chunks. Nominal: planning calls whose nominal plan passes estimated verification on its predicted trace, counting every call, including the last, cut-short chunk of each episode.}
    \label{tab:verification_full}
    \footnotesize
    \setlength{\tabcolsep}{4pt}
    \begin{tabular}{@{}llccc@{}}
        \toprule
        & & \multicolumn{2}{c}{Executed} & Nominal\\
        \cmidrule(lr){3-4}
        Task & Funnel & Sampled & Estimated & Estimated\\
        \midrule
        PushT & Plateau, $\delta=0.5$ & $73.3$ & $73.3$ & $48.7$\\
         & Plateau, $\delta=1$ & $68.9$ & $68.8$ & $45.8$\\
         & Plateau, $\delta=3$ & $69.0$ & $69.0$ & $46.2$\\
         & Linear & $15.5$ & $0.5$ & $5.5$\\
         & Calibrated, global & $71.2$ & $60.5$ & $15.4$\\
         & Learned add., $\alpha=0.5$ & $44.1$ & $38.0$ & $7.5$\\
         & Learned add., $\alpha=0.8$ & $64.2$ & $56.2$ & $18.6$\\
         & Learned add., $\alpha=0.95$ & $78.7$ & $62.9$ & $24.6$\\
         & Learned mult., $\alpha=0.5$ & $42.6$ & $40.5$ & $12.2$\\
         & Learned mult., $\alpha=0.8$ & $65.0$ & $62.6$ & $28.6$\\
         & Learned mult., $\alpha=0.95$ & $77.8$ & $70.7$ & $33.1$\\
        \midrule
        Stacking & Plateau, $\delta=0.5$ & $41.9$ & $41.9$ & $40.0$\\
         & Plateau, $\delta=1$ & $40.9$ & $40.9$ & $35.4$\\
         & Plateau, $\delta=3$ & $40.6$ & $40.6$ & $35.4$\\
         & Linear & $11.3$ & $3.3$ & $9.7$\\
         & Calibrated, per stage & $84.1$ & $57.6$ & $47.2$\\
         & Calibrated, global & $85.1$ & $61.2$ & $52.5$\\
         & Learned add., $\alpha=0.5$ & $55.0$ & $44.3$ & $23.6$\\
         & Learned add., $\alpha=0.8$ & $73.3$ & $50.8$ & $35.4$\\
         & Learned add., $\alpha=0.95$ & $81.1$ & $54.3$ & $41.9$\\
         & Learned mult., $\alpha=0.5$ & $51.5$ & $42.1$ & $23.5$\\
         & Learned mult., $\alpha=0.8$ & $71.0$ & $49.2$ & $34.2$\\
         & Learned mult., $\alpha=0.95$ & $81.6$ & $54.2$ & $43.6$\\
        \midrule
        Two-nut & Plateau, $\delta=0.5$ & $32.5$ & $32.5$ & $32.9$\\
        assembly & Plateau, $\delta=1$ & $31.5$ & $31.5$ & $32.2$\\
         & Plateau, $\delta=3$ & $32.7$ & $32.7$ & $33.5$\\
         & Linear & $14.2$ & $1.6$ & $13.9$\\
         & Calibrated, per stage & $96.3$ & $62.2$ & $60.0$\\
         & Calibrated, global & $97.1$ & $44.1$ & $44.2$\\
         & Learned add., $\alpha=0.5$ & $67.2$ & $37.4$ & $20.2$\\
         & Learned add., $\alpha=0.8$ & $77.7$ & $41.2$ & $30.9$\\
         & Learned add., $\alpha=0.95$ & $86.9$ & $44.0$ & $39.3$\\
         & Learned mult., $\alpha=0.5$ & $59.6$ & $38.9$ & $21.8$\\
         & Learned mult., $\alpha=0.8$ & $73.4$ & $41.5$ & $31.1$\\
         & Learned mult., $\alpha=0.95$ & $86.2$ & $44.7$ & $39.5$\\
        \bottomrule
    \end{tabular}
\end{table}

\subsection{Search Method Comparison}
\label{app:search_ablation}
\paragraph{Compared methods.}
All methods of Table~\ref{tab:search_ablation_all} refine the same policy, the PushT U-Net of Appendix~\ref{app:policy_training} used in the funnel ablation, against the same objective, the plateau funnel at $\delta=1$ with $\eta=0.1$, $\gamma=0.05$ and $\beta=10$ on the control grid, and run on 200 shared initial conditions with the hard reset of Table~\ref{tab:environment_settings}. The population searches score every candidate by a rollout in the native simulator and use a population of 20. Plain \ac{CMA-ES} samples the seed without projection. \ac{CMA-ES} on the band, \ac{CEM} (elite fraction 0.2, smoothing 0.1, minimum standard deviation 0.05), \ac{MPPI} (temperature 0.1) and best-of-$N$ use the typical-set projection of \Cref{sec:optimization} unchanged. Every sampled seed is mapped by $\Psi$ of \eqref{eq:seed_shell}, with one global group, before it is decoded, and the search mean is projected after each update. \ac{CEM}, \ac{MPPI}, best-of-$N$ and ours use $\zeta=4$, and two further configurations of ours use the tighter bands $\zeta\in\{2,3\}$. The sphere configuration fixes both radii of the band to the typical radius $\sqrt\nu$ of \eqref{eq:source_shell_radii}, the projection of standardized \ac{CMA-ES}~\citep{stolzlePhysibleMDMEnforcingInferenceTime2026}. Best-of-$N$ decodes one batch of 200 projected draws, the nominal seed first, and keeps the best. The gradient searches of Appendix~\ref{app:surrogate_gradient} move the seed by their own steps and are not projected. Every iterative search runs on a time budget of 1.5\,s per chunk, with patience and the generation cap disabled. The clock is checked after every generation or iteration, so a search exceeds it by at most one. A gradient search can still stop earlier, on a stationary gradient, a failed line search, or a plan that never reaches the block.

\paragraph{Similarity to the demonstrations.}
We measure how closely the executed pusher paths of a search resemble the demonstrated ones. Executed pusher paths are cut into 24-step windows at the chunk boundaries, expressed relative to the pusher position at the start of the window, and projected on a 32-component \ac{PCA} of the windows of the training demonstrations. A window is out of distribution when the mean distance to its 5 nearest training windows exceeds the 95th percentile of the same distance over held-out demonstration windows, and \ac{KID}~\citep{binkowskiDemystifyingMMDGANs2018} is computed against the training windows in the same space. Held-out demonstrations, cut at the chunk boundaries in the same way, have an \ac{OOD} rate of 9.6\% (95\% interval 5.4--15.4\%). Only unconstrained \ac{CMA-ES} exceeds this interval. It also has a \ac{KID} of 1.39, against 0.14--0.26 for every other method, nominal execution included. Among the band widths, only the sphere separates from $\zeta=4$ in \ac{OOD} rate. \ac{KID} has wide episode-bootstrap intervals (0.08--0.44 for ours) and does not separate the other methods. The commanded waypoints give the same ranking (Table~\ref{tab:search_ablation_full}). The similarity scores and \ac{LDJ} average over the chunks that each method executes, and successful methods end sooner, so the methods average different stretches of the task. Nominal execution therefore has a higher \ac{KID} than the searches despite its lowest \ac{OOD} rate: its windows include many more stretches of long failing episodes.

\paragraph{Outcome and smoothness.}
Table~\ref{tab:search_paired} gives the paired intervals behind Table~\ref{tab:search_ablation_all}. Here we add the comparisons that the main text does not discuss. Unconstrained \ac{CMA-ES} is also significantly better than ours in the mean final score and the mean \ac{DD}, and the sphere has a significantly lower mean final score. No search differs significantly from ours in the drawdown tail or in \ac{KL}. Within chunks, unconstrained \ac{CMA-ES} is the least smooth search (\ac{LDJ} $-1.03$ against ours), followed by \ac{CEM} and the gradient searches, while the sphere ($+0.31$) and the band at $\zeta=2$ ($+0.15$) are significantly smoother. At chunk boundaries, unconstrained \ac{CMA-ES}, the gradient searches, \ac{CEM} and \ac{MPPI} jump significantly more than ours, the sphere does not differ from it, and best-of-$N$ jumps significantly less ($-0.049$). At $\zeta=3$ the final score is slightly but significantly lower, in mean ($-0.005$) and $\text{\acs*{CVaR}}_{10}$ ($-0.022$). That interval is narrow because both methods end the same episode with score zero, which cancels in the paired resamples. The time to finish separates the methods more clearly than the mean score. An episode stops at its first success, when the score first exceeds 0.95, and an episode that approaches the threshold runs to its step budget with a similar score. On the band, the maximum score and the mean drawdown of all methods lie within 0.01 and 1.3 points of ours (Table~\ref{tab:search_ablation_full}). Over the episodes that each method solves, resampled jointly with ours, the narrower bands need 17 more steps ($\zeta=3$: $[+5,+28]$; $\zeta=2$: $[+7,+28]$), the sphere 38 ($[+26,+50]$) and best-of-$N$ 77 ($[+63,+91]$). This average is taken over the solved episodes of each method. Counting failures at the 400-step limit, the episodes of best-of-$N$ run 73 steps longer than ours ($[+55,+90]$) and those of the sphere 37 ($[+18,+55]$). Best-of-$N$ and the sphere therefore combine the most demonstration-like motion with longer times to completion, at a similar score and drawdown.

\paragraph{Gradient searches.}
Beyond success, \ac{L-BFGS} and Adam are significantly worse than ours on every score measure. The model's per-chunk error is small and compounds over contact switches. Adam, whose step moves each seed coordinate by at most about $0.05$ per iteration, stays close to the demonstrations (5.8\% \ac{OOD}) but solves the fewest episodes. A derivative-free search also applies unchanged to simulators that cannot be differentiated.

\begin{table}[h]
    \centering
    \caption{Search method comparison on PushT, all methods and measures. {\fboxsep=1pt\colorbox{green!25}{Better}/\colorbox{red!20}{Worse}} than {\fboxsep=1pt\colorbox{blue!10}{ours}} (significant, Appendix~\ref{app:search_statistics}), or \acs*{OOD} above held-out demonstrations. Score: mean of the maximum and of the final score. \acs*{DD}: drawdown mean and $\text{\acs*{CVaR}}_{90}$. Jump: relative velocity jump at chunk boundaries. Executed and Planned: similarity to the demonstrations of the executed path and of the commanded waypoints. $^\dagger$Gradient search on the differentiable model.}
    \label{tab:search_ablation_all}
    \label{tab:search_ablation_full}
    \scriptsize
    \setlength{\tabcolsep}{2pt}
    \resizebox{\linewidth}{!}{%
    \begin{tabular}{@{}lccccccccccccc@{}}
        \toprule
        & & & \multicolumn{2}{c}{Score} & \multicolumn{2}{c}{\acs*{DD} [pts]} & & & & \multicolumn{2}{c}{Executed} & \multicolumn{2}{c}{Planned}\\
        \cmidrule(lr){4-5}\cmidrule(lr){6-7}\cmidrule(lr){11-12}\cmidrule(l){13-14}
        Search & Succ.$\uparrow$ & $\text{\acs*{CVaR}}_{10}\!\uparrow$ & Max & Final & Mean & $\text{\acs*{CVaR}}_{90}$ & \acs*{KL} & \acs*{LDJ}$\uparrow$ & Jump & \acs*{OOD}\%$\downarrow$ & \acs*{KID} & \acs*{OOD}\% & \acs*{KID}\\
        \midrule
        Nominal & \sgW{31.5} & \sgW{.000} & .795 & .688 & 33.3 & 67.5 & .363 & $-9.92$ & .544 & 4.0 & .259 & 3.2 & .286\\
        Adam$^\dagger$ & \sgW{60.5} & \sgW{.647} & .927 & .918 & 19.1 & 51.6 & .045 & \sgW{$-10.12$} & .652 & 5.8 & .162 & 4.5 & .181\\
        \acs*{L-BFGS}$^\dagger$ & \sgW{70.0} & \sgW{.765} & .942 & .938 & 19.5 & 47.4 & .012 & \sgW{$-10.15$} & .662 & 12.0 & .196 & 10.7 & .154\\
        Best-of-$N$ & 82.5 & \sgW{.768} & .948 & .940 & 19.9 & 50.5 & .054 & $-9.87$ & .513 & 4.9 & .175 & 4.0 & .194\\
        \acs*{CEM} & 79.0 & .848 & .954 & .950 & 18.7 & 47.3 & .014 & \sgW{$-10.28$} & .600 & 8.0 & .143 & 6.2 & .141\\
        \acs*{MPPI} & 83.0 & .914 & .960 & .958 & 18.5 & 46.8 & .041 & $-9.97$ & .606 & 10.8 & .224 & 9.8 & .177\\
        \acs*{CMA-ES}, no band & \sgB{96.0} & \sgB{.941} & .963 & .963 & 15.3 & 45.9 & .022 & \sgW{$-10.96$} & .682 & \sgW{34.1} & 1.39 & 32.5 & .996\\
        \acs*{CMA-ES}, sphere & 84.5 & \sgW{.794} & .949 & .944 & 19.2 & 49.5 & .033 & \sgB{$-9.62$} & .557 & 6.6 & .166 & 5.1 & .152\\
        Ours, $\zeta{=}2$ & 87.0 & .873 & .957 & .955 & 18.0 & 45.3 & .012 & \sgB{$-9.78$} & .558 & 9.5 & .217 & 7.5 & .180\\
        Ours, $\zeta{=}3$ & 83.0 & \sgW{.855} & .954 & .951 & 18.3 & 50.3 & .008 & $-9.91$ & .563 & 9.4 & .177 & 8.2 & .155\\
        \oursrow Ours, $\zeta{=}4$ & 86.5 & .877 & .957 & .956 & 18.6 & 47.0 & .012 & $-9.93$ & .562 & 11.3 & .220 & 9.1 & .163\\
        \bottomrule
    \end{tabular}}
\end{table}

\newcommand{\pd}[2]{\begin{tabular}[c]{@{}c@{}}#1\\[-1pt]{\tiny[#2]}\end{tabular}}
\begin{table}[h]
    \centering
    \caption{Paired differences from our search (method minus ours) on the same 200 initial conditions: point estimate on the original paired episodes and 95\% percentile bootstrap interval, in bold when the interval excludes zero. Success and drawdown in points. Positive is better for success, score and \acs*{LDJ}, and worse for drawdown, \acs*{KL} and Jump.}
    \label{tab:search_paired}
    \scriptsize
    \setlength{\tabcolsep}{1.2pt}
    \renewcommand{\arraystretch}{1.25}
    \resizebox{\linewidth}{!}{%
    \begin{tabular}{@{}lcccccccc@{}}
        \toprule
        & & \multicolumn{2}{c}{Final score} & \multicolumn{2}{c}{Drawdown [pts]} & & & \\
        \cmidrule(lr){3-4}\cmidrule(lr){5-6}
        Search & Succ. [pts] & Mean & $\text{\acs*{CVaR}}_{10}$ & Mean & $\text{\acs*{CVaR}}_{90}$ & \acs*{KL} & \acs*{LDJ} & Jump\\
        \midrule
        Nominal & \pd{$\mathbf{-55.0}$}{$-62.5,-47.0$} & \pd{$\mathbf{-.267}$}{$-.312,-.224$} & \pd{$\mathbf{-.877}$}{$-.930,-.777$} & \pd{$\mathbf{+14.7}$}{$+11.6,+17.9$} & \pd{$\mathbf{+20.5}$}{$+11.6,+29.4$} & \pd{$\mathbf{+.352}$}{$+.208,+.554$} & \pd{$+0.01$}{$-0.09,+0.12$} & \pd{$-.018$}{$-.046,+.010$}\\
        Adam$^\dagger$ & \pd{$\mathbf{-26.0}$}{$-34.0,-18.0$} & \pd{$\mathbf{-.038}$}{$-.058,-.023$} & \pd{$\mathbf{-.230}$}{$-.398,-.095$} & \pd{$+0.5$}{$-1.3,+2.5$} & \pd{$+4.5$}{$-1.1,+9.6$} & \pd{$+.034$}{$-.083,+.086$} & \pd{$\mathbf{-0.19}$}{$-0.32,-0.06$} & \pd{$\mathbf{+.091}$}{$+.060,+.122$}\\
        \acs*{L-BFGS}$^\dagger$ & \pd{$\mathbf{-16.5}$}{$-24.5,-8.5$} & \pd{$\mathbf{-.017}$}{$-.033,-.006$} & \pd{$\mathbf{-.112}$}{$-.252,-.015$} & \pd{$+1.0$}{$-0.7,+2.7$} & \pd{$+0.4$}{$-4.7,+5.2$} & \pd{$+.000$}{$-.110,+.027$} & \pd{$\mathbf{-0.22}$}{$-0.36,-0.07$} & \pd{$\mathbf{+.100}$}{$+.064,+.137$}\\
        Best-of-$N$ & \pd{$-4.0$}{$-11.0,+3.0$} & \pd{$-.016$}{$-.034,+.000$} & \pd{$\mathbf{-.109}$}{$-.215,-.006$} & \pd{$+1.3$}{$-0.5,+3.0$} & \pd{$+3.5$}{$-1.6,+8.3$} & \pd{$+.042$}{$-.077,+.137$} & \pd{$+0.06$}{$-0.05,+0.16$} & \pd{$\mathbf{-.049}$}{$-.075,-.023$}\\
        \acs*{CEM} & \pd{$-7.5$}{$-15.0,+0.0$} & \pd{$-.006$}{$-.019,+.007$} & \pd{$-.029$}{$-.160,+.097$} & \pd{$+0.1$}{$-1.5,+1.7$} & \pd{$+0.3$}{$-5.0,+4.3$} & \pd{$+.003$}{$-.100,+.133$} & \pd{$\mathbf{-0.35}$}{$-0.47,-0.23$} & \pd{$\mathbf{+.037}$}{$+.009,+.066$}\\
        \acs*{MPPI} & \pd{$-3.5$}{$-10.0,+3.0$} & \pd{$+.003$}{$-.004,+.013$} & \pd{$+.037$}{$-.016,+.131$} & \pd{$-0.1$}{$-1.5,+1.2$} & \pd{$-0.2$}{$-5.2,+4.3$} & \pd{$+.029$}{$-.077,+.102$} & \pd{$-0.04$}{$-0.16,+0.08$} & \pd{$\mathbf{+.044}$}{$+.012,+.077$}\\
        \acs*{CMA-ES}, no band & \pd{$\mathbf{+9.5}$}{$+4.0,+14.5$} & \pd{$\mathbf{+.007}$}{$+.000,+.019$} & \pd{$\mathbf{+.064}$}{$+.008,+.163$} & \pd{$\mathbf{-3.3}$}{$-5.1,-1.6$} & \pd{$-1.1$}{$-6.8,+3.6$} & \pd{$+.011$}{$-.102,+.061$} & \pd{$\mathbf{-1.03}$}{$-1.15,-0.91$} & \pd{$\mathbf{+.119}$}{$+.082,+.157$}\\
        \acs*{CMA-ES}, sphere & \pd{$-2.0$}{$-9.0,+5.0$} & \pd{$\mathbf{-.012}$}{$-.024,-.003$} & \pd{$\mathbf{-.083}$}{$-.190,-.017$} & \pd{$+0.6$}{$-1.1,+2.3$} & \pd{$+2.5$}{$-3.1,+7.1$} & \pd{$+.021$}{$-.094,+.053$} & \pd{$\mathbf{+0.31}$}{$+0.18,+0.43$} & \pd{$-.006$}{$-.035,+.025$}\\
        Ours, $\zeta=2$ & \pd{$+0.5$}{$-6.5,+7.5$} & \pd{$-.001$}{$-.014,+.013$} & \pd{$-.004$}{$-.123,+.131$} & \pd{$-0.5$}{$-2.0,+0.9$} & \pd{$-1.7$}{$-6.0,+1.8$} & \pd{$-.000$}{$-.107,+.130$} & \pd{$\mathbf{+0.15}$}{$+0.03,+0.27$} & \pd{$-.004$}{$-.038,+.031$}\\
        Ours, $\zeta=3$ & \pd{$-3.5$}{$-10.0,+3.0$} & \pd{$\mathbf{-.005}$}{$-.009,-.001$} & \pd{$\mathbf{-.022}$}{$-.041,-.004$} & \pd{$-0.3$}{$-1.8,+1.2$} & \pd{$+3.3$}{$-2.2,+8.5$} & \pd{$-.004$}{$-.068,+.088$} & \pd{$+0.02$}{$-0.10,+0.14$} & \pd{$+.001$}{$-.032,+.036$}\\
        \bottomrule
    \end{tabular}}
\end{table}
\FloatBarrier

%% file: sections/appendix/nutassembly.tex
\subsection{Two-Nut Assembly}
\label{app:nut_results}
We repeat the funnel ablation on MimicGen's two-nut assembly~\citep{mandlekarMimicGenDataGeneration2023}, in which a Sawyer arm threads a square nut onto one peg and then a round nut onto the other. The policy is the \ac{DiT}-\ac{RMFM} of stacking, trained on the task's 1000 demonstrations, and it is refined against the phased potential of Appendix~\ref{app:nut_candidate}, against which every learned and calibrated funnel was fitted. Search and protocol are those of the main text, and Tables~\ref{tab:task_choices}--\ref{tab:controller_gains} give the remaining settings.

\paragraph{Measuring progress.}
PushT and stacking are scored by a measure separate from $V$. Two-nut assembly has no such measure finer than the share of seated nuts, $r\in\{0,\tfrac12,1\}$, which falls in only 2 of the 6400 episodes and therefore says little about transient regressions. We report success and this native score, neither of which depends on $V$, and compute the final score, drawdown and \ac{KL} on the V-score $1-V/V_0$, where $V_0=85.3$ is the potential at the reset state, the same in every episode. The V-score derives from the potential that also defines the refinement objectives, so its statistics describe how each configuration uses that potential.

\begin{table}[h]
    \centering
    \caption{Funnel ablation on two-nut assembly, 400 episodes per configuration. Native: share of seated nuts. V-score: $1-V/V_0$. Shading: {\fboxsep=1pt\colorbox{green!40}{\textbf{best}}, \colorbox{green!22}{second} and \colorbox{green!9}{third}} refined configuration in each column. $^{+}$/$^{-}$: significantly better/worse than endpoint-only (paired 95\% bootstrap). Verified: complete executed chunks and nominal plans passing estimated verification (Appendix~\ref{app:verification_details}).}
    \label{tab:nut_results}
    \scriptsize
    \setlength{\tabcolsep}{2.2pt}
    \def\gsep{\hspace{5pt}{\color{black!55}\vrule width 0.5pt}\hspace{2pt}}%
    \def\sp{\rlap{$^{+}$}}\def\sm{\rlap{$^{-}$}}%
    \setlength{\aboverulesep}{0pt}\setlength{\belowrulesep}{0pt}%
    \setlength{\extrarowheight}{1.5pt}
    \begin{tabular}{@{}ll@{\gsep}r@{\gsep}r@{\gsep}r@{\hspace{7pt}}r@{\gsep}r@{\hspace{7pt}}r@{\gsep}r@{\gsep}r@{\hspace{5pt}}r@{\hspace{4pt}}}
        \toprule
        & & \multicolumn{1}{c@{\gsep}}{Success} & \multicolumn{1}{c@{\gsep}}{Native} & \multicolumn{5}{c@{\gsep}}{V-score} & \multicolumn{2}{c}{Verified [\%]}\\
        \cmidrule(lr){5-9}\cmidrule(lr){10-11}
        & & \multicolumn{1}{c@{\gsep}}{[\%]} & \multicolumn{1}{c@{\gsep}}{score} & \multicolumn{2}{c@{\gsep}}{Final $\uparrow$} & \multicolumn{2}{c@{\gsep}}{Drawdown [pts] $\downarrow$} & \multicolumn{1}{c@{\gsep}}{\acs*{KL}} & & \\
        \cmidrule(lr){5-6}\cmidrule(lr){7-8}
        Funnel & & \multicolumn{1}{c@{\gsep}}{$\uparrow$} & \multicolumn{1}{c@{\gsep}}{mean $\uparrow$} & \multicolumn{1}{c}{Mean} & \multicolumn{1}{c@{\gsep}}{$\text{\acs*{CVaR}}_{10}$} & \multicolumn{1}{c}{Mean} & \multicolumn{1}{c@{\gsep}}{$\text{\acs*{CVaR}}_{90}$} & \multicolumn{1}{c@{\gsep}}{$\downarrow$} & \multicolumn{1}{c}{Executed $\uparrow$} & \multicolumn{1}{c}{Nominal}\\
        \midrule
        Nominal & & 80.3 & .873\sm & .875\sm & .205\sm & 5.4\sm & 43.2\sm & .192\sm & \multicolumn{2}{c}{n/a}\\
        \cmidrule[0.2pt]{1-11}
        Endpoint-only & & 84.5 & .915 & .923 & \rkC{.401} & 3.6 & 30.7 & .060 & \multicolumn{2}{c}{n/a}\\
        \cmidrule[0.4pt]{1-11}
        \multirow{2}{*}{No envelope} & $\dot V\le0$ & 86.8 & \rkC{.927} & \rkC{.928} & .394 & 3.4 & 30.6 & .080 & \multicolumn{2}{c}{n/a}\\
         & $\Sigma V$ & 86.3 & .920 & .923 & .358 & \rkB{2.8} & 24.3 & .057 & \multicolumn{2}{c}{n/a}\\
        \cmidrule[0.2pt]{1-11}
        \multirow{3}{*}{Plateau} & $\delta=0.5$ & 83.0 & .907 & .923 & \rkC{.401} & 3.5 & 30.2 & \rkC{.055} & 32.5 & 32.9\\
         & $\delta=1$ & 84.3 & .912 & .919 & .390 & 4.1 & 35.4 & .071 & 31.5 & 32.2\\
         & $\delta=3$ & \rkA{88.8}\sp & \rkA{.936}\sp & \rkB{.935} & \rkB{.410} & \rkC{3.1} & 25.3 & .060 & 32.7 & 33.5\\
        \cmidrule[0.2pt]{1-11}
        Linear & & 78.5\sm & .877\sm & .881\sm & .323\sm & 4.9 & 41.3 & .144\sm & 1.6 & 13.9\\
        \cmidrule[0.2pt]{1-11}
        \multirow{2}{*}{Calibrated} & per stage & 80.5 & .884\sm & .894\sm & .300\sm & 4.7 & 40.5 & .113 & \rkA{62.2} & 60.0\\
         & global & 81.5 & .895 & .898\sm & .338\sm & 4.7 & 40.9 & .133\sm & \rkC{44.1} & 44.2\\
        \cmidrule[0.2pt]{1-11}
        \multirow{3}{*}{Learned add.} & $\alpha=0.5$ & 86.5 & .925 & .926 & .372 & \rkB{2.8} & \rkC{23.3} & \rkB{.047} & 37.4 & 20.2\\
         & $\alpha=0.8$ & 86.5 & .922 & .927 & .353 & 3.2 & 27.3 & .061 & 41.2 & 30.9\\
         & $\alpha=0.95$ & \rkC{87.8} & \rkC{.927} & .927 & .338 & 3.4 & 28.9 & .085 & 44.0 & 39.3\\
        \cmidrule[0.2pt]{1-11}
        \multirow{3}{*}{Learned mult.} & $\alpha=0.5$ & \rkB{88.3} & \rkB{.934} & \rkA{.937} & \rkA{.432} & \rkA{2.6} & \rkA{21.2} & \rkA{.034} & 38.9 & 21.8\\
         & $\alpha=0.8$ & 85.3 & .915 & .923 & .349 & \rkA{2.6} & \rkB{22.0} & \rkC{.055} & 41.5 & 31.1\\
         & $\alpha=0.95$ & 87.3 & .926 & .921 & .349 & 3.4 & 29.2 & .083 & \rkB{44.7} & 39.5\\
        \bottomrule
    \end{tabular}
\end{table}

\begin{figure}[h]
    \centering
    \includegraphics[width=\linewidth]{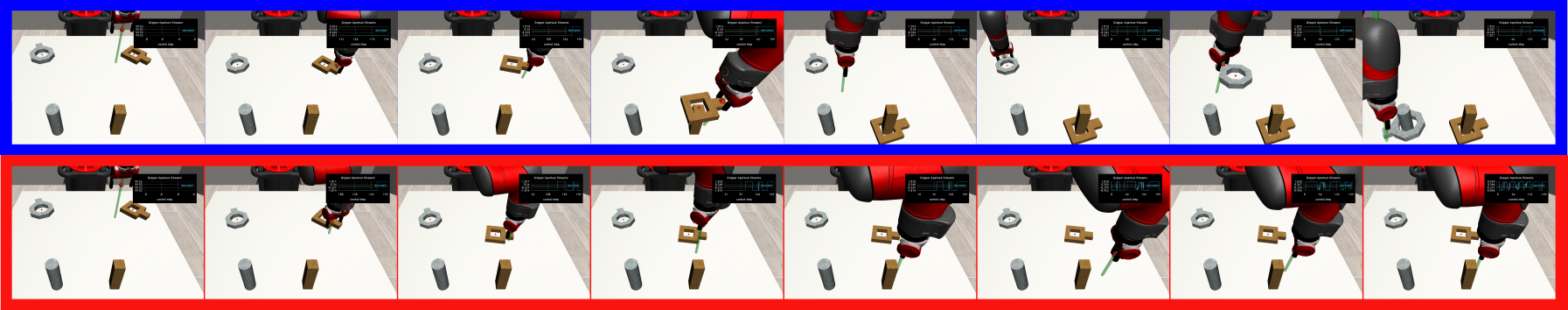}
    \caption{Representative \textcolor{blue}{refined} (top) and \textcolor{red}{nominal} (bottom) episodes on two-nut assembly from the same initial condition, time running left to right. The refined execution seats the square nut and then the round one; the nominal execution never lifts the square nut and stays stuck at it for the rest of the episode.}
    \label{fig:nut_episodes}
\end{figure}

\paragraph{Results.}
Refinement again improves clearly on nominal execution. It raises success from 80.3\% to 84.5\% with endpoint-only and to 85--89\% with $\dot V\le0$, $\Sigma V$, the learned funnels and the plateau at $\delta=3$ (Table~\ref{tab:nut_results}). Endpoint-only also raises the native score from .873 to .915 and lowers the $\text{\acs*{CVaR}}_{90}$ of the V-score drawdown from 43.2 to 30.7 points. The configurations that bound the interior stay at this level of progress retention, with confidence intervals for the V-score drawdown against endpoint-only that contain zero. The learned multiplicative funnel at $\alpha=0.5$ has the lowest values, a $\text{\acs*{CVaR}}_{90}$ of 21.2 points and a \ac{KL} of .034. The plateau at $\delta=3$ solves significantly more episodes than endpoint-only, 17 more. The calibrated funnels and the linear funnel lie close to nominal execution. They are significantly below endpoint-only in the final V-score and its $\text{\acs*{CVaR}}_{10}$, and the linear funnel solves 24 fewer episodes, as on stacking. The learned funnels keep the success of endpoint-only and make about 40\% of the executed chunks pass estimated verification, compared with 20 to 40\% of the nominal plans (Appendix~\ref{app:verification_details}). Figure~\ref{fig:nut_episodes} shows a pronounced case of the failures that refinement resolves. From the same initial condition, the nominal execution keeps pushing against the square nut and stays there for the rest of the episode. The refined execution grasps the square nut, seats it on its peg, and then picks up the round nut and seats it on the other peg, which completes the task.

\paragraph{Why terminal contraction suffices here.}
Progress on two-nut assembly comes in two discrete steps, one per seated nut, and the phased potential decreases along the motions that achieve them. Requiring $V$ to fall from one chunk boundary to the next therefore already selects the plans that advance toward the next seat or keep what was achieved. A nut dropped within a chunk ends it at a higher $V$, which terminal contraction rejects, as a toppled tower does on stacking. The V-score is consistent with this. The drawdown under the interior constraints equals that under endpoint-only within the confidence intervals, and its largest values, 40 to 48 points, have the size of the progress of a whole nut. Two-nut assembly is the opposite case to PushT. On PushT the task score registers progress inside a chunk, and the interior bound improves retention (\Cref{sec:experiments}). On two-nut assembly the interior bound of the funnel contributes estimated verification of the transient at the success level of endpoint-only.

%% file: sections/appendix/compute.tex
\section{Computational Cost}
\label{app:compute}
\paragraph{Setup.}
Refinement splits its work between the GPU and the CPU. The flow decodes the whole population of a generation in one batched call on the GPU, with its integration steps unrolled and compiled into a single kernel by Reactant.jl. The decoded plans are then rolled out on CPU threads, where each worker process keeps a pool of simulator copies, resets them to the planning state, and evaluates the candidates in parallel. All timings were measured on a workstation with a 32-core AMD Ryzen Threadripper 9970X and an NVIDIA RTX PRO 5000 Blackwell GPU. Each experiment runs several worker processes that share the GPU, with the worker layouts given in Tables~\ref{tab:compute_funnels} and~\ref{tab:compute_search}. Times are wall-clock per planning call, including flow decoding, rollouts and the search, and exclude compilation: the first episode of each worker, during which compilation may occur, is left out, and in the search ablation the samplers were compiled beforehand.

\begin{table}[!htbp]
    \centering
    \caption{Wall-clock planning time per chunk [s] in the funnel ablation. Each refined chunk is a \acs*{CMA-ES} search with population 20, at most 30 generations and the patience of Table~\ref{tab:search_settings}. Nominal execution decodes one plan. PushT uses the U-Net-\acs*{OT} policy, stacking and two-nut assembly the \acs*{DiT}-\acs*{RMFM} policy.}
    \label{tab:compute_funnels}
    \small
    \begin{tabular}{@{}llccc@{}}
        \toprule
        Funnel & & PushT & Stacking & Two-nut assembly\\
        \midrule
        Nominal &  & 0.01 & 0.01 & 0.01\\
        Endpoint-only &  & 1.17 & 3.84 & 6.18\\
        \midrule
        No envelope & $\dot V\le0$ & 0.77 & 2.02 & 3.26\\
         & $\Sigma V$ & 1.41 & 4.97 & 8.34\\
        \midrule
        Plateau & $\delta=0.5$ & 1.06 & 3.93 & 6.11\\
         & $\delta=1$ & 1.12 & 3.76 & 6.20\\
         & $\delta=3$ & 1.14 & 4.09 & 6.79\\
        \midrule
        Linear &  & 1.10 & 4.75 & 6.96\\
        \midrule
        Calibrated & per stage & -- & 3.38 & 4.53\\
         & global & 1.30 & 3.44 & 4.87\\
        \midrule
        Learned add. & $\alpha=0.5$ & 1.14 & 4.02 & 5.94\\
         & $\alpha=0.8$ & 1.02 & 3.80 & 5.75\\
         & $\alpha=0.95$ & 0.94 & 3.74 & 5.60\\
        \midrule
        Learned mult. & $\alpha=0.5$ & 1.17 & 3.90 & 5.84\\
         & $\alpha=0.8$ & 1.00 & 3.81 & 5.62\\
         & $\alpha=0.95$ & 0.99 & 3.81 & 5.56\\
        \midrule
        \multicolumn{2}{@{}l}{Workers $\times$ threads} & $10\times4$ & $14\times4$ & $14\times4$\\
        \bottomrule
    \end{tabular}
\end{table}

\paragraph{Funnel ablation.}
Refinement costs 2.0 to 5.0\,s per chunk on stacking, 3.3 to 8.3\,s on two-nut assembly and 0.8 to 1.4\,s on PushT, against 0.01\,s for the nominal policy (Table~\ref{tab:compute_funnels}). Every candidate costs one rollout regardless of the objective, so the objectives differ in cost through when the search stops: it ends when its best plan stops improving meaningfully, after fewer generations once that plan is verified (Table~\ref{tab:search_settings}). In all three tasks, $\dot V\le0$ is the cheapest objective and $\Sigma V$ the most expensive. On stacking and two-nut assembly the MuJoCo rollouts dominate. On PushT, where the simulator is cheap, decoding with the flow does. Per episode, refinement adds about 18.5\,s of planning to a stacking episode when episodes run one at a time, and about 36\,s when 14 worker processes of 4 threads run episodes in parallel, against 0.1\,s for the nominal policy. Decoding the flow on the CPU instead of the GPU roughly doubles the planning time.

\begin{table}[!htbp]
    \centering
    \caption{Compute in the PushT search ablation, ten worker processes of four threads sharing one GPU. Iterations: generations, batches or gradient steps per chunk. Evaluations: plans evaluated per chunk.}
    \label{tab:compute_search}
    \small
    \begin{tabular}{@{}lccc@{}}
        \toprule
        Search & Time [s] & Iterations & Evaluations\\
        \midrule
        Nominal & 0.01 & -- & 1\\
        Adam$^\dagger$ & 1.33 & 20.3 & 21\\
        \acs*{L-BFGS}$^\dagger$ & 1.35 & 14.5 & 36\\
        Best-of-$N$ & 0.20 & 1 & 200\\
        \acs*{CEM} & 1.52 & 40.4 & 808\\
        \acs*{MPPI} & 1.52 & 42.0 & 841\\
        \acs*{CMA-ES}, no band & 1.52 & 40.8 & 815\\
        \acs*{CMA-ES}, sphere & 1.52 & 40.6 & 811\\
        Ours, $\zeta=2$ & 1.5 & 40 & 800\\
        Ours, $\zeta=3$ & 1.5 & 40 & 800\\
        Ours, $\zeta=4$ & 1.52 & 40.3 & 806\\
        \bottomrule
    \end{tabular}
\end{table}

\paragraph{Search ablation.}
Within the 1.5\,s budget, the population searches run 40 to 42 generations, or 806 to 841 evaluated plans (Table~\ref{tab:compute_search}). Of their 1.52\,s per chunk, decoding the population with the flow takes 1.45\,s and scoring the candidates in the native simulator 0.06\,s. Best-of-$N$ decodes and scores its single batch of 200 plans in 0.20\,s. For the gradient searches, each evaluation is a flow pass at batch size one plus, for every gradient, its vector--Jacobian product. Per chunk, Adam and \ac{L-BFGS} spend 0.84 and 0.63\,s in the flow's vector--Jacobian product, 0.30 and 0.57\,s in its forward pass, and 0.16 and 0.11\,s in the model gradient.
\FloatBarrier

%% file: sections/appendix/limitations.tex
\section{Extended Limitations}
\label{app:limitations}

\paragraph{The task potential.}
Refinement depends on the task potential. Our potentials are designed by hand from task geometry, with constants calibrated on demonstrations, and they read privileged simulator state. The search optimizes $V$, and the native task outcome improves when $V$ follows the task through its intermediate stages, as with the phased potentials of Appendix~\ref{app:potential_ablation}. Future work should replace the hand-designed potential with a learned progress measure that relates to completion and is regular enough for verification. The learned representations and reward models discussed in Appendix~\ref{app:related_work} could be starting points.

\paragraph{World model.}
Refinement rolls out candidates in the simulator that also executes them, so the evaluated rollout matches execution, and it reads the full state of the scene. On a physical system, the rollouts would come from a calibrated simulator or from a learned world model, and the mismatch between predicted and executed motion would enter the convergence result as the endpoint error of \Cref{cor:practical_contraction}. Learned simulators of contact-rich rigid-body dynamics~\citep{allenLearningRigidDynamics2023} and latent world models~\citep{zhouDINOWM2025,assranVJEPA22025} are candidates for this role. How model error affects funnel satisfaction is an open question.

\paragraph{Search and guarantees.}
Refinement reaches useful motions through the prior within a finite search budget and executes its lowest-objective plan together with its estimated-verification status. The convergence results hold under funnel satisfaction of the executed chunks and agreement between simulated rollouts and execution. Two directions would strengthen them: conditions under which a satisfying chunk exists at every planning step, which would give recursive feasibility, and certified bounds on the residual rate, such as the one of \Cref{lem:residual_regularity}, in place of the estimated rates.

\paragraph{Optimization freedom and fidelity to the prior.}
Any refinement of a generative prior trades optimization freedom against fidelity to the prior. The more the optimizer can change a motion, the lower it can drive the objective and the constraint violation, and the further the motion can move from the demonstrated distribution, where the policy, the tracking controller and the rollouts are less reliable. The radial band is one way to set this tradeoff, through its width $\zeta$. On PushT, the band-constrained search reaches 86.5\% success with 11.3\% of its motion windows out of distribution, and the unconstrained search reaches 96.0\% with 34.1\% (Table~\ref{tab:search_ablation}). The band constrains the radius and the mean of the seed and leaves its direction free, so it limits the shift of each chunk. The states that a refined chunk leads to still determine the next observation, which can differ from the demonstrated observations, the covariate shift familiar from imitation learning~\citep{rossReductionImitationLearning2011c}. A band width that adapts to the state, wider where the nominal plan is far from satisfying its funnel, is a possible extension.

\paragraph{Computation and deployment.}
The candidate rollouts within each generation run in parallel, and the generations run in sequence, which gives a planning delay of 0.8 to 8.3\,s per chunk in our experiments (Appendix~\ref{app:compute}). Physical deployment would need a scene that stays still during planning, or a prediction of its evolution over the computation interval. Learned rollout models would shorten each generation. The refined seeds could also be amortized into a network that predicts them from the observation, as in learned noise policies for diffusion models~\citep{wagenmakerSteeringDiffusionPolicy2025,eyringNoiseHypernetworks2025}.